\documentclass[
11pt, 
oneside, 
english, 
onehalfspacing, 
nolistspacing, 
headsepline, 
]{MastersDoctoralThesis} 

\usepackage[T1]{fontenc} 

\usepackage{mathpazo} 

\usepackage[autostyle=true]{csquotes} 

\usepackage{changepage}

\usepackage{enumerate}
\usepackage{bm}
\usepackage{amsthm}
\usepackage{amsmath}
\theoremstyle{definition}

\newtheorem*{theorem*}{Theorem}

\newtheorem*{lemma*}{Lemma}
\usepackage{physics}

\usepackage[export]{adjustbox}
\usepackage{subfig}
\usepackage{wrapfig}
\usepackage{threeparttable}
\usepackage{enumitem}
\usepackage{hyperref} 
\usepackage{amsmath}
\usepackage{epigraph}
\newcommand{\q}[1]{"#1"}

\usepackage{nicefrac} 
\usepackage{float} 

\thesistitle{An unsupervised tour through the hidden pathways of deep neural networks} 
\supervisor{Alessandro Laio} 
\examiner{} 
\degree{Doctor of Philosophy} 
\author{Diego Doimo} 
\addresses{} 

\subject{Biological Sciences} 
\university{\href{https://www.sissa.it/}{SISSA\\International school for advanced studies}} 
\department{\href{http://department.university.com}{Department or School Name}} 
\group{\href{http://researchgroup.university.com}{Research Group Name}} 
\faculty{\href{http://faculty.university.com}{Faculty Name}} 

\AtBeginDocument{
\hypersetup{pdftitle=\ttitle} 
\hypersetup{pdfauthor=\authorname} 
\hypersetup{pdfkeywords=\keywordnames} 
}

\begin{document}

\frontmatter 

\pagestyle{plain} 


\begin{titlepage}
\begin{center}

{\scshape\LARGE \univname\par}\vspace{1.5cm} 
\textsc{\Large Doctoral Thesis}\\[0.5cm] 

\HRule \\[0.4cm] 
{\huge \bfseries \ttitle \par}\vspace{0.4cm} 
\HRule \\[1.5cm] 
 
\begin{minipage}[t]{0.4\textwidth}
\begin{flushleft} \large
\emph{Author:}\\
\authorname
\end{flushleft}
\end{minipage}
\begin{minipage}[t]{0.4\textwidth}
\begin{flushright} \large
\emph{Supervisor:} \\
\supname
\end{flushright}
\end{minipage}
 
\vspace{50pt}
\includegraphics[width = 0.18\columnwidth]{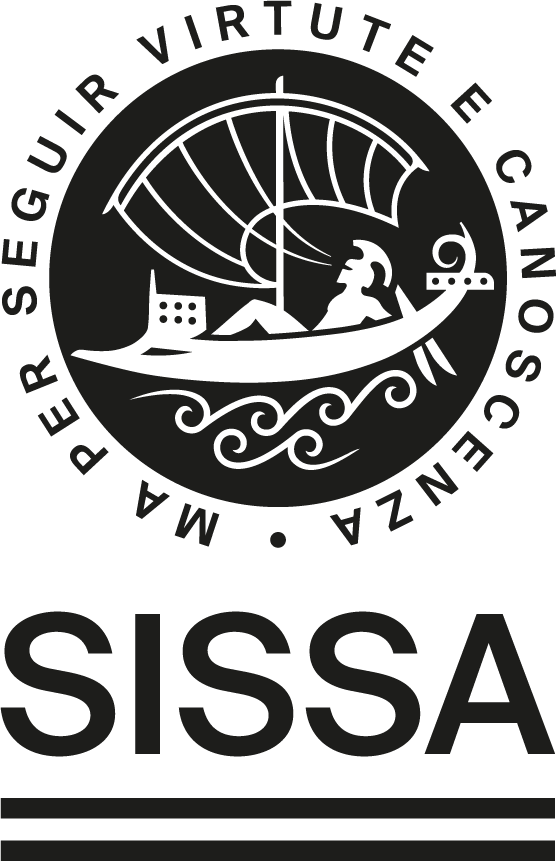} 

\vspace{30pt}
\textsc{\large PHD course in \\  Physics and Chemistry of Biological Systems}

\vspace{10pt}{Academic Year 2021/2022}

\vfill

\end{center}
\end{titlepage}

\begin{acknowledgements}
\addchaptertocentry{\acknowledgementname}

\begin{adjustwidth}{0cm}{2.5cm} 
\itshape
\vspace{25pt}
Ad Alessandro per la sua presenza e il tempo che mi ha dedicato. Per avermi mostrato un esempio di dedizione, competenza e soprattutto umiltà \mbox{orientate alla ricerca e alla supervisione scientifica}. 
\vspace{6pt}

A Mara per la strada fatta assieme e per essermi stata vicina nei momenti \mbox{in cui} le mie fragilità si sono fatte maggiormente sentire.
\vspace{6pt}

Alla mia famiglia, a Erica, a mio papà e a mia mamma, a cui devo le parti migliori di me.

\bigbreak

\hfill\noindent{\large Ottobre 2025}

\end{adjustwidth}
\end{acknowledgements}

\vfill
\vfill




\clearpage


\cleardoublepage

~\vfill

\begin{flushright}
\begin{minipage}{0.64\textwidth}
\itshape \large
Non domandarci la formula che mondi possa aprirti.
\medskip

\normalfont Eugenio Montale, \emph{Ossi di Seppia}, 1925.
\end{minipage}
\end{flushright}

\vfill\vfill


\begin{abstract}

The goal of this thesis is to improve our understanding of the internal mechanisms by which deep artificial neural networks create meaningful representations and are able to generalize.
We focus on the challenge of characterizing the semantic content of the hidden representations with unsupervised learning tools, partially developed by us and described in this thesis, which allow harnessing the low-dimensional structure of the data. 

Real-world data are indeed hosted in manifolds that can be topologically complex, but are low-dimensional. Chapter 2. introduces Gride, a method that allows estimating the intrinsic dimension of the data as an explicit function of the scale without performing any decimation of the data set. Our approach is based on rigorous distributional results that enable the quantification of uncertainty of the estimates. Moreover, our method is simple and computationally efficient since it relies only on the distances among nearest data points.

In Chapter 3, we study the evolution of the probability density across the hidden layers in some state-of-the-art deep neural networks. We find that the initial layers generate a unimodal probability density getting rid of any structure irrelevant to classification. In subsequent layers, density peaks arise in a hierarchical fashion that mirrors the semantic hierarchy of the concepts. This process leaves a footprint in the probability density of the output layer, where the topography of the peaks allows reconstructing the semantic relationships of the categories.

In Chapter 4, we study the problem of generalization in deep neural networks:  adding parameters to a network that interpolates its training data will typically improve its generalization performance, at odds with the classical bias-variance trade-off. 
We show that wide neural networks learn redundant representations instead of overfitting to spurious correlation and that redundant neurons appear only if the network is regularized and the training error is zero.

\bigbreak

\hfill\noindent{\large 17 November 2022}

\end{abstract}


\tableofcontents 

\mainmatter 

\pagestyle{thesis} 


\chapter{Introduction}
\label{ch:introduction}


\begin{flushright}
\begin{minipage}{0.52\textwidth}
\itshape  
There is no doubt whatever that all our cognition \\ begins with experience.
\medskip

\normalfont Immanuel Kant, \emph{Critique of Pure Reason}, 1781.
\end{minipage}
\end{flushright}


\vspace{20pt}

The opening  statement of Kant's Critique of Pure Reason, 
is undoubtedly correct for today's machine learning algorithms. 
These draw their knowledge from the \enquote{experience}, consisting of web text accumulated over more than ten years \cite{common_crawl} and from tens or hundreds of millions of images \cite{jft_distilling, tencent_large_scale}. This allowed reaching astounding performances in automatic translation, computer vision, and several other tasks \cite{gpt3, dalle, stable_diffusion, clip}.
This huge amount of data contains information with an intricate interplay of semantic relations on multiple levels of abstraction.
Deep learning models, if fed with large amounts of data, can unveil this broad and diverse spectrum of relations with little or almost no need for human-crafted features and with a simple, general-purpose learning algorithm \cite{backprop1, backprop2}.

The term abstraction etymologically comes from the Latin \emph{abstrahere} (\enquote{\emph{abs-}} and  \enquote{\emph{-trahere}}) meaning \enquote{to drag away from},  \enquote{to split}, and denotes a separation between two things, for instance, an idea or concept that is separated from material reality. 
The remote connection between, say, the concepts represented in an image and the raw input pixels mathematically corresponds to a complicated, highly non-linear dependence between them, extremely hard to capture in a single step, or \enquote{layer}, for any learning algorithm \cite{bengio2009learning}.
Deep learning decomposes the complex relation between pixels and high-level concepts in multiple levels of intermediate abstractions, which gradually disentangle a multitude of features.

Learning a meaningful hierarchy of representations from a data set is not easy. Indeed, before the introduction of Deep Belief Networks and layerwise pretraining in 2006 \cite{deep_belief}, training more than two layers of representations was considered detrimental to performance \cite{bengio2009learning, bengio2013representation, lecun2015deep}. 
Between 2009 and 2012, Graphic Processing Units (GPUs) allowed training deep models on extensive amounts of data \cite{imagenet09}, showing that deep architectures could be successfully trained end-to-end without layerwise pertaining \cite{2010gpu_ciresian, 2011gpu_xavier, krizhevsky2012imagenet}.
Perhaps one of the most celebrated and paradigmatic successes of deep learning was the winning of the ILSVRC competition in 2012 by AlexNet's 8-layer convolutional network that reduced the classification error on the ImageNet data set \cite{imagenet09} by almost a factor of two with respect to traditional shallow models.
From 2010 onwards more and more models were proposed, all moving towards making the hierarchy of representations deeper \cite{vgg, 2015inceptionv3, he2016deep}. By 2015, with the introduction of residual units \cite{resnet}, architectures with more than 1000 layers could be trained.

As the number of layers grew to several hundred and the number of parameters to hundreds of millions the working principle of deep learning models became less and less easy to understand \cite{efron_prediction_attribution, xai_concepts_taxonomies}. 
Improving our understanding of the internal mechanisms by which  deep architectures build meaningful representations is not an academic question: deep learning models are not only widespread in scientific applications such as
bioinformatics \cite{deep_learning_science_bioinformatics1, deep_learning_science_bioinformatics2, deep_learning_science_bioinformatics3}
 and
astronomy  \cite{deep_learning_astronomy}, but have also \enquote{invaded}  many sectors of our lives, from self-driving cars \cite{deep_learning_autonomous1, deep_learning_autonomous2}, to recommender systems \cite{deep_learning_recommender} and, very importantly,  medicine \cite{deep_learning_medicine1, deep_learning_medicine2, deep_learning_medicine3}.
Interpretability is a particularly sensitive issue when the decision-making process has an impact on fields such as medicine, targeted advertising, and judicial systems, since in those applications, ethical concerns are at stake. In these situations, trustworthiness and transparency are a \emph{conditio sine-qua-non} for the adoption of an algorithm in practice.

The goal of this thesis is to take part in a big effort that involves many scientists worldwide,  looking inside the black box of deep models. In particular, we focus on the challenge of characterizing the semantic content of the hidden representation.
While we will mostly focus on deep vision models, the methods we use are model-agnostic and can be applied to any machine learning algorithm and, we hope, will also help in understanding how deep models work in other fields.
Due to the pivotal importance of \enquote{explainebleAI} \cite{arrieta2020xai, zhu2018explainable, xu2019explainable}, a great deal of research has been done to make deep learning more transparent.
In the context of deep vision models, the methods used can be broadly divided into two groups \cite{arrieta2020xai}: 
attribution methods aimed at understanding the relevance of the input features or input patches for explaining the output prediction \cite{pixelwiseLRP, zeiler2010deconvolutional, goyal2016towards, saliency}, 
and methods that aim at characterizing how intermediate representations see the external world
\cite{mahendran2015understanding,  2017dissection, 2017tsne_repr, 2015pca_repr}. 
Our approach belongs to the second class of methods.
%
%
Like \cite{2017dissection, 2017tsne_repr, 2015pca_repr}, we analyze the semantics of a representation, looking at the global structure of the data set, but we neither rely on low-dimensional visualizations  \cite{2017tsne_repr, 2015pca_repr} nor we require extensive pixel-wise labeling of the images to infer the semantics of a layer \cite{2017dissection}.

We will instead analyze the hidden layer representations of the networks with unsupervised learning tools, partially developed by us and described in this Thesis, which allow harnessing the low-dimensional structure of the data without performing any explicit dimensional reduction.
Indeed, despite the wide range of applications, tasks, and sources, a \emph{fil rouge} characterizing real-world data is that they do not fill all the available space uniformly, but instead are typically hosted in manifolds that can be topologically complex, but which are typically low-dimensional.
%
%
In Chapter \ref{ch:hier-nucl} we will see how the density distribution of the data points in every single layer can be characterized explicitly, extracting the topography of the probability peaks.
As we will see, this analysis allows capturing quantitatively an essential feature of deep models: the semantic content of the representation in deep models, measured by this topography,  vary significantly across the layers. In deep models trained on classification tasks, the representation typically becomes linked to "abstract" categories very close to the output, but in other architecture, semantic richness can emerge also in the middle of the network.  

Another problematic aspect that defies our understanding of deep learning networks is related to their ability 
to generalize outside the training data.
Classical learning theory provides guarantees to bound the deviation of the error on unseen samples $E_{test}$ from the error measured on the training set 
$E_{train}$ with a quantity $\Omega$, connected to the model complexity: 
\begin{equation}
\label{eq:intro_error_bound}
E_{test} \leq E_{train} + \Omega.
\end{equation}
\citet{vapnik_chevr} showed that $\Omega$ depends on the data set size $N$, on a measure of \enquote{model complexity} known as VC-dimension $d_{VC}$, and on a tolerance $\delta$ on  Eq. \eqref{eq:intro_error_bound} as \cite{abu} :
\begin{align}
\label{eq:vc_omega}
    \Omega = \sqrt{\dfrac{8}{N}\log \left(\dfrac{4((2N)^{d_{VC}}+1)}{\delta}\right)}
\end{align}
$d_{VC}$ can be informally thought of as the number of effective degrees of freedom of the model. 
If it is small, then $E_{train}$ gives an approximately correct estimate of $E_{test}$ with probability $1-\delta$ for sufficiently large $N$.
For some simple algorithms $d_{VC}$ can be derived analytically. For instance, for the linear perceptron, $d_{VC}$ is exactly equal to the number of parameters \cite{abu}.
In a more general setting, when $d_{VC}$ con not be computed, the \enquote{Occam Razor} principle is applied: simple models with a moderate number of parameters are preferred, and regularization techniques are used to reduce the model complexity.
%

Deep networks go against conventional wisdom, with their number of parameters being orders of magnitude larger than the data set size. 
For instance, the networks analyzed in this Thesis have more than 10 million parameters and are trained on data sets of 50 thousand (CIFAR10/CIFAR100) or 1.2 million (ImageNet) examples.
\citet{zhang2016understanding} showed that the capacity of such models is so large that they can fit with zero training error ($E_{train}=0$) random labels or images corrupted by noise.
Under these conditions, $E_{test}$ is no better than random guessing, and the generalization gap, $E_{test}-E_{train}$, can be made arbitrarily large only by changing properties of the data set which are not captured by Eq. \ref{eq:vc_omega}.

A crucial aspect of Eq. \ref{eq:vc_omega} and of similar bounds based on model complexity \cite{bartlett2002rademacher, alon1997scale, bousquet2002stability} is that they provide general, \enquote{data-agnostic}, \emph{sufficient} conditions for generalization, often defined on worst-case data distributions \cite{generalization2019bengio}.
In recent years, more focus has been put to understand the generalization problem in deep learning in light of the relationship between the algorithm and the data structure. 
This approach is particularly meaningful in practice since deep networks are not designed to minimize some abstract notion of complexity, but 
to exploit the structure and the symmetries of the data
\cite{cohen2016group, satorras2021graphs, weiler20183d, hoogeboom2022equivariant, thomas2018tensofield, nequip2022, satorras2021flows}.

Under this perspective, \citet{generalization_error_invariant} showed that the generalization error of a classifier built to be invariant to the symmetries of the data can be much smaller than that of a non-invariant one.
\citet{generalization2019bengio} showed that if a classifier learns to concentrate a \emph{given} data set in a small region of the representation space, the generalization error is small even if on other data sets or tasks the error can be arbitrarily large.

These works point to the importance of measuring the effective \enquote{volume} in the representation space occupied by the data. 
This brings in another key working tool that can be used to characterize the representation: an estimator of the intrinsic dimension of the data.
Recent studies have shown that the intrinsic dimension of the input data is a key factor enabling good generalization in neural networks, both empirically \cite{pope2021intrinsic} and theoretically \cite{sharma2020neural, goldt2020modelling, wyart_id_generalization}.
In Chapter \ref{ch:gride} and Chapter \ref{ch:repr-mitosis}, we will explore these aspects, analyzing how the intrinsic dimension varies as a function of the depth, and how deep neural networks develop redundant representations and generalize by exploiting the low-dimensional structure of the data manifold.


%

\section{Outline of the thesis}
\paragraph{Scale analysis of the intrinsic dimension with Gride.}
To measure correctly the intrinsic dimension (ID) of complex and twisted manifolds ID estimators have to be local \cite{Campadelli2015}.
Local estimators assume that in a small neighborhood of each data point, the data distribution is less affected by curvature and can be approximated by a uniform distribution on $d$-dimensional balls.
%
One difficulty of local estimators comes from the noise in the data that typically leads to an overestimation of the ID if the size of the neighborhoods is comparable to the scale of the noise.
%
In Chapter \ref{ch:gride}, we propose an extension of the Two Nearest Neighbor estimator (TwoNN)\cite{Facco2017}, which improves its reliability on noisy data sets.

\begin{figure}
\centering
{\includegraphics[width=1.\columnwidth]{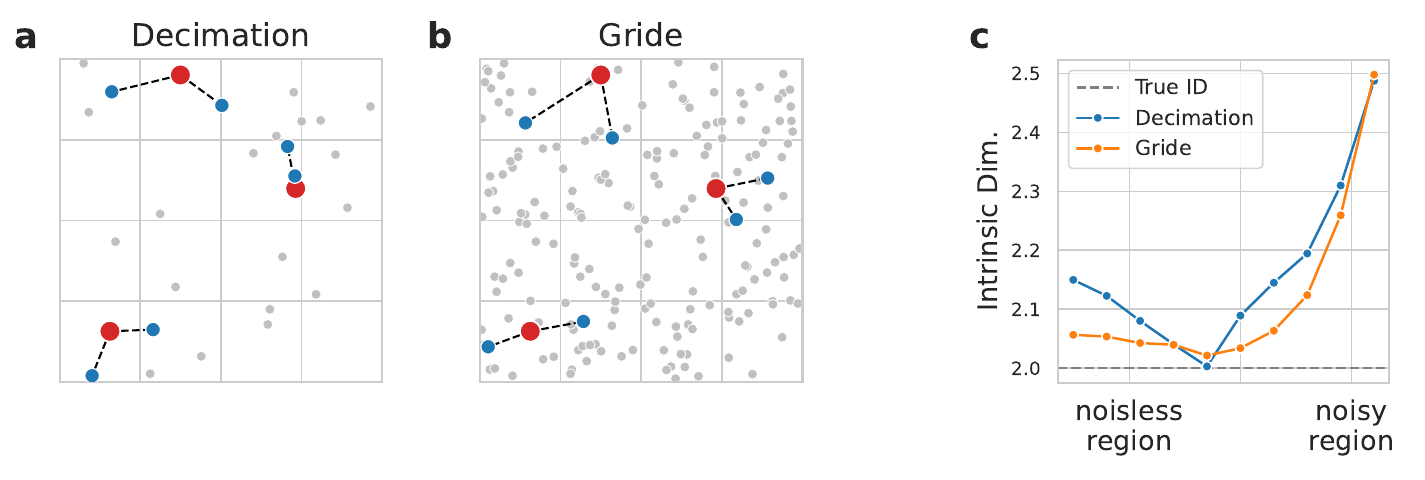}
	\caption{{\bf Comparison between Gride and TwoNN}:
\label{fig:intro_id}}}
\end{figure}

Assuming that locally the manifold is flat and the data is distributed according to a Poisson process, \citet{Facco2017} show that the ratio $\mu$ between the distances to the second and first nearest neighbor of each point is Pareto distributed, with a shape parameter given by the ID. 
The ID is the only free parameter of the likelihood and can be estimated given the observed values of $\mu$. 
Since noise in the data can lead to an overestimation of the ID, TwoNN increases the neighborhood size by decimating the data set (Fig. \ref{fig:intro_id}-a) to find a range of scales where the ID is less affected by noise.
%
%
We propose Gride, an estimator that considers nearest neighbors of higher order instead of decimating the data set (Fig. \ref{fig:intro_id}-b). 
%
This approach reduces the variance of the likelihood, giving more stable estimates across the noise scale.
Figure \ref{fig:intro_id}-c shows the case of a 2-dimensional Gaussian distribution embedded in $\mathbb{R}^3$ and perturbed by Gaussian noise of standard deviation $\sigma$. When the neighborhood size is larger than $\sigma$ Gride converges faster to the true ID and is more stable than TwoNN.
Moreover, in Sec. \ref{sec:time_benchmark} we show that the computational cost to perform the scale analysis of the ID with Gride is two times smaller than TwoNN and orders of magnitude smaller than many other neighbor-based estimators.

Due to its computational efficiency, Gride can estimate the ID on very high-dimensional spaces like those embedding the hidden representations of deep neural networks.
We conclude Chapter \ref{ch:gride} with a scale analysis of the ID of the representations of ResNet152 \cite{he2016deep} and Image GPT (iGPT) \cite{igpt} trained on ImageNet, showing an intriguing possible connection between the ID of a representation and its semantic expressiveness.

%
\paragraph{Characterization of the probability distribution of the hidden representations of deep neural networks.}
Using unsupervised learning techniques, including the tools developed in Chapter \ref{ch:gride}, in Chapter \ref{ch:hier-nucl}, we analyze the evolution of the probability distribution across the layers of several deep neural networks.
We estimate the probability density with the approach described in \citet{cluster}. In short, the local volume density $\rho_i$ around each data point is determined with a $k$NN estimator: $\rho_i = k/(NV_{k_i})$.
In this step, the volume $V_{k_i} = \omega_d r_{k_i}^d$ is measured on the data manifold using the value $d$ of the ID estimated with the approach described in Chapter \ref{ch:gride}.
The local density maxima are then found, but only those maxima that are judged sufficiently robust with respect to spurious fluctuations are considered genuine probability modes (see Sec. \ref{sec:hier_nucl_methods}). 
In the procedure, the data points are grouped around the probability modes, leading to a clustering of the data landscape.

\begin{figure}
\centering
{\includegraphics[width=1.\columnwidth]{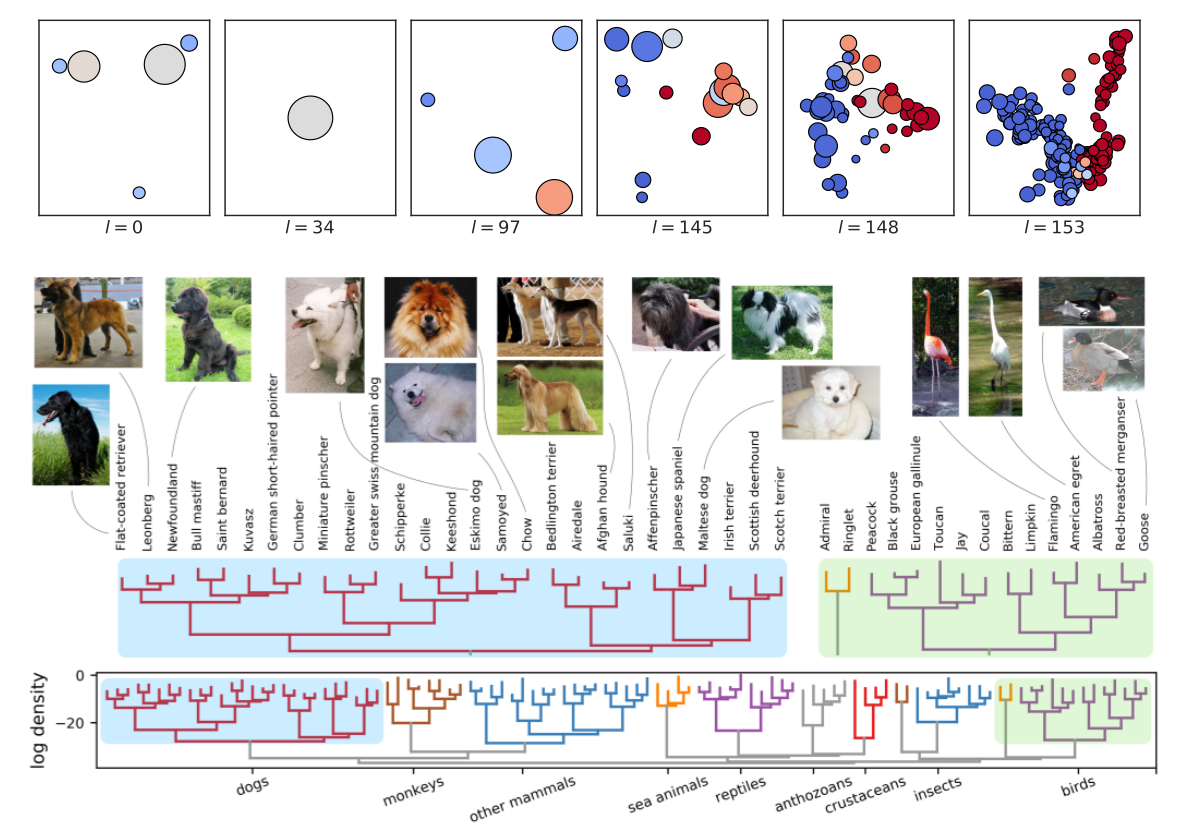}
	\caption{{\bf Density landscape of the hidden representations of ResNet152}:}
\label{fig:intro_resnet152}}
\end{figure}

We use this method to characterize the evolution of the probability density of the hidden representation of convolutional networks, of the Vision Transformer \cite{dosovitskiy2021vision_transformer}, and of the iGPT, all trained on ImageNet. 
The typical picture of a trained convolutional network is illustrated in Fig. \ref{fig:intro_resnet152} for the case of ResNet152.
The development of the probability modes is not gradual, as one would expect in a deep network with more than one hundred layers.
Instead, the greatest changes to the neighborhood composition and the emergence of the probability peaks are localized in a few layers close to the output. 
In the first layers, the network destroys the patterns found in the input that are not useful for classification, and the probability density becomes unimodal (layer 34). 
In the second stage, the probability distribution becomes multimodal with each peak being populated by a single class or group of classes according to the semantic relationships of the categories.
Indeed, the splitting follows a hierarchical order. At layer 97, two peaks appear, separating the animals from the objects, then around layer 140, the modes corresponding to the final classes emerge.
The knowledge of the general concepts is not destroyed when the fine-grained categories are learned, as in the logit space, a hierarchical clustering of the density peaks allows recovering the semantic relationships between the classes as illustrated by the dendrogram of Fig. \ref{fig:intro_resnet152}.

The analysis of Vision Transformer reveals a similar hierarchical evolution, but the process is more gradual, and the peaks appear already in the middle of the network. In iGPT the probability modes are not associated with the ImageNet categories, as the network is a generative model trained without labels. Nonetheless, we find that at a local scale, in the middle of the network, the neighborhoods are populated by data points belonging to the same class. This happens when the ID of the representation reaches a minimum.
We finally study how the density peaks arise during training by looking at the probability density in the logit representation (Sec \ref{sec:hier_nucl_dynamics_resnet152}).
We show that hierarchical learning occurs also as a function of time, first recognizing animals and objects (Fig. \ref{fig:training_dynamics}) and only gradually developing the peaks associated with the fine-grained categories of ImageNet once the neighborhoods of each data are mostly populated by alike data points (Fig. \ref{fig:dynamics_ari}).

\begin{figure}
\centering
{\includegraphics[width=1.\columnwidth]{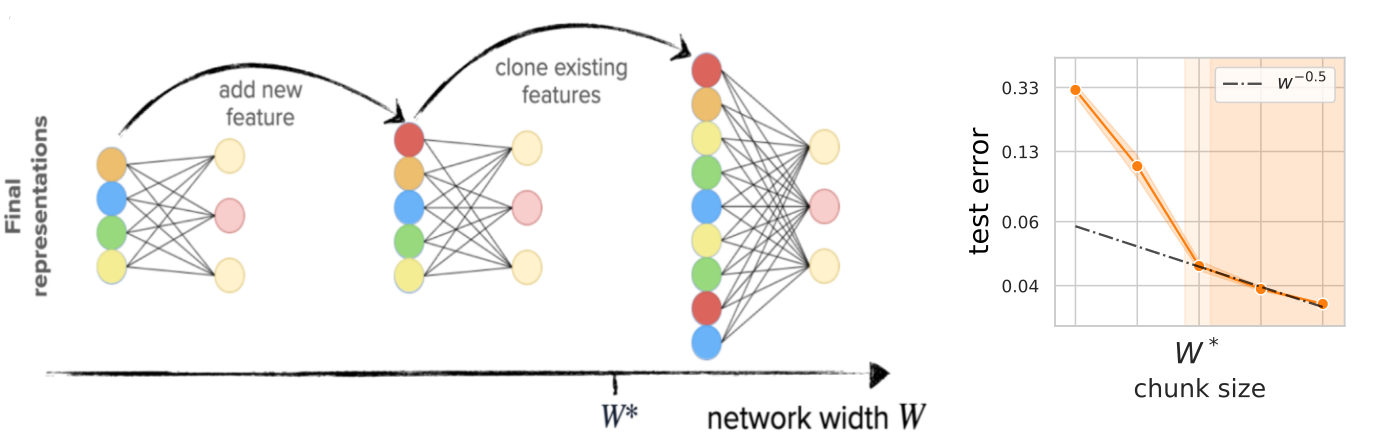}\caption{{\bf Redundant representations in wide neural networks.}:
}
\label{fig:intro_mitosis}}
\end{figure}

\paragraph{Generalization through redundant representations}
Due to the low-dimensional structure of the representation, the number of coordinates required to describe the data manifold is typically much smaller than the number of layer activations.
In Chapter \ref{ch:repr-mitosis}, we study various state-of-the-art convolutional networks (ResNets, Dense-Nets) and show that if the number of activations of the last hidden representation is large enough, the neurons tend to split into groups that carry identical information and differ from each other only by statistically independent noise. 
In other words, in the over-parametrized regime, when the width of the network is wide enough to fit the training set with zero error, neural networks learn redundant representations instead of overfitting spurious correlations.
The mechanism, pictorially represented in the left panel of Fig. \ref{fig:intro_mitosis}, is motivated by the statistical properties of the subsets of activations picked at random from those belonging to the last hidden representation.

We find that when the width of the network $W$ is above the interpolation width $W^*$, subsets of $w_c > W^*$ activations can themselves fit the training set with zero error, thereby showing that chunks with more than $W^*$ activations have captured all the relevant features for classification.
Moreover, on the test set, the decay of the classification error of random chunks as a function of $w_c$ shows different trends above and below $W^*$ (see right panel of Fig. \ref{fig:intro_mitosis}): when $w_c<W^*$ the decay of the test error is faster than $w_c^{-\nicefrac{1}{2}}$ when $w_c>W^*$ is instead compatible with $w_c^{-\nicefrac{1}{2}}$. 
We show that, above $W^*$, the decay of the test error of the chunks is similar to that of full networks that have the same number of activations $W$ as the chunks ($W = w_c$).

From a geometrical perspective, the important features of the final representation correspond to directions in which the data landscape shows large variations because the features irrelevant to classification have been suppressed by the network. 
Above $W^*$ a chunk encode almost exactly the relevant directions as it is able to \emph{linearly} reconstruct the full data representation with an $R^2$ coefficient equal to one. 
In addition, the correlations between the components of the error along different axes are particularly small and compatible with uncorrelated Gaussian noise.

\paragraph{Dadapy: a package for the distance-based unsupervised analysis of data sets.}
All the methods used in this thesis have been released in a package named Dadapy: Distance-based Analysis of DAta-manifolds in PYthon.
The details of the package are described in the companion paper published in \emph{Patterns}, a result of the work of our lab. Being a purely methodological work, the methods contained in the paper will not be described in this Thesis. 
My contribution to the collaboration consists of the implementation of the methods for the intrinsic dimension estimation, as well as the migration and optimization of the density-based clustering methods from the original implementation written in Fortran to Python. 

\vspace{10pt}
\noindent {\bf This thesis is based on the following papers}:
\vspace{5pt}


\noindent \fullcite{me_gride}
\vspace{7pt}

\noindent \fullcite{me_hier_nucl}
\vspace{7pt}

\noindent \fullcite{me_repr_mit}
\vspace{7pt}

\noindent \fullcite{me_dadapy}

\chapter{Multiscale neighbor-based intrinsic dimension estimation}
\label{ch:gride}
\section{Introduction}
Learning complex functions defined in high-dimensional spaces is an extremely hard task because the number of samples must grow exponentially with the dimension of the ambient space \cite{sample_complexity_cuts, tesiting_manifold_hypotheses}.
Luckily most real-world data sets are characterized by regularities that induce strong constraints and correlations between the features of the data. 
In natural images, nearby pixels have strongly correlated colors and intensities and often have few degrees of freedom \cite{isomap}.  Regularities occurring in text and in the sound of spoken letters \cite{id_voice_text, isolet} similarly result in a low number of relevant variations.
In molecular dynamics simulations, the region of the configuration space visited by a complex molecule of $N$ atoms has a dimensionality much smaller than the $3N$ degrees of freedom available, due to the constraints imposed by the bonds between the atoms and the steric hindrance between parts of the molecule \cite{Facco2017}. 
The number of possible variations in proteins belonging to the same family is much smaller than the sequence length \cite{granata_carnevale}
because of the conserved amino acids at some positions of their sequences and correlated variation in others.
Therefore, despite real-world data possessing a very large number of features, in practice, they often tend to concentrate around low-dimensional, often highly non-linear, manifolds. 
The dimension of the manifold is referred to as \emph{intrinsic dimension} (ID) of the data set.
In many circumstances, the knowledge of the ID also gives valuable information about the inner properties of the dataset \cite{Grassberger1983, id_phase_transition, ansuini2019intrinsic}.

Due to its theoretical and practical importance, an extensive amount of studies have proposed methods to characterize the ID of data sets based either on Principal Component Analysis or Multidimensional Scaling \cite{isomap, proj_id_lle, poj_id_sammond, proj_id_som, proj_bruske, kernelPCA, localPCA, probabilisticPCA, little2012, recanatesi2022}, on fractal methods \cite{Grassberger1983, hein2005, fractal_camastra, fractal_id_binary_data, fractal_id_time_series, fractal_id_vector_quant}, or on nearest neighbor-based methods \cite{Facco2017, Levina2004a, idea2011, mindkl2011, danco2014, ess2015, geomle2019, iuri_discrete_id}.
A good review of the existing approaches and their properties can be found in \cite{Campadelli2015}. 
In particular, nearest-neighbor-estimators estimate the ID by exploiting the distribution of some geometrical quantity related to the nearest neighbors of each data point, such as distances between pairs of points or angles between pairs of vectors.
They provide an effective measure of the ID on curved and twisted manifolds, but due to their local nature, the estimate can be heavily affected by noise in the data.
Real-world samples are always perturbed by noise, and if the magnitude of the perturbation is comparable to the size of the neighborhoods, local estimators can largely overestimate the ID of the data. 
In this chapter, we propose an extension of the Two Nearest Neighbor estimator (TwoNN) introduced by \citet{Facco2017} to improve its reliability on noisy data sets.

\subsection{Related work on scale-dependent intrinsic dimension} 
A common strategy used by local estimators to make the ID estimate robust on noisy data sets is to increase the neighborhood size to find a range of scales where the ID is less affected by noise.

\citet{little2012} show that when the typical distance $r$ between a point and its nearest neighbors is above the noise scale but at the same time small enough not to be affected by the curvature, the spectrum $\lambda$ of the covariance matrix has a gap between the first $\hat{d}$ eigenvalues and the remaining $D-\hat{d}$. The ID is estimated by $\hat{d}$, applying local PCA on hyperspheres of increasing radius and looking at gaps in the plot $\lambda(r)$.
In a similar vein, \citet{recanatesi2022} look at the ratio $D_{PR}$ between the first and second moment of the spectral density of the covariance matrix $D_{PR}= \nicefrac{( \sum_j^D \lambda_j)^2}{\sum_j^D \lambda_j^2}$ on hyperspheres of increasing radii. The ID is identified at the plateau in $D_{PR}(r)$.
\citet{Facco2017} propose to estimate the ID by looking at a plateau of a local statistic, the ratio of the distances between a point and its second and first nearest neighbors, as a function of the scale (see Sec. \ref{sec:motivation}). 
\citet{hein2005} found that the optimal scale $h^*$ should depend on the data set size $N$ and on the assumed dimensionality $d$ and proposed to adapt the bandwidth $h$ of a correlation integral kernel to estimate the ID. 
An alternative method to identify an \enquote{optimal scale} \cite{idea2011}, is based on the empirical observation that the ID as a function of the data set size $N$ has a horizontal asymptote for $N \to \infty$.  \citet{idea2011} repeatedly measures the ID on random subsets of different sizes and estimates the asymptote.
In most cases, however, local estimators \cite{Levina2004a, mindkl2011, danco2014, ess2015, geomle2019} set the scale implicitly by fixing the number of $k$ nearest neighbors used in the estimate.

\section{Motivation}
\label{sec:motivation}
In Sec. \ref{sec:derivation_mu}, we will extend the TwoNN estimator \cite{Facco2017} to neighbors of higher order to improve the estimated ID on larger scales. 
In this section, we show with a simple example why using neighbors of higher orders typically leads to a more robust ID estimate than the decimation approach. 

\begin{figure*}
  \centering
  \includegraphics[width=\textwidth]{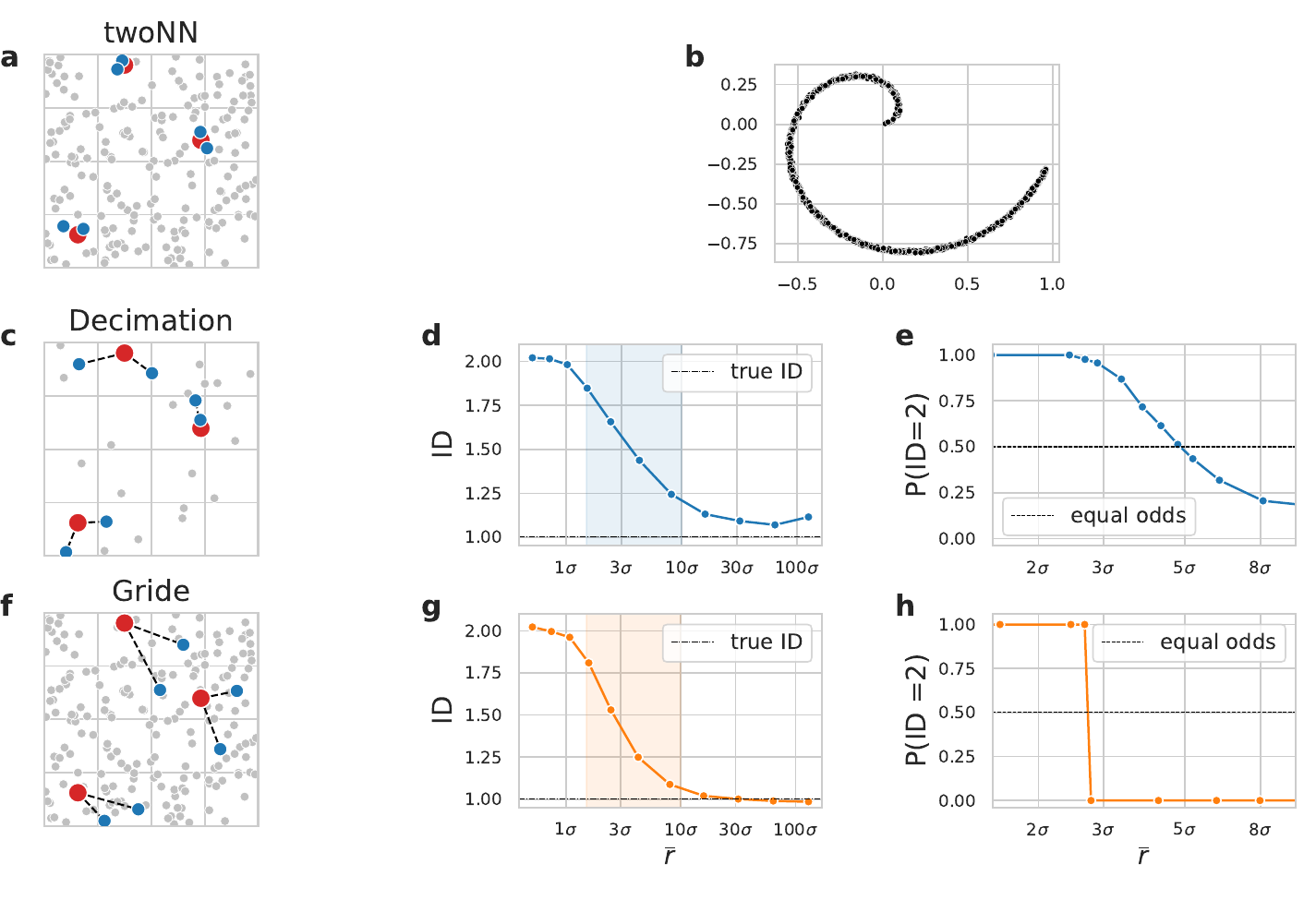}
  \caption{\label{fig:id_cartoon} 
  \textbf{A comparison between the scale analysis of TwoNN and Gride} 
  {\bf a}: Schematic representation of the TwoNN nearest neighbors for three data points sampled from a 2-dimensional uniform distribution; 
    {\bf b}: one dimensional noisy spiral. The spiral coordinates are parametrized by $t$ as $(t \cdot \text{cos}(6t); t \cdot \text{sin}(6t))$. We generate N = 20 thousand data points sampling $t$ uniformly on the unit interval and add to each coordinate of the points a Gaussian noise with standard deviation $\sigma = 10^{-3}$.
  {\bf c-d-e}: scale analysis of the ID with the decimation approach. {\bf c}:
  nearest neighbors on the decimated data set, {\bf d}: ID profile as a function of the neighborhood size, shown on the x-axis in $\sigma$ units, {\bf e}: probability ID = 2 as a function of the neighborhood size.
  {\bf f-g-h}: scale analysis of the ID with the Gride approach.
  }
\end{figure*}

\citet{Facco2017} show that the ID can be estimated using only the distances $r_{i_1}$, $r_{i_2}$ between each point $x_i$ and its first two nearest neighbors (Fig. \ref{fig:id_cartoon}-a).
Under the assumption of the Poisson point process in space, it can be shown that the ratios $\mu_i = \nicefrac{r_{i_2}}{r_{i_1}}$ between the distances to the second and first nearest neighbor follow a Pareto distribution:
\begin{equation}\label{eq:pareto}
p(\mu_i|d) = d\mu_i^{-d-1}
\end{equation}
Further assuming that the data-likelihood factorizes over the $N$ single-point likelihoods, the ID can be inferred given the empirical distribution of ${\boldsymbol \mu}$. 
In practice, due to its very local nature, the TwoNN estimator can be affected by noise, which typically leads to an overestimation of the ID. To identify the most plausible value of the data intrinsic dimension,
\citet{Facco2017} apply the TwoNN estimator on the full data set and then 
on random subsets of decreasing size: $[N, N/2, N/4, ...]$. 
The relevant data ID is chosen as the value of $\hat{d}$ where the graph $\hat{d}(N)$ exhibits a plateau and the ID is less dependent on the scale.

The first drawback of this approach is that to increase the scale, the ID must be estimated on a fraction of the available data points. 
A second issue is illustrated in Fig. \ref{fig:id_cartoon} where we have applied the "decimation" protocol on a one-dimensional noisy spiral embedded in $\mathbb{R}^2$ (see Fig. \ref{fig:id_cartoon}-b). 
We applied uncorrelated Gaussian noise with standard deviation $\sigma$ to each coordinate.
We estimate the ID on samples of decreasing size $[N, N/2, N/4, ..., N/512]$ and plot in Fig. \ref{fig:id_cartoon}-d the ID  as a function of the average neighbor distance $\overline{r} = \nicefrac{(\overline{r}_1 + \overline{r}_2)}{2}$,  shown on the $x$-axis in $\sigma$ units. 
As we decrease the subset size from left to right of the figure, the typical neighborhood size $\overline{r}$ increases. 
For $\overline{r} < \sigma$, the neighborhoods of each data point consist of samples from a two-dimensional Gaussian, and the ID is two. For $\overline{r} > 10\sigma$, the impact of the noise is small and the estimated ID is one. 
Between these two regimes, the geometry of the neighbor distances becomes more and more similar to that of a one-dimensional manifold as the neighborhood size increases and the importance of the noise reduces.
To quantify the scale at which the TwoNN model starts to assign more probability weight to ID = 1, we compute the TwoNN posterior probability $p(d| {\boldsymbol \mu} )$, assuming a uniform prior on $d$ and the likelihood of Eq. \eqref{eq:pareto}:
\begin{equation}
   p(d| {\boldsymbol \mu} ) = \dfrac{p({\boldsymbol\mu}|d )p(d)}{\sum p({\boldsymbol \mu}|d )p(d)}
\end{equation}
In Fig. \ref{fig:id_cartoon}-e we plot $p(2| {\boldsymbol \mu} )$, for $\overline{r} \in [1.5\sigma, 8.5 \sigma]$. 
The posterior probability for ID = 2 is larger than $p(1| {\boldsymbol \mu} )$ even when the average neighbor distance $\overline{r}$ is of the order of $4\sigma$.
This happens because of the presence of some data points whose second nearest neighbor distances are smaller than $\sigma$ and whose estimated local ID is close to 2.

Accordingly, reducing the fluctuations of $r$ around its mean is a promising strategy to observe more reliably the transition to an ID estimate equal to one. 
An intuitive way to reduce the variability of the neighborhood size is to consider neighbors of higher order instead of decimating the data set  \ref{fig:id_cartoon}-f.
This is because the sum of the nearest neighbors' distances has a relative fluctuation around its mean smaller than that of distances to the first nearest neighbors.
Figure \ref{fig:id_cartoon}-(g-h) shows that an estimator based on higher nearest neighbor ranks improves the ID estimates close to noise scale and that an ID = 1 is more likely already at $\overline{r} \sim 3\sigma$.

The main idea of this work is that the ID estimates near the noise scale can be improved using nearest neighbors of higher rank instead of decimating the data set. 
Our main contributions are as follows:
\begin{itemize}
\item we extend the TwoNN estimator approach to neighbors of higher order and derive a {\itshape Generalized ratios ID estimator} (Gride);
\item we benchmark Gride on synthetic real-world data sets showing that it improves the TwoNN ID estimates near the noise scale. We also show that Gride is competitive with other state-of-the-art estimators;
\item we show that Gride is faster than its competitors.
\end{itemize}

\section{Background}
\label{sec:background}
Before describing our algorithm, we briefly discuss the ideas behind the local estimators we will use for comparison in later sections. 
These estimators are the Maximum Likelihood Estimator (MLE, \citet{Levina2004a}), DANCo (\citet{danco2014}), ESS (\citet{ess2015}),  TwoNN (\citet{Facco2017}), GeoMLE (\citet{geomle2019}), and have been chosen to represent the current state-of-the art methods for ID estimation \cite{Campadelli2015}.
They assume that locally, in the range of the first neighbors of each point, the manifold is isomorphic to a hyperplane where the data are distributed according to a uniform distribution (DANCo, ESS) or a Poisson process in space (MLE, TwoNN, GeoMLE), and use some nearest neighbor statistics to infer the local intrinsic dimension of the data.

MLE derives the \emph{pointwise} ID estimate $\hat{d}(x_i)$ under the assumption of the Poisson point process in space. $\hat{d}(x_i)$ depends only on the distances $r_{i, j}$ between a query point $x_i$ and its $k$ nearest neighbors indexed by $j$:
\begin{equation}
\label{eq:levina}
    \hat{d}_k(x_i) = \left(\dfrac{1}{k}\sum_{j=1}^k\dfrac{r_{i, k+1}}{r_{i, j}}\right)^{-1}
\end{equation}
\citet{Levina2004a} aggregate the pointwise ID estimates with an average over the $N$ data points $\hat{d}_k = \nicefrac{\sum_{i=1}^N  \hat{d}_k(x_i)}{N}$. \citet{mackay2003} later showed that a correct likelihood-based combination of the $N$ individual estimates requires taking their harmonic average:
\begin{equation}
\label{eq:mackay}
    \hat{d}_{MLE_k} =\left( \dfrac{1}{N} \sum_{i=1}^N \dfrac{1}{d_k(x_i)}\right)^{-1}
\end{equation}
We will use Eq. \eqref{eq:mackay} in our experimental tests. 

Due to its simplicity and computational efficiency (see also Sec. \ref{sec:results_id}), MLE is one of the most used estimators
but tends to underestimate the ID on manifolds of high dimensionality when the density is not uniform \cite{Campadelli2015}. 
\citet{geomle2019} introduced a polynomial correction to the rate of the Poisson process, defining the model likelihood to take into account density variations. 
Their algorithm, named GeoMLE (Geometric Maximum Likelihood Estimator), estimates the ID as the intercept of a polynomial regression fit of $\hat{d}_{MLE_k}(r)$ where $r$ is the average distance to the $k^{th}$ neighbor. 
In practice, they take 20 random subsets of the data, compute $\hat{d}_{MLE_k}$ in each of them, and use the 20 pairs $\{ \hat{d}_{MLE_k}, \: \overline{r} \}$
to regress a quadratic polynomial and estimate the intercept.

A different approach to reduce the underestimation of MLE is to compare some nearest neighbor statistics estimated on $N$ data points, to the same statistic measured on a synthetic data set of \emph{known} ID where the number of points $N$ is the same. 
DANCo (Dimensionality from Angle and Norm Concentration) \cite{danco2014} similarly to MLE  computes a likelihood for the normalized nearest neighbor distances $\mathcal{L}_k(d, r)$ and adds a second likelihood to describe the distribution of the pairwise angles $\mathcal{L}(d, \theta)$ within the first $k$ neighbors of each point. 
The ID is computed by minimizing the sum of the Kullback-Leibler divergences for the distance and angle distributions between the data set and a synthetic sample of $N$ points uniformly drawn on a hyperball. 

\citet{ess2015} models the data with a uniform distribution in a hyperball in $\mathbb{R}^d$ to derive an analytical expression for the Expected Simplex Skewness (ESS) $s_d^n$. 
ESS is defined as the ratio between the volume of an $n$-dimensional simplex in $d$ dimensions constructed with a vertex in the centroid of the data and the remaining $n$ on random data points, and the volume the same simplex if the $n$ edges were orthogonal.   
This ratio is a monotonically increasing function of $d$. When $d \to \infty$ the pairwise angles tend to be orthogonal and the ratio approaches one.
In practice, \citet{ess2015} set $n=2$ and the simplex skewness is just a function of the sine of the angles between pairs of vectors.
The ID is estimated by comparing the empirical simplex skewness $\hat{s}^2$ with the analytical ones.

\section{Methods}
\subsection{Derivation of the probability distribution for \texorpdfstring{$\mu_k$}{u}}
\label{sec:derivation_mu}
We model a random distribution of data points in space with a homogeneous  Poisson process with rate $\rho$. In a homogeneous Poisson process (see  \citet{moltchanov2012}):

\begin{enumerate}
  \item the number of points contained in a region of space of volume $V$ is Poisson distributed;
  \item the numbers of points contained in two non-overlapping regions are independent random variables.
\end{enumerate}
The latter property characterizes the Poisson process, and motivates its adoption as a model of randomness in space \cite{Levina2004a, Facco2017, geomle2019} since, under the Poisson assumption, the inter-point distance distribution has tractable analytical expressions \cite{moltchanov2012}. 
We now show that the probability distribution $p(\mu_k)$ of the ratios of the distances between a point and its nearest neighbors of rank $k_1 \doteq k$ and $k_2 = 2 \cdot k$,
$\mu_k = \nicefrac{r_{2k}}{r_k}$ is:
\begin{equation} \label{eq::texttt{Gride}}
    p(\mu_k) = \dfrac{d(\mu_k^d-1)^{k-1}}{\mu_k^{(2k-1)d+1} B (k, k)}
\end{equation}
where $B (k, k)$ is the beta distribution.

\begin{figure}
  \centering
  \includegraphics[width=1.\textwidth]{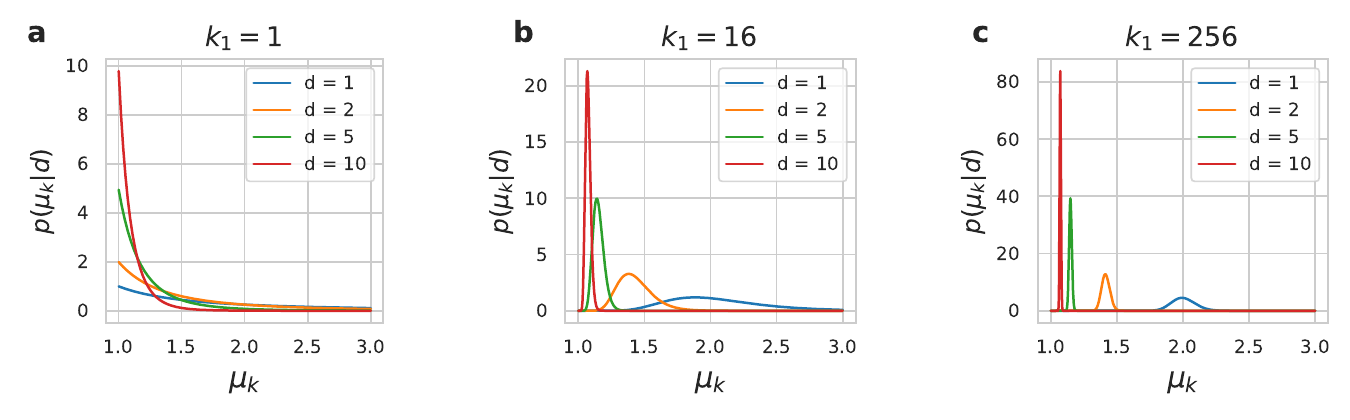}
  \caption{\label{fig:pdf} \textbf{Gride likelihood functions for different orders of $k$}. 
  We show the shape of the Gride likelihood functions for a $k_1=1$ ({\bf a}), $k_1=16$ ({\bf b}), and $k_1=256$ ({\bf c}). Inside each panel, we illustrate the distributions for increasing intrinsic dimensions equal to $(1, 2, 5, 10)$. When k=1 $p(\mu_k|d)$ coincides with the TwoNN Pareto likelihood (see Eq. \eqref{eq:pareto}). Increasing $k$ (panels b-c), the likelihoods of different IDs are less and less overlapped.
  }
\end{figure}

We prove that 1) $\mu_k$ is a function of the volumes of the spherical shells $v_i$ enclosed by successive nearest neighbors of a point, and 2) use the probability distribution of $v_i$ to derive the pdf of $\mu_k$
 \begin{proof}
1)
Consider a point of a data set in $\mathbb{R}^d$ and $r_i, r_2, r_3, ... r_k$ the sorted distances to its first $k$ nearest neighbors.
The volumes of the spherical shells between the $i^{th}$ and the $i+1^{th}$ neighbors of a point are $v_i = \omega_d (r_{i+1}^d - r_i^d)$ where $\omega_d = \pi^{d/2} / \Gamma(d/2 + 1)$ is the measure of the volume of the unit ball in $\mathbb{R}^d$. 
Consider now the following sums: $N = \sum_{k}^{2k-1} v_i$ and $D = \sum_1^k v_i$. These are telescopic sums in $r_i$ and equal $N = \omega_d (r_{2k}^{d}-r_k^{d})$ and $D = \omega_d r_{k-1}^{d}$ respectively. $\mu_k = \nicefrac{r_{2k}}{r_k}$ can be written in terms of $R = \nicefrac{N}{D}$ as:
\begin{equation}
    R = \mu_k^d-1
\end{equation}
2) If the data set is distributed according to a Poisson process, the joint distribution of $v_i$ is an exponential distribution with rate $\rho$ \cite{moltchanov2012, Facco2017}:
\begin{equation} \label{eq::gv}
    p(v_1, v_2, ..., v_{2k}) = \rho^{2k}e^{-\rho(v_1+v_2+...+v_{2k})}
\end{equation}
We use this result to derive $p(N, D)$, than $p(R)$ and finally $p(\mu_k)$.
The joint distribution $p(N, D)$ can be obtained with the following change of variables:
\begin{align*}
    & v'_1 = v_1, v'_2 = v_2, ... , v'_k = D \\
    & v'_{k+1} = v_{k+1}, v'_{k+2} = v_{k+2}, ... , v'_{2k} = N
\end{align*} 
This is a linear transformation with a Jacobian determinant ($\det J$) equal to one. By marginalizing over $v_i = \{v'_1, ... v'_{k-1},v_{k+1}, ..., v'_{2k-1} \}$ we get:
\begin{equation}
    p(N,D) = \dfrac{N^{k-1}}{(k-1)!}\dfrac{D^{k-1}}{(k-1)!}\rho^{2k}e^{- \rho (N+D)}
\end{equation}
Changing again the variables as $R=N/D$, $D'=\rho D$, with $\det J = \nicefrac{\rho}{D}$ we get:
\begin{equation}
    p(R, D') = \dfrac{R^{k-1}}{(k-1)!^2} D'^{2k-1} e^{-D'(R+1)}
\end{equation}
Marginalizing over $D'$ and recalling that  $\Gamma(k) = (k-1)!$  yields:
\begin{equation}
    p(R) = \dfrac{R^{k-1}}{\Gamma(k)^2} \int_0^\infty D'^{2k-1} e^{-D'(R+1)} \,dD' = \dfrac{R^{k-1}}{\Gamma(k)^2} \dfrac{\Gamma(2k)}{(R+1)^{2k}}
\end{equation}
The final transformation is $\mu_k^d =R+1$ ($\text{d} R = d \mu_k^{d-1} \text{d}\mu_k $) and recalling the relationship between the beta and gamma functions $\text{B} (k_1, k_2) = \nicefrac{ \Gamma(k_1) \Gamma(k_2) }{ \Gamma(k_1+k_2) }$, we get:
\begin{equation} 
    p(\mu_k) = \dfrac{d(\mu_k^d-1)^{k-1}}{\mu_k^{(2k-1)d+1} B (k, k)}
\end{equation}
\end{proof}
When $k=1$ Eq. \eqref{eq::texttt{Gride}} is the Pareto distribution $p(\mu_1) = d \mu_1^{-d-1}$ used in \citet{Facco2017} to derive the TwoNN estimator. 
As we increase the order $k$, for a given intrinsic dimension $d$, the variance of $p(\mu_k)$ decreases and the distribution concentrates around its mean (see Fig. \ref{fig:pdf}-(a-c)). 

\subsection{The \texttt{Gride} estimator}
Given a set of distance ratios ${ \boldsymbol \mu_k} = \{\mu_{k_i}: i = 1, ... N \} $ between each point ${x}_i$ and its $2k^{th}$ and $k^{th}$ we assume that the data likelihood $p({ \boldsymbol \mu_k})$ factorizes over the single-point likelihoods:
\begin{equation}\label{eq::likelihood}
    p({ \boldsymbol \mu_k}|d) = \prod_i^N p(\mu_{k_i}|d)
\end{equation}

By maximizing Eq. \eqref{eq::likelihood}, we get an intrinsic dimension estimator for $d$ that generalizes the TwoNN estimator to higher order ratios. For this reason, we call our ID estimator {\itshape Generalized Ratios Intrinsic Dimension Estimator}, in short, \texttt{Gride}.

Unfortunately, we can not give a closed form for the value of $d$ that maximizes the likelihood:
\begin{equation}\label{eq:score_function}
    \frac{\partial \mathcal{L}}{\partial d} = \dfrac{N}{d} + (k-1)\sum_{i=1}^N\dfrac{\mu^d_{k_i} \text{log} \mu_{k_i}}{\mu^d_{k_i}-1} - (2 k -1) \sum_{i=1}^N \text{log} \mu_{k_i}=0
\end{equation}
However, the solution of Eq. \ref{eq:score_function}
can be obtained numerically, as $\mathcal{L}$ is concave on the entire parameter space $d \in [0, +\infty)$:
\begin{equation}\label{eq:fisher_gride}
    \pdv[2]{\mathcal{L}}{d} = - \dfrac{N}{d^2} - (k-1)\sum_{i=1}^N\dfrac{\mu^d_{k_i} (\text{log} \mu_{k_i})^2}{(\mu^d_{k_i}-1)^2} < 0 
\end{equation}
and $\mathcal{L}$ has a unique maximum. 
Empirically, we found that a simple bisection rule with initial values $d_0 = 10^{-3}$, $d_1 = 10^3$ allows finding the root of Eq. \eqref{eq:score_function} in a fraction of a second (see Sec. \ref{sec:time_benchmark}).

The negative of Eq. \eqref{eq:fisher_gride}, evaluated in $\hat{d}$, is the so-called observed information $I(\hat{d}) = -\pdv[2]{\mathcal{L}}{d} $
and we use it to estimate the large-sample confidence intervals of the intrinsic dimension
\cite{wackerly2014, gelman2013}:
\begin{equation}\label{eq:confidence_gride}
  d = \hat{d} \pm \dfrac{1.96}{\sqrt{I(\hat{d})}}
\end{equation}

\subsection*{Reproducibility} The source code to reproduce the experiments of this section is available at the \texttt{GitHub} repository \href{https://github.com/diegodoimo/intrinsic_dimenson}{https://github.com/diegodoimo/intrinsic\_dimenson} 

\section{Results}
\label{sec:results_id}
\subsection{Simulated data sets}
\label{sec:simulated_data sets}
In this section, we evaluate the ID on multiple scales on several data sets of known ID taken from \cite{geomle2019}. 
These data sets are standard benchmarks used extensively \cite{hein2005, idea2011, mindkl2011, danco2014, Campadelli2015, geomle2019} to assess the performance of an estimator on manifolds linearly and non-linearly embedded in high-dimensional spaces. 
%
%

\begin{figure}
\centering
\includegraphics[width=0.37\columnwidth]{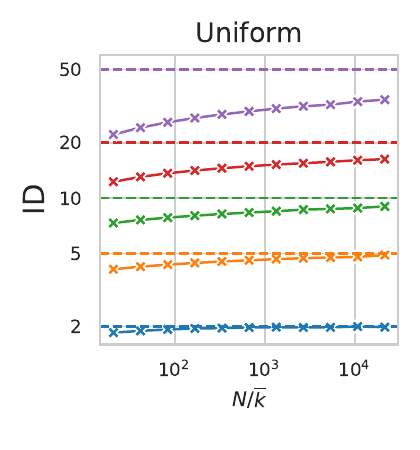}
\caption{{\bf Scale analysis of the ID done with Gride on uniformly distributed data sets.} Different profiles show the ID of uniformly distributed data sets of increasing dimension $d = (2, 5, 10, 20, 50)$. On the x-axis, $N$ is the total number of data points, and $\overline{k}$ is the average neighbor rank.$N/\overline{k}$ is plotted in a logarithmic scale. The ground truth value is plotted with horizontal dashed lines. 
}
\label{fig:syntetic_uniform}
\end{figure}

%
We first describe the scale analysis of the ID with Gride on $N_{tot} = 32000$ data points sampled from a uniform distribution on $d$-hypercubes where $d = \{2, 5, 10, 20, \\50\}$ (Fig. \ref{fig:syntetic_uniform}.
To increase the scale, we double each time the range $(k_1, k_2)$ of nearest neighbors' ranks: $(k_1, k_2) = \{(1, 2), (2, 4), (4, 8), ...\}$ and plot the estimated ID as a function of $N_{tot}/\overline{k}$ with $\overline{k}= \nicefrac{(k_1+k_2)}{2}$.
$N_{tot}/\overline{k}$ plays the role of the "scale parameter": the higher $N_{tot}/\overline{k}$, the smaller on average the size of the neighborhood.
%
%
%
The $x$-axis of Fig. \ref{fig:syntetic_uniform} is in logarithmic scale and to identify the ID, we look for plateaus in the $\{ \log(\nicefrac{N}{\overline{k}}), \: \hat{d}\}$  plot. 
When the ID is low,  say 2 or 5, $\hat{d}$ is almost constant over several orders of magnitude and equal to the true intrinsic dimension of the hypercube, displayed with dotted lines in the figure. 
As the dimension of the data grows, Gride starts to underestimate the ID and shows a decreasing trend as a function of the neighborhood size.
For instance, when the true dimension is $50$, the estimated ID at $(k_1, k_2) = \{ (1, 2),  (16, 32) \}$ are $\hat{d} = \{ 34, 30\}$ respectively, while when the true ID is $2$ the estimates are $\hat{d} = \{ 1.97, 1.98\}$.
This phenomenon is typical of most neighbor-based estimators \cite{Levina2004a, mindkl2011, Campadelli2015} and is related to the growing amount of data points lying close to the boundary of the manifold \cite{danco2014}, which makes the constant density assumption of the Poisson process less realistic when the ID is high.

We now compare Gride with the estimators described in Sec. \ref{sec:background} using the official implementations provided by the authors at \href{https://github.com/stat-ml/GeoMLE}{github.com/stat-ml/GeoMLE} for GeoMLE, 
\href{https://it.mathworks.com/matlabcentral/fileexchang/40112-intrinsic-dimensionality-estimation-techniques}{it.mathworks.com/matlabcentral/fileexchange/40112-intrinsic-dimensionality-estimation-techniques} for DANCo, 
\href{https://github.com/kjohnsson/intrinsicDimension}{github.com/kjohnsson/intrinsicDimension} for ESS.

In all cases, the neighborhood range is defined by choosing the value of $k$-nearest neighbors considered in the estimate. With the only exception of the TwoNN estimator, where, by definition, $(k_1, k_2) = (1, 2)$, $k$ is a hyperparameter chosen empirically by the methods. 
The default values are $k<10$  for DANCo \cite{danco2014}, ESS and MLE ($\overline{k} = 5.5$), and  $20<k<55$ for GeoMLE ($\overline{k} = 38.5$). 
To analyze the ID on different scales, we repeatedly evaluate the IDs on bootstrap subsets of sizes $N = \{N_{tot}, N_{tot}/2,N_{tot}/4,... \}$ following the protocol of \citet{Facco2017}.
Since GeoMLE measures the ID on random bootstrap samples, the average number of data points is $0.632\cdot N_{tot}$ (see Sec. \ref{sec:background}).
The scale parameter $\nicefrac{N}{\overline{k}}$ allows comparing on consistent scales the IDs measured on data sets with a different number of samples and values of $\overline{k}$. Indeed, the neighborhood size grows either by making $\overline{k}$ bigger (Gride), or by decimating the data set when $\overline{k}$ is fixed.
\begin{figure}
  \centering
  \includegraphics[width=1.\textwidth]{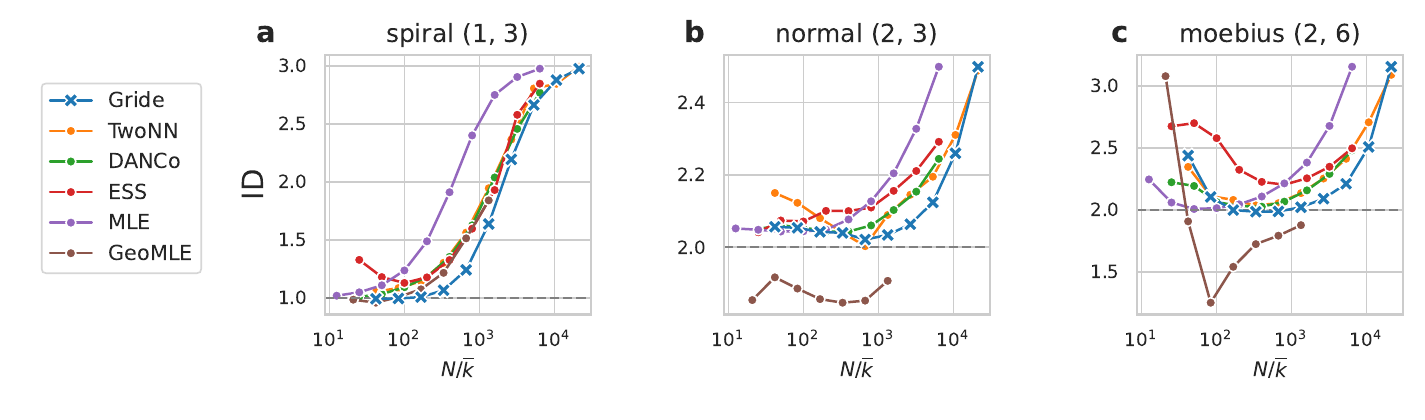}
  \caption{\label{fig:syntetic_datasets} \textbf{Scaling analysis of the ID on simulated data sets with noise.} We estimate Gride, TwoNN, and the other four nearest neighbor-based estimators described in Sec. \ref{sec:background}.
  The data sets are:
  {\bf a}: one dimensional spiral embedded in $\mathbb{R}^3$;
  {\bf b}: bivariate gaussian embedded in $\mathbb{R}^3$;
  {\bf c}: Moebius strip embedded in $\mathbb{R}^6$.
  Each data coordinate is perturbed with random Gaussian noise with a standard deviation equal to $0.01/\sqrt{D}$.
  }
\end{figure}

Figure \ref{fig:syntetic_datasets} shows the scale analysis of the ID on $N_{tot} = 16000$ points sampled from a 1-$d$ spiral, a  2-$d$ standard normal distribution, and a 2-$d$ Moebius strip.
The embedding dimensions $D$ are 3 ({\bf a}-{\bf b}) and 6 ({\bf c}). Gaussian noise of standard deviation $\sigma = \nicefrac{0.01}{\sqrt{D}}$ is independently added to each coordinate of the manifolds.
At large $N/\overline{k}$, the typical distance between a point and its $k$-nearest neighbors is smaller then the noise standard deviation $\sigma$, and all the methods
overestimate the ID of the underlying manifold. 
Increasing the neighborhood size, when the typical distance becomes larger than $\sigma$ the estimated IDs approach the true dimension of the manifolds.
However, the rate and the scale at which the ID reaches the ground truth value are different for each estimator, and, in all cases, Gride stabilizes around $d$ at smaller scales than the competitors.

More precisely, for the 1$d$ spiral ({\bf a}), the noise standard deviation $\sigma \sim 6.0 \cdot 10^{-3}$ and is similar to the average distance to the first two nearest neighbors and $N/\overline{k} \sim 10^4$ corresponds to distances of the order of  $3\sigma$.
TwoNN and Gride are the only estimators able to evaluate the ID at very local scales $N/\overline{k} >10^4$ where the ID is almost 3,  equal to the embedding dimension. 
DANCo, ESS, and MLE reach $N/\overline{k} \sim 10^4$ where the ID is close to 3. The GeoMLE profile instead starts from $N/\overline{k} \sim 2 \cdot 10^3$, a scale at which the ID is around 1.9.
In the range  $10^2 < N/\overline{k} < 10^3$, the Gride profile decreases faster than those of the other estimators and reaches 1.006 around $N/\overline{k} \sim 5 \cdot 10^2$. 
At the same scale, the IDs measured by TwoNN, DANCo, ESS, MLE, and GeoMLE are  $[1.3, 1.32, 1.4, 1.48, 1.35]$ respectively.
Increasing further the neighborhood size ($N/\overline{k} < 5 \cdot 10^2$), Gride remains constant and equal to one and the other estimators gradually approach the same value with the exception of ESS which shows a small increase for $N/\overline{k} \sim 10^1$. In this range, $\hat{d}$ equals the true ID of the manifold. 

\begin{figure}[t]
  \centering
  \includegraphics[width=1.\textwidth]{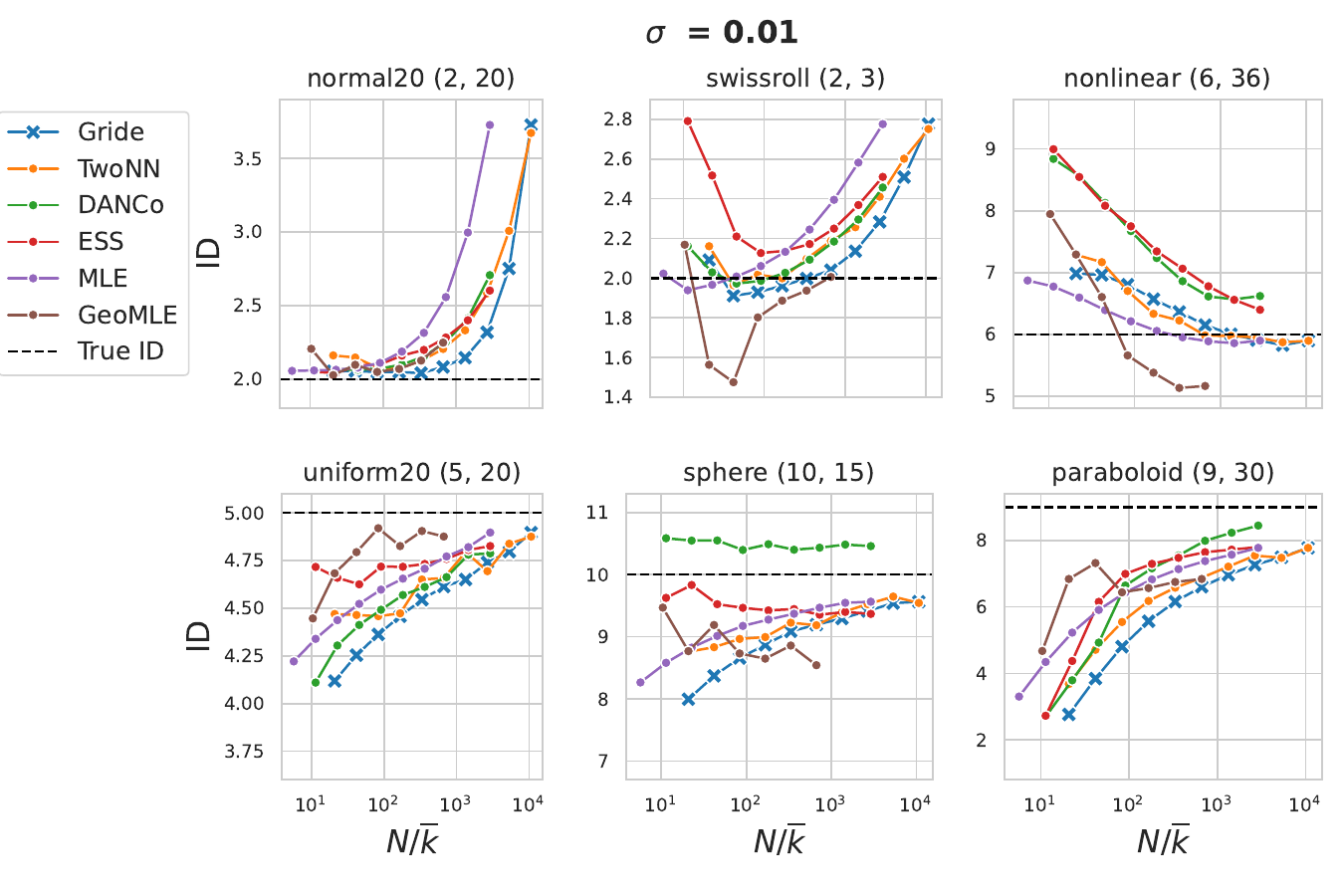}
  \caption{\label{fig:syntetic_datasets_b} \textbf{Other simulated data sets.} 
  As in Fig. \ref{fig:syntetic_datasets}, we show the ID estimates with Gride, TwoNN, and the other four nearest neighbor-based estimators described in Sec. \ref{sec:background}.
  The data sets are:
  {\bf a}: bivariate gaussian embedded in $\mathbb{R}^{20}$;
{\bf b}: swiss roll embedded in $\mathbb{R}^3$;
  {\bf c}: highly non-linear data set embedded in $\mathbb{R}^{36}$ (see \cite{Campadelli2015});
  {\bf d}: uniform 5-dimensional distribution embedded in $\mathbb{R}^{20}$;
  {\bf e}: 10-dimensional sphere embedded in $\mathbb{R}^{15}$;
    {\bf f}: 9-dimensional paraboloid  embedded in$\mathbb{R}^{30}$.
  Each data coordinate is perturbed with random Gaussian noise with a standard deviation equal to $0.01/\sqrt{D}$. 
  }
\end{figure}

The profiles of the IDs on the noisy normal distribution ({\bf b}) and on the Moebius strip ({\bf c}) show similar trends: the Gride estimator exits at smaller neighborhood sizes and on average the plateau is larger and more stable than the competitors;
Figure \ref{fig:syntetic_datasets_b} 
TwoNN in addition to reaching the ground truth ID at larger scales than Gride, in the case of the normal distribution ({\bf b}) does not show a plateau at all;
DANCo measures very stable ID profiles over a wide range of scales; 
ESS tends to overestimate the IDs with plateaus that are on average shorter than those of Gride, TwoNN, DANCo, and MLE;
\begin{figure}
  \centering
  \includegraphics[width=0.93\textwidth]{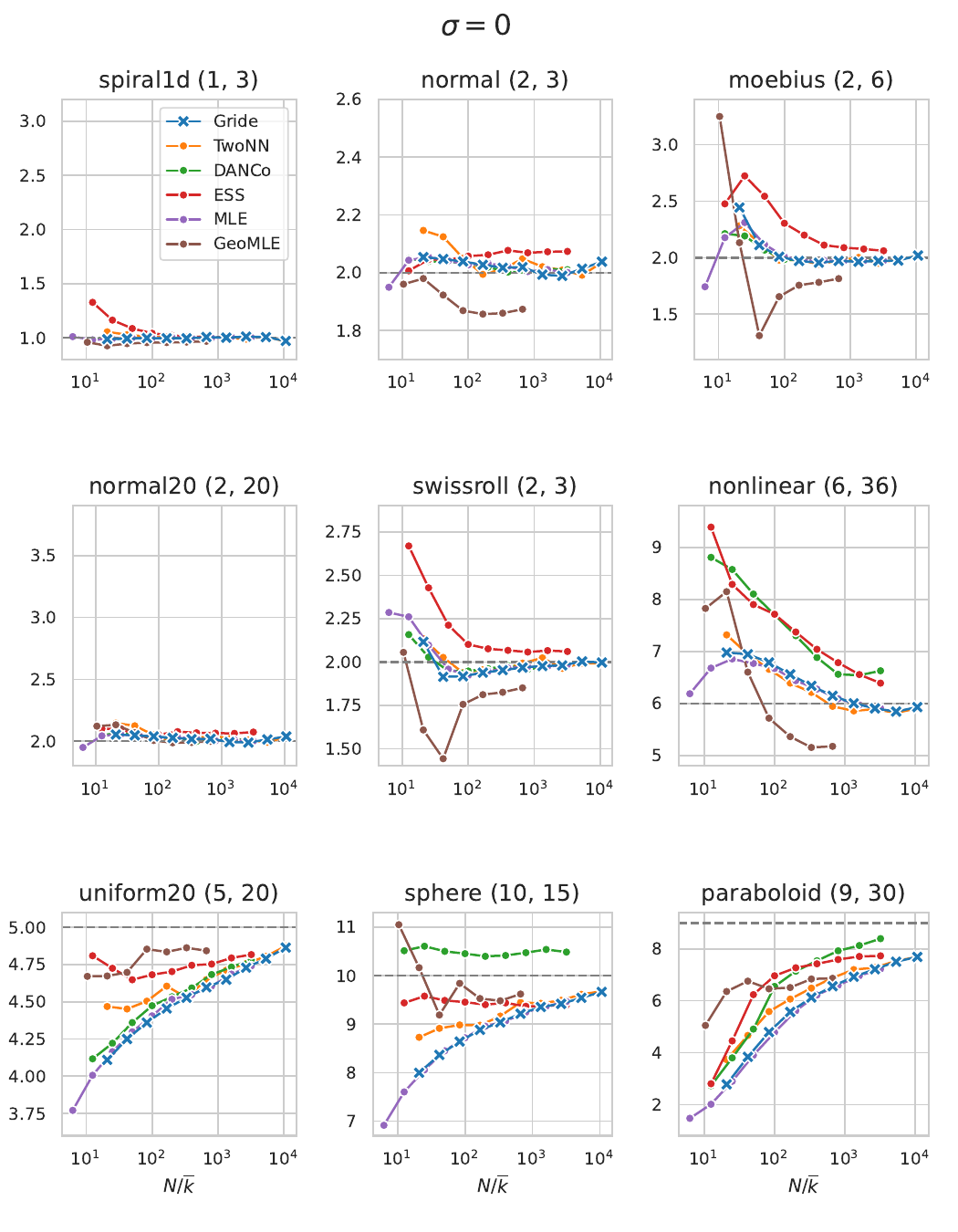}
  \caption{\label{fig:syntetic_noiseless}
  \textbf{Simulated data sets without additional noise.} 
  The figure shows the ID as a function of different neighborhood sizes on ten simulated data sets without noise. The manifold where the points lie is written on the top of each panel, and the analytic equations are reported in Fig. \ref{fig:syntetic_datasets} and Fig. \ref{fig:syntetic_datasets_b}.
  We compare six estimators as reported in the legend of the first panel.  We plot the neighborhood size on the x-axis in logarithmic scale as a function of $\nicefrac{N}{\overline{k}}$: $N$ is the data set size and $\overline{k}$ is the neighborhood range.
  }
\end{figure}
MLE is very similar to TwoNN and Gride. It is more robust than TwoNN because it considers a larger set of points to infer the ID (e.g $k = 10$, see \cite{Levina2004a}), but its estimates are less local and larger than those of TwoNN and Gride for a given scale;
GeoMLE is most unstable likely due to the large neighborhood size used to fit the ID.
shows the analysis of six other types of manifolds. For comparison, Fig. \ref{fig:syntetic_noiseless} shows the same results without additional noise. 
\subsection{Real data sets}\label{sec:real_datasets}
We now turn to the analysis of the ID on three widely used  real benchmarks  \cite{hein2005, danco2014, geomle2019}: 
the \href{http://yann.lecun.com/exdb/mnist/}{MNIST} \cite{mnist} digit number one, a collection of $N_{tot} = 6742$ gray scale images with 784 pixels, 
the \href{https://web.archive.org/web/20160913051505/http://isomap.stanford.edu/data sets.html}{ISOMAP faces} \cite{isomap} , which consist of $N_{tot} = 698$ gray scale images with 4096 pixels, 
and the \href{https://archive.ics.uci.edu/ml/data sets/isolet}{ISOLET} (Isolated Letter Speech Recognition) data set \cite{isolet}, generated by recording the sound alphabet letters spoken by 150 different subjects $N_{tot} = 7797$ samples with $617$ features each. 
The data sets are generated by processes (the recorded voice that spoke a letter, or the scanned handwritten digit) affected by intrinsic noise that can change the intrinsic dimension estimated at different scales. 
Indeed, the ground truths for these data sets are not known but there is a consensus around plausible ID intervals: between 8 and 11 for MNIST, 3 for the ISOMAP faces (determined by the presence of two main axes of variation for the face pose, vertical and horizontal, and one axis for the light direction) and between 16 and 22 for the ISOLET letters.

The ID is estimated following the procedure described in Sec \ref{sec:simulated_data sets}: for TwoNN, DANCo, ESS, GeoMLE, and MLE we decimate the data sets and compute the ID on subsets with $N= \{N_{tot}, N_{tot}/2, N_{tot}/4, ...\}$; on Gride we perform the scale analysis on the original data set doubling at each step the neighbor ranks. 
%
We reduce the GeoMLE nearest neighbor range to $(k_1, k_2) = (5, 15)$ because we find that the algorithm diverges for the ISOMAP data set using the default values $(k_1, k_2) = (20, 55)$.

\begin{figure}
  \centering
  \includegraphics[width=1.\textwidth]{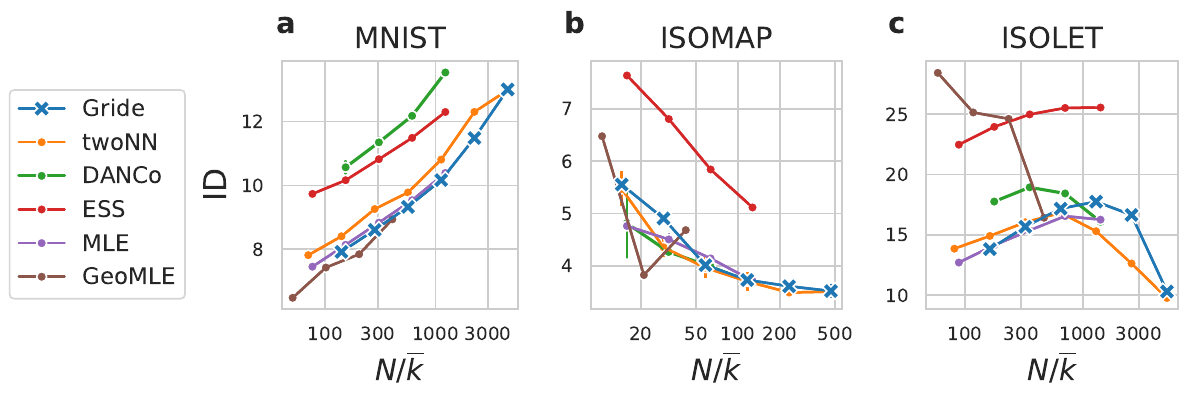}
  \caption{\label{fig:real_datasets} \textbf{ID profiles of the MNIST digit one, ISOMAP faces, and ISOLET data sets.} Figure shows the ID as a function of different neighborhood sizes on three real-world data sets: the MNIST digit one ({\bf a}), the ISOMAP faces ({\bf b}), and ISOLET letters ({\bf c}). We compare six estimators as reported in the legend.  We plot the neighborhood size on the x-axis in logarithmic scale as a function of $\nicefrac{N}{\overline{k}}$: $N$ is the data set size and $\overline{k}$ is the neighborhood range.}
\end{figure}
On MNIST digit one (Fig. \ref{fig:real_datasets}-a), the ID increases with $N/\overline{k}$
from 8 to 13, for all the estimators. Gride, TwoNN, MLE, and GeoMLE are consistent with each other in the range $100 < N/\overline{k} < 700$, while ESS and DANCo estimate a slightly larger ID. 
This example shows the importance to compare the IDs on the same "scale" given by $N/\overline{k}$. For instance, at $N/\overline{k} \sim 700$, Gride, TwoNN, MLE, and GeoMLE measure the ID on data sets sizes equal to  6742, 1685, 3371, and $6742 \cdot 0.632 \sim 4260$ respectively, but the estimated ID is similar for all of them. 

The ISOMAP data set  (Fig. \ref{fig:real_datasets}-b) is challenging because it has only 698 images, an order of magnitude less than MNIST and ISOLET.
At small scales ($N/\overline{k} >100$), the profiles of Gride and TwoNN show that the ID is approximately constant and equal to 3.5.
At larger neighborhood sizes $30< N/\overline{k} <100$, the ID starts to grow and the estimates of Gride, TwoNN, MLE, and DANCo are comparable ($ID \sim 4$) while the ID measured by ESS is around 5.8. 
GeoMLE was particularly hard to fit to this data, probably due to the high neighbor ranks ($k_1 = 5$, $k_2 = 15$). Its estimates oscillate between 3.9 and 6.5 in the range $N/\overline{k} <30$. 

On ISOLET (Fig. \ref{fig:real_datasets}-c), the IDs estimated by Gride, TwoNN and DANCo are not monotonic functions of $N/\overline{k}$. Near the maxima, $\hat{d}$ is almost constant and equal to $17.2 \pm 0.5$ for Gride ($N/\overline{k} \in [650, 2500]$), $16.0 \pm 0.7$ for TwoNN ($N/\overline{k} \in [300, 1300]$) and $16.0 \pm 0.7$ for DANCo $N/\overline{k} \in [170, 700]$.
The IDs estimated by ESS and MLE slowly increase with $N/\overline{k}$ to become almost constant around their maximum values reached at the smallest scales, where $\hat{d} \sim 25$ and 16, respectively.
GeoMLE is again more unstable than the other estimators, and $\hat{d} $ varies between 16 and 28.

\subsection{Computational cost of  \texttt{Gride} vs other ID estimators}\label{sec:time_benchmark}

In sections \ref{sec:simulated_data sets} and \ref{sec:real_datasets}, we have seen that the scale analysis done with Gride is competitive with the other estimators. 
When a plateau can be identified in the plot $\{ \hat{d}$, : $\log ( N/\overline{k})\}$, 
Gride often predicts a wider range of values over which the ID can be considered constant, making the choice of the ID easier.
Another crucial practical advantage of GRIDE is that the ID computation is much faster than that of the other methods analyzed in this chapter.

To compare the speed of the estimators, we measure how the execution time grows as a function of the number of samples $N$ and of the number of input coordinates $P$
using the CIFAR10 data set \cite{cifar10} which consists of 50000 color images of size $3 \times 32 \times 32$. 
We vary $N$ taking random subsets of size $\{N, N/2, N/4, ... N/64\}$, and vary $P$ by resizing the resolution of 5000 images (those of the cats) from $3\times 16 \times 16$ ($P = 768$) to $3\times 181\times181$ ($P = 98283)$. Each estimator is evaluated once on each data size.
The hyperparameters are the same as those used in Sec. \ref{sec:simulated_data sets} and \ref{sec:real_datasets}: $k\leq10$ for DANCo, ESS and MLE, $5\leq k \leq15$ for GeoMLE. For TwoNN by definition $k\leq2$, and for Gride we consider $k\leq64$.
The computation is done on an Intel Xeon E5-2690v3 processor with 12 cores and 24 threads.

\begin{figure*}
  \centering
  \includegraphics[width=1.\textwidth]{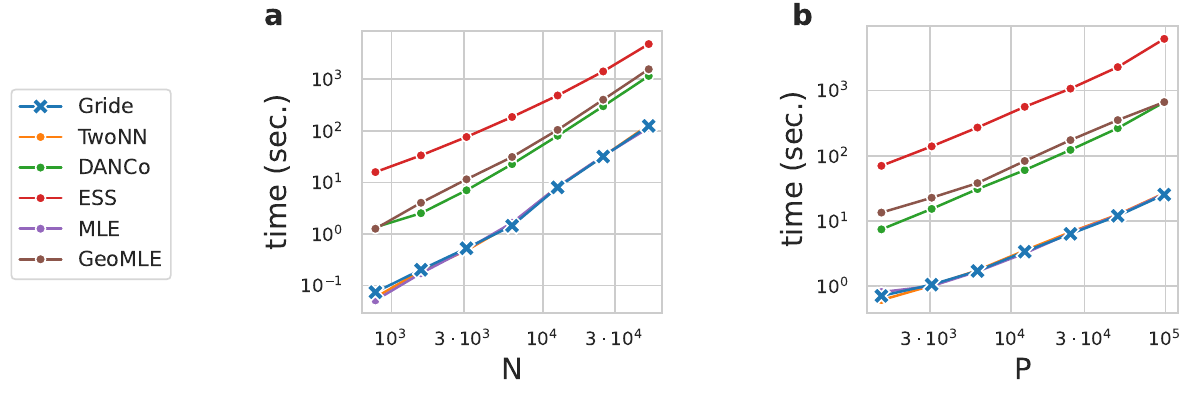}
  \caption{\label{fig:time_benchmarks} \textbf{Time benchmarks for different estimators} We compare the computational time of the estimators reported in the legend as a function of the data set size $N$ {\bf a}:, and the number of features $P$ {\bf b}: of each data point.
  The data set is CIFAR10. We vary $N$ choosing a random subset of increasing size, keeping $P = 1024$. We vary $P$ by increasing the resolution of $N = 5000$ images.
  }
\end{figure*}

In Fig. \ref{fig:time_benchmarks} we compare the increase in execution time with the number of samples $N$ (left) and with the number of features $P$ of each sample (right). 
In both cases, Gride, TwoNN, and MLE are always one order of magnitude faster than DANCo and GeoMLE, and almost two orders of magnitude faster than ESS.
When $N = 50000$ (a) the ID evaluation takes 2 minutes for Gride, TwoNN and MLE, 19 and 26 minutes for DANCo and GeoMLE (x10/15 speedup), 1 hour and 20 minutes for ESS (x40 speedup). 
The comparison is even more favorable when $P$ is large (Fig. \ref{fig:time_benchmarks}-b): when $P \sim 10^4$, the execution time is 3.3 seconds for Gride TwoNN and MLE, 60 and 80 seconds for DANCo and GeoMLE (x20/25 speedup), almost 10 minutes for ESS (x160 speedup).

Gride, TwoNN, and MLE are much faster than the other estimators because they just need a single evaluation of the full distance matrix $O(N^2P)$ plus $O(N(k_2\log{k_2}+N))$ operations to find the first $k_2$ neighbors for each data. 
The cost required to maximize the Gride likelihood with a bisection approach (see Eq. \eqref{eq:score_function}) grows linearly in $N$ and is 0.28 seconds for $N = 5 \cdot 10^4$ data points, a negligible amount of time with respect to the calculation of the distances.
On small data sets ($N \le 10^4, P\le 10^3$), Gride is slower than a single evaluation of the TwoNN estimator, due to the time required to sort the distances to higher ranges (we typically set $k_2=64$ or $k_2 = 128$ in practical applications). 
On large data sets ($N \ge 10^4, P\ge 10^4$), the computational time is dominated by the distance matrix evaluation and is the same for Gride, TwoNN, and MLE when the comparison is made on a \emph{single} ID measure.
In Sec. \ref{sec:id_resnet152}, we will see that a full-scale analysis with Gride requires half of the time of TwoNN for large $P$ values but in this section, due to the computational cost of ESS DANCo and GeoMLE, we limited the comparison just on a single ID estimate.
GeoMLE is approximately 20 times slower because it requires applying 20 times the MLE estimator on random subsets before fitting the ID (see Sec. \ref{sec:background}). 
The bottleneck of ESS and DANCo is instead related to the additional operations to compute the $\binom{k}{2}$ pairwise angles for each data point.

The high computational cost is also the reason why we chose not to increase the neighborhood size by adjusting the nearest neighbor order $k$ in DANCo, ESS.
In ESS, increasing $k$ from 10 to 30,  the computational time becomes three times larger than what is shown in Fig. \ref{fig:time_benchmarks}.
In MLE (and GeoMLE), from a computational time perspective, it is feasible to increase $k$ to evaluate the ID on multiple scales. 
However, as we saw in the previous section, GeoMLE is very unstable for high $k$ values, especially when the number of data is small.
MLE has a memory overhead with respect to Gride since all the neighbors up to $k$ are needed to compute the ID (see Eq. (\ref{eq:levina}, \ref{eq:mackay})) and the memory scales as $N \cdot k$. In Gride the memory grows as $N \cdot \log_2 k$, because only the neighbors' ranks of an integer power of two are stored. 

\section{Discussion}
In this chapter, we described Gride, an extension of the TwoNN estimator to efficiently estimate the ID on multiple scales.
Gride is orders of magnitude faster than other nearest neighbor-based estimators (DANCo, GeoMLE, ESS), does not depend on any hyperparameter because it evaluates the IDs on multiple scales, and does not require to decimate the data set like TwoNN. 
On data sets with noise, Gride allows exiting the "noisy region" at shorter length scales, making the ID easier to identify.

\begin{figure*}
  \centering
  \includegraphics[width=1.\textwidth]{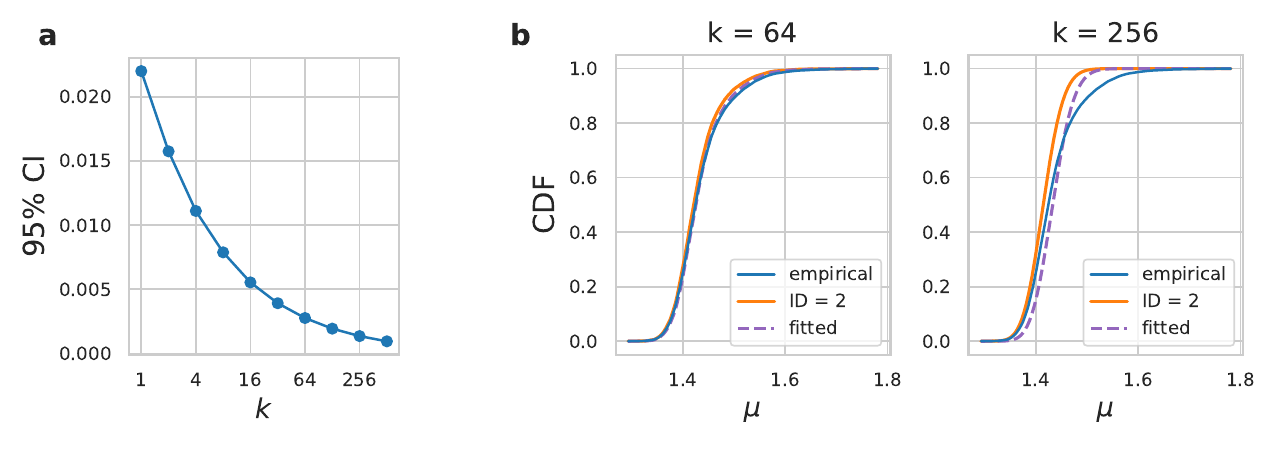}
  \caption{\label{fig:id_cumulative} \textbf{Confidence intervals of the Gride estimator.} In panel {\bf a} we plot the 95\% large-sample confidence intervals (CI) of the Gride estimator for an increasing value of $k_1$. As $k_1$ grows, the (CI) becomes small as a result of the decrease of the variance (see Fig. \ref{fig:pdf}). 
  Panel {\bf b} shows the cumulative distribution functions (CDF) of the Gride likelihood. 
  }
\end{figure*}

The estimator is efficient and reliable, but some issues related to its properties are still open.
The first issue is related to the error estimated for large nearest-neighbor ranks. 
The variance of the Gride likelihood as a function of $d$ is a decreasing function of $k$. As a result, by increasing $k$, the observed information grows and the estimated confidence intervals become small (see Eq. \eqref{eq:confidence_gride}).
Figure \ref{fig:id_cumulative}-a shows the width of 95\% confidence intervals estimated on the 2-dimensional uniform distribution analyzed in Fig. \ref{fig:syntetic_datasets}-a as a function of $k$.
When $k= 256$, the estimated ID with its confidence is $1.930 \pm 0.001$.
However, when the size of the neighborhood is large, the Poisson process assumption used to derive Gride is violated more often, mainly due to the finite size of the data set. This is an unfortunate case where the model becomes more confident when the prediction is wrong.

To overcome this difficulty one should always perform an additional model assessment step. A visual inspection of the deviation of the empirical cumulative distribution (CDF) of the Gride likelihood (Eq. \eqref{eq::likelihood}) from the analytical one, gives a first, qualitative hint about the consistency between the model and the data. 
In  Fig. \ref{fig:id_cumulative}-b, we compare the empirical CDF (blue) with the analytical one based on the ground truth (ID = 2, orange) and the estimated ID for $k = 64$ and $k = 256$. 
As expected, the deviation from the ground truth is larger for higher values of $k$ and, in this setting, concerns the tail where the $\mu_k$ are larger which is more populated in the empirical CDF.
In general, the higher the ID, the smaller the $\mu_k = \nicefrac{r_{2k}}{r_k}$ because in high dimension it is more likely for nearest neighbors to occupy regions of space at a similar distance from a given point (see also Fig. \ref{fig:pdf}). 
The presence of boundaries limits portions of space that could be occupied by data points, making the higher tail of the likelihood fatter and the estimated ID smaller.
Unfortunately, it is hard to make quantitative statements about the goodness of the fit in the general case, but the inspection can nonetheless rule out cases where the fit is manifestly wrong.

A second limitation of Gride is the assumption of constant ID for all the data points. This is a common hypothesis for most ID estimators \cite{Campadelli2015}, but \citet{Allegra2020} showed that in real-world data sets the ID is in general not constant, and being able to identify the homogeneous ID components can add valuable information about the data.
To segment the data according to the local ID, \citet{Allegra2020} proposed Hidalgo, a Bayesian mixture model based on the TwoNN Pareto likelihood Eq. \eqref{eq:pareto}. 
A difficulty of Hidalgo is related to the large overlap of the Pareto likelihoods (see Fig. \ref{fig:pdf}-a), which makes a single $\mu_i$ compatible with models of different IDs.
The Gride likelihoods at relatively low neighbor ranks (e.g. $k_1 = 4$, $k_2 = 8$) are less overlapped and could be used in place of the Pareto ones to improve the discriminative power of Hidalgo.
%
A promising future direction of research would be to combine Hidalgo and Gride for a multiscale ID estimation where the intrinsic dimension is not constant in the data set.

\section{Intrinsic dimension of hidden representations in deep neural networks}\label{sec:imagenet_igpt}
Using the tools described in the previous section, we study the ID of the ImageNet representations in a ResNet152, a standard architecture for image classification, and in Image GPT model (iGPT, \cite{igpt}) an attention-based model trained to generate images. 
In many models for image classification, the ID grows in the first layers and then decreases, reaching its lowest value in the output layer \cite{ansuini2019intrinsic}.
In contrast, we will see that in the large iGPT architecture the profile of the ID has a "double hunchback" shape, and that the two peaks arise in different stages during the training dynamics.

\subsection{Intrinsic dimension of the ResNet152 representations.}\label{sec:id_resnet152}
\citet{ansuini2019intrinsic} analyzed the ID of the hidden representations of convolutional networks (AlexNet, VGGs, ResNets). The networks were trained on ImageNet and the ID was measured with the TwoNN estimator on a subset of 3500 images from the training set. 
In this paragraph we compare Gride with the TwoNN estimator on the hidden representations of a ResNet152, extending the analysis of \citet{ansuini2019intrinsic} to $N =  90000$ images, and study the evolution of the ID during training. 

\begin{figure*}[t]
  \centering
  \includegraphics[width=0.95\textwidth]
  {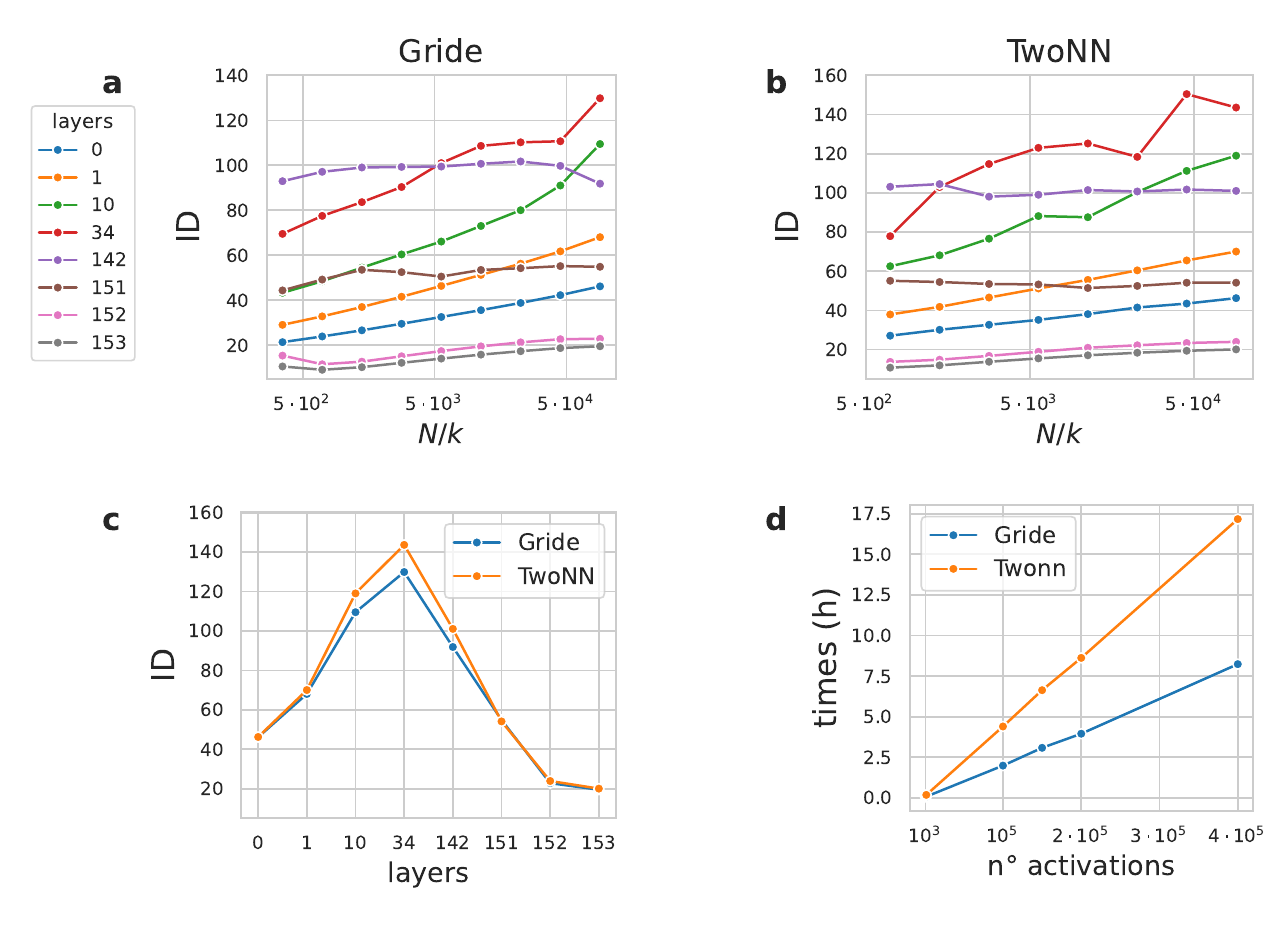}
  \caption{\label{fig:id_resnet152} \textbf{ID evaluated with Gride and TwoNN on ResNet152 hidden representations.} {\bf a} Scaling analysis of the ID done with Gride on eight hidden representations of ResNet152. $N/k$ is plotted in logarithmic scale on the x-axis; $N = N_{tot}$, $k = (k_1+k_2)/2$.
  {\bf b}: scale analysis done with TwoNN on the same representations ($N = N_{tot}/2^i$, $k = 1.5$, see main text)
   {\bf c}: ID vs hidden representation estimated with Gride ($k_1 = 2$, $k_2=4$) and TwoNN ($N = N_{tot}/2$).
  {\bf d}: Computational time (hours) required to estimate the ID of the ResNet152 representations with Gride (blue) and TwoNN (orange). The number of activations in a representation is plotted on the x-axis. The scale analysis with Gride is approximately two times faster than TwoNN.
  }
\end{figure*}

The ResNet152 architecture has four blocks. In each block, the number of activation of each channel $H^2$ and the number of channels $C$ are kept constant. Between two consecutive blocks, the side of the channels $H$ is halved and the number of channels is doubled. We measure the ID at the output of the ResNet blocks, corresponding to layers = $(10, 34, 142, 151)$, after the input stem (layer 1), and after the last average pooling  (layers = 152). 
We use the standard input resolution $224 \time 224$, so that the size of a representation ($D = H^2 \times  C$)  are
$D = (150528, \; 200704, \; 401408, \;200704, \;100352, \\ \;2048, \;1000)$ for layer $(0, \, 1, \, 10, \, 34, \, 142, \, 151, \, 152, \, 153)$ respectively, where $"0"$ stands for the raw input and $"153"$ for the logits. In our numbering, there are 153 layers because we increment the counter also for the last average pool, between representations $"151"$ and $"152"$, where there are no convolutions.
We sample the 90000 images from the training set keeping 300 classes and 300 images per class. 
We measure the IDs with Gride using $(k_1, k_2) = \{(1, 2), (2, 4), ..., (256, 512)\}$ and with TwoNN we decimate the data set size in powers of two up to $N = N_{tot}/256$.
The scale analysis shown in  Fig. \ref{fig:id_resnet152} reveals that the ID measured by Gride and TwoNN are quantitatively similar. %
In both cases, the ID decreases as the neighborhood size grows, more markedly for high ID values.
%
At representation "34", when the ID reaches its peak, the scale analysis done with the TwoNN estimator is more unstable than that done with Gride. 
In panel \ref{fig:id_resnet152}-c we summarize the profile of ID across all the hidden representations, choosing for simplicity, for all the representations, the values measured at the same scale: $k_1 = 2, k_2 = 4$ for Gride, $N = \nicefrac{N_{tot}}{2}$ for TwoNN. 

A sizeable difference between Gride and TwoNN concerns computational time.
Panel \ref{fig:id_resnet152}-d shows that Gride is two times faster than TwoNN when the number of activations $P$ is greater than $10^5$. This is a regime where the elapsed time is dominated by the computation of the distance matrix $O(N^2 P)$. 
Indeed, when the data set size is reduced by a factor of $2$, TwoNN computes the ID twice and averages the bootstrap estimates.
This operation requires half the computational time $T_{N_{tot}}$ to compute the ID on the original data set because $T_{N_{tot}} \sim N_{tot}^2 P$ while 
$T_{N_{tot}/2} \sim 2 \cdot {(\nicefrac{N_{tot}}{2})} ^2 P \sim N_{tot}^2 P/2$. 
Repeating this protocol up to $N = \nicefrac{N_{tot}}{256}$, the overall computational time required to evaluate the ID on different scales with TwoNN, $\sim N_{tot}^2  P (1+\nicefrac{1}{2}+\nicefrac{1}{4}+...+\nicefrac{1}{256})$, is almost the double of that of Gride, $\sim N_{tot}^2 P$. 

\begin{figure*}
  \centering
  \includegraphics[width=.65\textwidth]{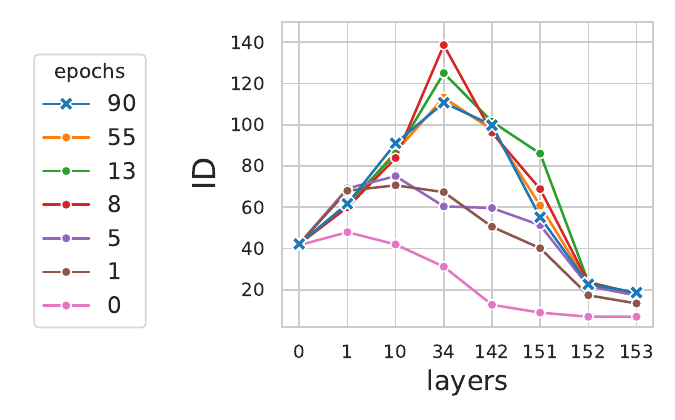}
  \caption{\label{fig:id_dynamics_resnet152} \textbf{Image GPT ID evolution during training.} We plot the ID of the ResNet152 hidden representations at seven training stages: before the training begins, at epochs 1, 5, 8, 13, 55, and at the end of training. The ID profile of the randomly initialized network decreases over the layers. The largest changes occur in the first 13 epochs.
  }
\end{figure*}

In Fig. \ref{fig:id_dynamics_resnet152} we show the dynamics of the ID measured with Gride; as in Fig. \ref{fig:id_resnet152} we take for simplicity the values computed with $k_1 = 2, k_2 = 4$.
The ResNet152 is trained for 90 epochs on ImageNet with weight decay = $10^{-4}$, batch size = 256, and initial learning rate = 0.1. We decrease the learning rate by a factor of 10 at epochs $\{30, 60, 80\}$. The input size is $224 \times 224$. We reach a top1 test accuracy of 78.22, consistent with \href{https://pytorch.org/vision/stable/models/generated/torchvision.models.resnet152.html#torchvision.models.ResNet152_Weights}{Pytorch\_ResNet152\_v1}. 
When the network is not trained, the ID after the first convolution decreases from 45 to 18. Then, after the first $8/13$ epochs the ID rapidly reaches its peak in all the layers, forming the hunchback shape observed at the end of training. 
The largest part of training is spent adjusting the ID of the representation toward the final value. This entails a steady decrease of the ID in the layers towards the output, while in the first 10/30 layers, the ID remains stable during training.

We conclude this section by showing the intra-class ID measured on the ImageNet training set in the space of the logits.
The average ID is around 20 in the output space (see Fig. \ref{fig:id_resnet152}-c), but we do not know how much the ID differs from class to class. 
\citet{ansuini2019intrinsic} showed that the ID of the output representation is correlated to the top5 test accuracy of the network: the lower the ID the higher the accuracy.

\begin{figure}
\includegraphics[width=0.9\columnwidth]{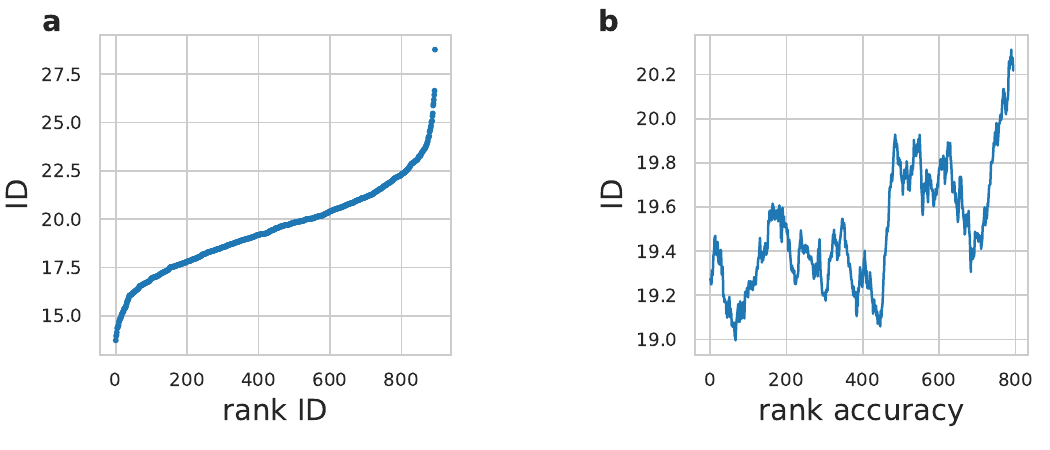}
\caption{{\bf Class-wise ID ImageNet training set.} 
{\bf a}: IDs measured with Gride $(k_1 = 2, k_2 = 4)$ on the ImageNet training set classes containing 1300 examples, sorted for increasing value of ID.
{\bf b}: Class ID as a function of the accuracy plotted with a moving average with a window size of 100.
}
\label{fig:classwise_id}
\end{figure}

It is therefore interesting to test if this property holds also at the class level. Since the number of images per class in the validation set is only 50, we evaluate the IDs and accuracies on the training set where there are 895 classes that contain 1300 images, and compute the ID with Gride using ($k_1, k_2) = (2, 4)$.
Figure \ref{fig:classwise_id}-a shows that the intra-class IDs are quite variable with an average of 19, 
the first and third quartiles are equal to 18 and 21, and maximum and minimum values are equal to 13 and 29. 

To capture the presence of a trend between IDs and class conditional accuracy in Fig. \ref{fig:classwise_id}-b we sort the data in order of increasing top1 training accuracy and plot the ID as a running mean over the sorted list with a window size equal to 100. 
This result shows that the correlation between ID and accuracy at the class level is very small.
Therefore the trend observed \citet{ansuini2019intrinsic} holds only when the comparison is done at the level of the full representation, on different architectures.

\subsection{Intrinsic dimension of the iGPT representations.} 
++Differently from the ResNet152, the Image GPT network (iGPT) is an attention-based architecture trained without labels to generate images \cite{igpt}. We analyze the small (S), medium (M), and large (L) versions of iGPT trained on ImageNet which have been publicly released
(\href{https://github.com/openai/image-gpt}{https://github.com/openai/image-gpt}).
After the pixel embedding and positional embedding layers, the architecture is composed of a sequence of identical attention blocks.
The number of blocks $l$, and the size of the embeddings $D$, are: ($l=24$, $D=512$),  ($l=36$, $D=1024$), ($l=48$, $D=1536$), for $S$, $M$, and $L$ respectively.
Due to the high memory footprint required by the attention layers, the image resolution is reduced from the standard ImageNet size ($r = 224^2$) to $r = 32^2$ (S), $r = 48^2$ (M), $r = 64^2$ (L) and the 3 color channels are encoded in an "embedding axis" with size $D_{input} = 512$. 
In practice, the $\mathbb{R}^3$ color space in which each pixel is represented by a triplet of real numbers $(R, G, B)$ is quantized with $k$-means clustering ($k=512$), and each pixel is described by the
\emph{discrete} "code" 
of the cluster where it belongs.
For each pixel $i$ the network outputs a probability distribution
\begin{equation}
    p(x_i) = p(x_i|x_0, ... x_{i-1}, \theta)
\end{equation}
$p(x_i) \in \mathbb{R}^{512}$ conditioned on the values of the previous pixel codes.
The network is trained to predict pixel codes by minimizing a negative log-likelihood loss \cite{igpt}: 
\begin{equation}
    L = \mathbb{E}_{x \sim X}\left[ -\log \prod_{i=1}^s p(x_i)\right]
\end{equation}
Once the network is trained, an image can be generated sampling from $p(x_i)$ one pixel at a time \cite{igpt}.

\begin{figure*}
  \centering
  \includegraphics[width=1.\textwidth]{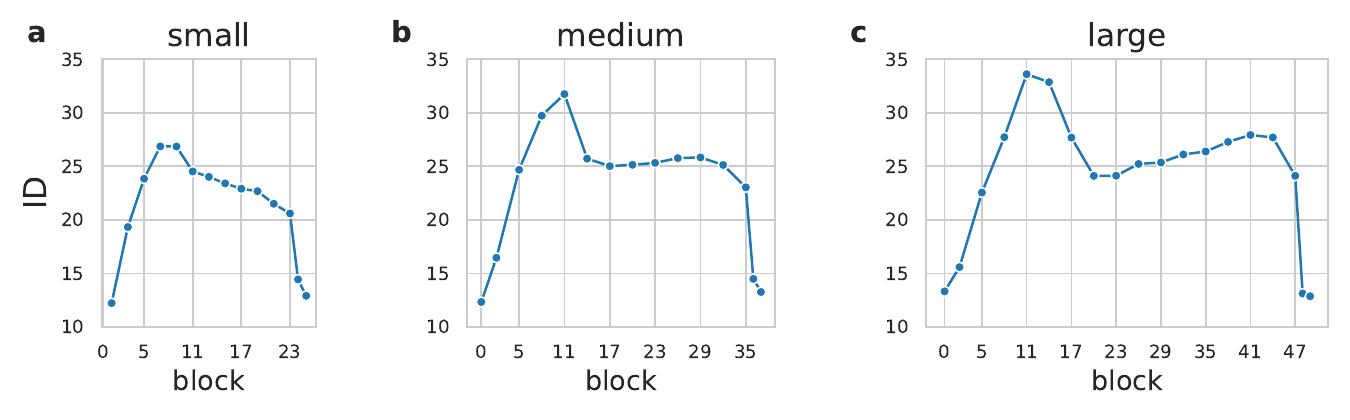}
  \caption{\label{fig:id_image_gpt} \textbf{Intrinsic dimension of Image GPT hidden layers}
  ID evaluated on the normalization layers before some attention blocks of iGPT small ({\bf a}),  medium ({\bf b}), and large ({\bf c}). The ID is measured with Gride setting $(k_1 = 2, k_2 = 4)$.
  As the size of the mode grows, a second peak of the ID profile can be observed. The peak can be clearly detected for iGPT large around layers 41/44.
  }
\end{figure*}

\begin{figure*}
  \centering
  \includegraphics[width=0.65\textwidth]{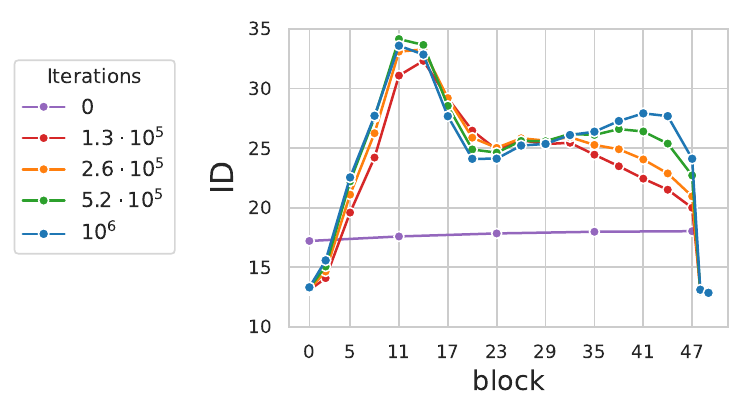}
  \caption{\label{fig:id_dynamics_image_gpt} \textbf{Evolution of the ID of iGPT representations during training.}  ID evaluated on the normalization layers before some attention blocks of iGPT Large during training. We measure the ID of the randomly initialized network, after $(1.3, 2.6, 5.2)\cdot 10^6$ iterations, and at the end of training. After $(2.6\cdot 10^6$ iterations, the first hunchback is formed, and during the remaining training epochs, the second hunchback grows.
  The checkpoints are available at https://github.com/openai/image-gpt. }
\end{figure*}

We consider 90000 images (from 300 classes using 300 images per class) of the ImageNet training set, as we did for the ResNet architecture (see Sec. \ref{sec:id_resnet152}), and measure the ID of the representations after the pixel embedding, at the normalization layers before the attention blocks, and at the output of the network. 
In all cases, we average over the sequence dimension, so that the size of the representation of an image is ($X_l \in \mathbb{R}^D$). 
Figure \ref{fig:id_image_gpt} shows the profile of the ID estimated with Gride $(k_1, k_2) = (2, 4)$ for the small, medium, and large iGPT networks from left to right. In all cases, the IDs of the output are similar to those of the input.
This is not surprising: since the network is trained to reconstruct a realistic completion of the images the geometric aspects of the data manifold, like the ID, should not be very different between output and input.
Across the hidden layers, the ID grows; but while on the small iGPT the ID profile has a hunchback shape quite similar to that observed in ResNet152 and in many convolutional discriminative models \cite{ansuini2019intrinsic},
on the medium and, most notably, large iGPT the ID has two peaks with a local minimum in the middle of the network.
The formation of the two peaks occurs at different stages during training. Figure \ref{fig:id_dynamics_image_gpt} shows the ID profile after
$\{0, \:1.3,\: 2.6,\: 5.2, \:10\}\cdot 10^5 $ iterations\footnote{These checkpoints are available at  https://github.com/openai/image-gpt}. 
After $2.6 \cdot 10^5$ iterations, the development of the first peak around layer 12 is completed but the second one at layer 40 is not yet formed. 
In the course of the last $7.4 \cdot 10^5$ iterations, the intermediate minimum after block 23 slightly decreases from 25  to 24.2, and the ID at block 41 grows from 24.3 to 26.3 (iteration $5.2 \cdot 10^5$) to finally reach 28.

We can not give a neat interpretation of these two-stage dynamics and of the presence of the double hunchback shape itself, but in the next chapter, we will see that the layer ID can be associated with the classification accuracy of a given representation. More precisely, the layer where the ID is lowest is also the one that achieves the highest classification accuracy. 
Indeed, in the original work, \citet{igpt} noted that the features extracted by the intermediate layers are those that reach the highest accuracy when fitted with a linear probe on ImageNet as well as on downstream tasks like CIFAR10, CIFAR100, and STL.
If we push the connection between ID and semantic expressiveness even further, a hypothesis is that up to iteration $2.6/5.2 \cdot 10^5$ the network tries to learn the abstractions needed for the generative process and encodes this knowledge in the middle layers of the architecture.
Once the first hunchback is formed and some degree of abstraction is extracted from the data, the second stage of training is devoted to refining the details of the generated images.

It is in any case interesting that in the examples we have seen, the most expressive representations in terms of high-level abstractions seem to be those in which the ID is low. 
In discriminative models, these representations are at the output of the network \cite{ansuini2019intrinsic} as the extraction of abstract features useful for classification is the main training task.
On the contrary in deep generative models, the semantically meaningful representation is the latent space, in the middle of the network \cite{vae_intro_welling, stable_diffusion}. 
The recent study of \citet{style_gan_id} analyzed the local dimension of the hidden layers in a StyleGAN, showing that the ID tends to be low in the "disentangled" $W$-latent space.
In a StyleGAN \cite{style_gan} the generator network is split in two subnetworks: a \emph{mapping network} $f: \mathcal{Z}\to\mathcal{W}$ and a \emph{syntesis network} $f: \mathcal{W}\times\mathbb{R}^n_0\to\mathcal{X}$. The mapping network is fully connected and is used to map the latent code $z$ into a space $W$ where the abstract features of the data set (\emph{styles}) are better disentangled. 
\citet{style_gan_id} showed that in the mapping network the ID decreases to reach its minimum in the $\mathcal{W}$ space, but they do not analyze the synthesis subnetwork.
An interesting future line of research is exploring in detail the connection between low ID and expressive latent spaces from a wider spectrum of generative models, from variational autoencoders \cite{nvae, vdvae, efficient_vdvae} to the more recent diffusion-based models \cite{ho_ddpm, diffusion_beats_gans, stable_diffusion}. 

\chapter{Hierarchical nucleation in deep neural networks}
\label{ch:hier-nucl}
\section{Introduction}
Deep convolutional networks (CNNs) have become fundamental tools of modern science and technology.
They provide a powerful approach to supervised classification, allowing the automatic extraction of meaningful features from data. 
The capability of CNNs to discover representations without human input has attracted the interest of the machine-learning community.
In the intermediate layers of a CNN,  the data are represented with a set of features (the activations) embedded in  a manifold whose tangent directions capture the relevant factors of variation of the inputs \cite{bengio2013representation, goldt2020modelling}.
Accordingly, understanding these data representations requires both studying the geometrical properties of the underlying manifolds and characterizing the data distributions on them.

In the present chapter, we analyze how the probability density of the data changes across the layers of a CNN. We consider in particular CNNs trained for classifying ImageNet; as we will see, the complexity and heterogeneity of this dataset critically affect the results of our analysis. 

Comparisons between representations based on generalizations of multivariate correlation have already been performed with the methods in  \cite{svcca} (SVCCA), \cite{morcos2018insights} (PWCCA), and, more extensively, in \cite{cka} (CKA).
Representational similarity analysis (RSA) \citet{kriegeskorte2008representational} -- introduced originally in neuroscience -- investigates artificial representations as well, and in each layer, a matrix of pairwise distances (representation dissimilarity matrix (RDM) ) between the activation vectors of the data points tells which data is similar or dissimilar in that layer.
The introduction of RDMs allowed performing multiple comparisons including those between artificial and biological networks \cite{khaligh2014deep, yamins2014performance, cadieu2014deep}.

More recently many techniques have been proposed to understand representations. \citet{2017dissection} shows that feature maps are often aligned to a specific explanatory factor of the data distribution and therefore behave as detectors of specific concepts. In \citet{2017linear_probes} linear separability of the features was shown to increase with a smooth monotonic trend across the internal layers.
In \citet{bilal2017convolutional} the question of whether CNNs learn a hierarchy of classes was addressed by exploiting class confusion patterns.
Other studies investigated more specifically the geometrical and structural properties of the representations. 
\begin{figure}[!t]
\centering
\includegraphics[width=\textwidth]{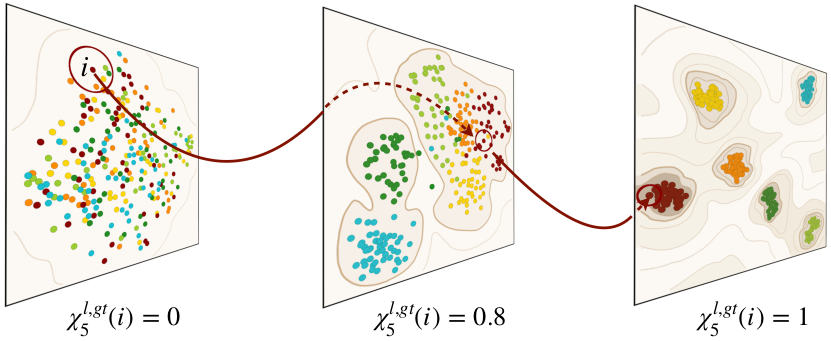}
\caption{{\bf Evolution of the data representations in ResNet152}. Projections of the representations of the input (left), conv4 (middle), and output layers in ResNet152 for six ImageNet classes. Contours schematically portray the density isolines on the data manifold. The dark red circles surround the five nearest neighbors of a point $i$; $\chi_5^{l,gt}(i)$ represents the fraction of these points that are in the same class as $i$.}
\label{fig:cartoon}
\end{figure}
In \citet{id}, a common trend in the intrinsic dimension was found across several architectures; in \citet{salakhutdinov2007learning}, the soft-neighbor loss was used as a tool to reveal structural changes in neighbors organization across layers, also during training \cite{frosst2019analyzing}.

Here we take a complementary perspective. One can view a CNN as an engine capable of iteratively shaping a probability density. Input data can be seen as instances harvested from a given probability distribution. 
This distribution is then modified again and again by applying, at each layer, a non-linear transformation to the data coordinates. The result of this sequence of transformations is well understood:  in the output layer of a trained network, data belonging to different categories form well-separated clusters, which can be viewed as distinct peaks of a probability density (see Fig. \ref{fig:cartoon}).
But where in the network do these peaks appear? Do they develop slowly and gradually or all of a sudden? Is this change model-specific, or is it shared across architectures? And what is the probability \emph{flux} between one layer and the next? 
In the input layer, the data points are mixed: data with different ground truth labels are close to each other. In the output layer, the neighborhood of each data point is ordered, namely, it contains mostly data points belonging to the same category. Where in the network does the transition from disordered  to ordered neighborhoods take place? 
Is it simultaneous with the formation of the probability peaks?

The pivotal role of depth in determining the accuracy of a neural network suggests that the transformation of the probability density should be slow and gradual to be effective. The analysis reported in 
\citet{svcca, morcos2018insights, cka} 
are consistent with this scenario. However, we will see that, especially in a CNN trained for a complex classification task the evolution of the probability density is not smooth, with spikes in the probability flux and sudden changes in the modality.  
We analyze the probability landscape in the intermediate layers of a CNN 
by a technique that estimates the probability density and characterizes its features even if this is defined as a function of hundreds of thousands of coordinates, provided that the data are embedded in a relatively low-dimensional manifold \cite{cluster, 2014density_peak}.
This approach, described in detail in Sec. \ref{sec:density_peak}, \emph{does not} require a dimensional reduction of the data and has a great advantage that the embedding manifold does not necessarily have to be a hyperplane, but can be arbitrarily curved, twisted, and topologically complex.
To analyze the probability flux between the layers, we use an extension of neighboring hit \cite{NH}, described in Sec. \ref{sec:ov}.
The main results of this analysis also sketched in Fig. \ref{fig:cartoon} can be summarized as follows:

\begin{itemize}
    \item Representations in CNNs trained for complex classification tasks  do not evolve smoothly, but through nucleation-like events, in which the neighborhood of a data point changes rather suddenly (Sec. \ref{sec:nucl}); 
    
    \item In the first layers of the network, any structure which is initially present in the probability density of the input is washed out, reaching a state  with a single probability peak where the neighborhoods mainly contain simple images characterized by elementary geometrical shapes (Sec. \ref{sec:entropy});
    
    \item In the successive layers, a structure in the landscape starts to emerge, with probability peaks appearing in an order that mirrors the semantic hierarchy of the data set: neighborhoods are first populated by images that share the same high-level attributes (Sec. \ref{sec:hier_nucl_evolution_cluster_ari});
    
    \item In the output layer, the probability landscape is formed by density peaks containing data points with the same ground truth classification; interestingly, these peaks are organized in complex \q{mountain chains} resembling the semantic kinship of the categories (Sec. \ref{sec:hier_nucl_evolution_cluster_ari}). 
    
\end{itemize}
In short, we find that the disorder-order transition induced by a trained CNN can be characterized without any reference to the ground truth categories as a sequence of changes in the modality of the probability density of the representations.
These changes are achieved by reshuffling the neighbors of the data points again and again, in a process that resembles the diffusion in a heterogeneous liquid, followed by the nucleation of an ordered phase. 

\section{Methods}
\label{sec:hier_nucl_methods}
The CNNs we consider in this work are classifiers ${\bf y} = f({\bf x})$ that map a data point ${\bf x}_i \in \mathbb{R}^p$, for example, an image, to its categorical target ${\bf y}_i \in \{0, 1\}^q $ typically encoded with a one-hot vector of dimension $q$ equal to the number of classes.
Feedforward networks achieve the task via a function composition  $f = f^{(1)} \to f^{(2)} \to ... \to f^{(L)}$ that transform the input sequentially ${\bf x}_i \to  {\bf x}^{(1)}_i \to ... \to  {\bf x}^{(L)}_i $.
We call the vector ${\bf x}^{(l)}_i$ containing the value of the activations of the $l$-th layer for data point $i$  the {\it representation} of ${\bf x}_i$ at the layer $l$. 
The sequence of representations of these data points on a trained network can be seen as a \q{trajectory} in a very high dimensional space.
The relative positions of the $N$ inputs change from an initial state where the neighborhood of each point contains members of different classes to a final state where images of the same class have been mapped close together to the same target point.
We study this process with two approaches, one aimed at describing the probability flux across the layers (Sec. \ref{sec:ov}) and the other aimed at characterizing the features of the probability density in each layer (Sec. \ref{sec:density_peak}).

\subsection{The data set and the network architecture}\label{sec:hier_nucl_data}
We  perform our analysis on the ILSVRC2012 data set, a subset of $1000$ mutually exclusive classes of ImageNet which can be considered leaves of a hierarchical structure with 860 internal nodes.
The highest level of the hierarchy contains seven classes, but $95\%$ of the ILSVRC2012 images belong to only two of these (artifacts or animals) and are split almost evenly between them ($55\%$ and $45\%$ respectively). 
Unless otherwise stated, the analysis in this work is performed on a subset of $300$ randomly chosen categories, including $300$ images for each category, for a total of $90,000$ images.

We extracted the activations of the training set of ILSVRC2012 from a selection of pre-trained PyTorch models: ResNets \cite{resnet}, DenseNets \cite{2017densenet}, VGGs \cite{vgg} and GoogleNet \cite{2014googlenet}. 
To compare  architectures of different depths we will use as {\it checkpoints} the layers that downsample the channels and the final fully connected layers. In these layers, the learned representations become more abstract and invariant to details of the input irrelevant for the classification task \cite{bengio2013representation}.

\subsection{The neighborhood overlap}\label{sec:ov}
We now introduce an observable to characterize the similarity between a pair of representations, looking at the neighborhood composition of each data point. 
Let $\mathcal{N}_{k}^l(i) $ be the set of $k$ points nearest to  ${\bf x}_i^{(l)} $ in euclidean distance at a given layer $l$, and let $A^l$ be an $N \times N$ adjacency matrix with entries 
$A^l_{{ij}} =1$ if $j \in \mathcal{N}_k^l(i) $ and $0$ otherwise.
Through $A$ we define an index of similarity $\chi_k^{l,m} \in [0, 1]$ between two layers $l$ and $m$  as: 
\begin{equation}
\chi_k^{l,m} = \dfrac{1}{N} \sum_i \dfrac{1}{k} \sum_j {A^l}_{ij} A^m_{ij}   
\label{eq:overlap}
\end{equation}
The similarity just introduced has a very intuitive interpretation: it is the average fraction of common neighbors in the two layers considered: for this reason, we will refer to $\chi_k^{l,m}$ as the \emph{neighborhood overlap} between layers $l$ and $m$.

In the same framework, we also compare the similarity of a layer with the ground truth categorical classification, defining the \q{ground truth} adjacency matrix
$A^{gt}_{{ij}} =1$ if $y_i = y_j  $ and $0$ otherwise. In this case, $\chi_k^{l,gt}= \dfrac{1}{N} \sum_i \dfrac{1}{k} \sum_j {A^l}_{ij} A^{gt}_{ij} $ is the average fraction of neighbors of a given point in $l$ that are in the same class as the central point (see Fig.~\ref{fig:cartoon}).
We set $k=30$, equal to one-tenth of the number of images per class, but we verified that our findings are robust with respect to the choice of $k$ over a wide range of values (see Sec. \ref{sec:nucl}).
${A^l}_{ij}$ and $A^{gt}_{ij} $ are built using the euclidean distances between images, as such they are invariant to orthogonal transformations but not to any arbitrary linear transformation of the activations. These convenient properties for similarity indices between representations \cite{cka} are inherited by $\chi_k^{l,m}$ and $\chi_k^{l,gt}$.
When calculated using the ground truth adjacency matrix as a reference,  $\chi_k^{l,gt}$ reduces  to the neighboring hit \cite{NH}. 
A measure of overlap quantitatively similar to $\chi_k^{l,m}$ can be obtained by using the CKA method \cite{cka} with a Gaussian kernel of a very small width (see Sec. \ref{sec:hier_nucl_discussion}).

\subsection{Estimating the probability density}\label{sec:density_peak}
We  analyze the structure of the  probability density of data representations following the approach in \citet{cluster, 2014density_peak}, which allows finding the peaks of the data probability distribution  and the location and the height of the saddle points between them.
This in turn provides information on the relative hierarchical arrangement of the probability peaks.
The methodology works as follows.

\paragraph{Finding the density peaks.}
Using a $k$NN estimator the local volume density $\rho_i$ around each point $i$ is estimated:
\begin{align}
\label{eq:knn_density}
    \rho_i &= \dfrac{k}{ N \omega_d r_{k_i}^d}
\end{align}
where $k$ is a hyperparameter, here fixed to 30, $N$ is the total number of points in the data set, $r_{k_i}$ is the distance to the $k^{th}$ nearest neighbor of the point $i$, and $\omega_d$ is the volume of the unit ball. 
In Sec. \ref{sec:pak} we will see that the density estimation can be made fully unsupervised with the procedure proposed in \cite{pak} which allows determining for each point, the optimal value of $k_i$ to compute the local density.
In either case, crucially $d$ is the value of the ID estimated with an appropriate approach, as discussed in chapter \ref{ch:gride}. 
As we saw in Sec. \ref{sec:id_resnet152} despite the number of activations $D$ in a representation can be very large  ($D \sim 10^5$) the volume is computed using the manifold ID which is much smaller, typically between 10 and 100. 

The maxima of $\rho_i$ (namely the probability peaks) are then found.
Data point $i$ is a maximum if the following two properties hold: (I) $\rho_i>\rho_j$ for all the points $j$ belonging to $\mathcal{N}_k(i)$; (II) $i$ does not belong to the neighborhood $\mathcal{N}_k(j)$  of any other point of higher density \cite{cluster}.
(I) and (II) must be jointly verified as the neighborhood ranks are not symmetric between pairs of points.
A different integer label $\mathcal{C} = \{c^1, ... c^n\}$ is then assigned to each of the $n$ maxima, and the data points that are not maxima are iteratively linked to one of these labels, by assigning to each point the same label of its nearest neighbor of higher density.
The set of points with the same label corresponds to a density peak.

\paragraph{Saddle points and peaks significance}
Not all the density peaks can be regarded as genuine \emph{probability modes} because of spurious fluctuations of the density due to finite sampling. 
The significance of the peaks can be assessed by comparing the difference between the density maxima and the density of the saddle points between the peak relative to the statistical error of Eq. \ref{eq:knn_density}. We approximate the error with the asymptotic standard deviation \cite{pak}:
\begin{equation}
    e_{\rho_k} = \sqrt{\dfrac{4k_i +2}{(k_i-1)k_i}}
\end{equation}

The saddle points between two peaks are identified as the points of maximum density between those lying on the borders between the peaks.
A point  ${\bf x}_i \in c^\alpha$ is assumed to belong to the border $\partial_{c^\alpha, c^\beta}$ with a different peak $c^\beta$ if exists a point ${\bf x}_j \in \mathcal{N}_k(i) \cap c^\beta$ whose distance from $i$ is smaller than the distance from any other point belonging to  $c^\alpha$.
The saddle point between $c^\alpha$ and $c^\beta$ is the point of maximum density in $\partial_{c^\alpha, c^\beta}$.
The statistical reliability of the peaks is then assessed  as follows:
let $\rho^\alpha $ be the maximum density of peak $\alpha$, and $\rho^{\alpha,\beta} $ the density of the saddle point between $\alpha$ and $\beta$.
If:
\begin{equation}
\label{eq:z_clusters}
\log \rho^\alpha - \log \rho^{\alpha,\beta} < 2 Z \sqrt{e_{\rho_\alpha}^2 +e_{\rho_{\alpha, \beta}}^2}
\end{equation} 
peak $\alpha$ is merged with peak $\beta$ since the value of its density peak is considered indistinguishable from the saddle point at statistical confidence fixed by the parameter $Z$ \cite{cluster}.
The process is repeated until all the peaks satisfy this criterion, and are therefore statistically robust with a confidence $Z$.

\subsection{Local density estimation with PAk}\label{sec:pak}
Instead of fixing the number $k$ a priori, \citet{pak} showed that a better strategy to evaluate the density $\rho_i$ (see Eq. \ref{eq:knn_density}) around each point is based on the maximum neighborhood size at which $\rho_i$ can be considered constant around each point. 

Pa$k$ assumes that the data are locally distributed according to a Poisson process in space (see Sec. \ref{sec:derivation_mu}).
We recall that under the Poisson assumption, the points are distributed according to:
\begin{equation}
    P(k, \rho) = \dfrac{(\rho V)^k}{k!}e^{-\rho V},
\end{equation}
which gives the following log-likelihood written as a function of the density $\rho$:
\begin{equation}
\label{eq:rho_lik}
    \mathcal L (\rho) = k \log\rho -\rho V;
\end{equation}
If $k$ is fixed, the  value of $\rho$ that maximizes $\mathcal L$ is $\hat \rho = \nicefrac{k}{V}$. 
To find the optimal $k$, \citet{pak} compare, for increasing value of $k$, two models that we denote, for clarity, $M1$, $M2$.
$M1$ assumes that the value of the density of a point $i$, $\rho_{i, k}$, and that of the $k+1$ nearest neighbor of $i$, $\rho_{j, k}$, are different. Both the densities of $i$ and $j$ are evaluated up to their $k$ nearest neighbors. 
The likelihood of M1 then is: $\mathcal{L}_{M1} = \mathcal{L}(\rho_{i, k})+ \mathcal{L}(\rho_{j, k}) =  (k \log \rho_{i,k}-\rho_{i,k}V_{i, k}) +(k \log \rho_{j,k}-\rho_{j,k}V_{j, k})$.
In M2, the densities at the points $i$ and $j$ are equal, $\rho_{i, k}=\rho_{j, k} \doteq \rho_k$, and the likelihood reads: $\mathcal{L}_{M2} = 2k\log \rho_{k}-\rho_{k}(V_{i, k}+ V_{j, k})$. %
Fixing $k$, and maximizing  $\mathcal{L}_{M1}$ with respect to $\rho_{i, k}$, $\rho_{j, k}$ and $\mathcal{L}_{M2}$  with respect to $\rho_{k}$  gives:
\begin{align}
&\mathcal L^*_{M1} (k) = k \log \dfrac{k^2}{V_{i, k} V_{j, k}} -2k \\
&\mathcal L^*_{M2} (k) = 2k \log \dfrac{2k}{V_{i, k}+ V_{j, k}} -2k
\end{align}
The two models are compared with a likelihood ratio test and considered distinguishable if the difference between $\mathcal L^*_{M1}$ and $\mathcal L^*_{M2}$ exceeds a threshold corresponding to a $p$-value of $10^{-6}$. 
According to this criterion, the optimal neighborhood size $k_i^*$ is the maximum value of $k$ for which $M1$ and $M2$ are consistent. Therefore, the density at point $i$ is equal to $\rho_i^* = \nicefrac{k_i^*}{V_{i, k}}$.
To correct a residual bias of the estimate \citet{pak}, impose a linear correction to $\log(\rho_i^*)$ as a function of the distance to the point $i$. 
The final value of $\rho_i^*$ and the slope of the linear correction are found by maximizing the likelihood of Eq. \ref{eq:rho_lik} with $k$ kept fixed and equal to $k^*_i$.

\subsection{Analysing the composition of the peaks}
\label{sec:hier_nucl_methods_ari}
The saddle point density $\log \rho^{\alpha, \beta}$ provides a notion of similarity between pairs of peaks: the higher $\log \rho^{\alpha, \beta}$, the more similar the two clusters can be considered. We construct a pairwise \emph{dissimilarity} as follows:
\begin{equation}
\label{eq:similarity}
    S_{\alpha, \beta} = \log \rho_{max} - \log \rho_{\alpha,\beta}
\end{equation}
$\rho_{max}$ is the density of the highest peak and allows to normalize $S$ in a way that when the two clusters can be considered similar, $S_{\alpha, \beta}$ is close to zero.
According to $S$, the peaks can be further grouped in a hierarchical structure joining at each level of the hierarchy the closest peaks according to $\log \rho^{\alpha, \beta}$. 
Once two peaks $c^\alpha$, $c^\beta$ have been merged, the saddle point density between the broader peak $c^\gamma$ representing their union and the rest of the peaks $c^{\delta_i}$ is determined according to the Weighted Pair Group Method with Arithmetic Mean (WPGMA) \cite{sokal1958statistical}:
\begin{equation}
   \log \rho^{\gamma, \delta_i}= \dfrac{\log \rho^{\alpha, \delta_i}+\log \rho^{\beta, \delta_i}}{2}
\end{equation}
In Sec. \ref{sec:hier_nucl_results} we will use this agglomerative procedure to build a hierarchical representation of the peaks.

To determine to which degree the density peaks are consistent with the abstractions of the data, we will compare the clustering induced by the density peaks algorithm with the partition given by the 300 classes and that given by the high-level subdivision in animals and objects.
Among the many possible scores, we choose the Adjusted Rand Index (ARI) \cite{ari_hubert}, one of the best clusters external evaluation measures according to \cite{ref-rand_indx}. 
The Rand Index (RI) \cite{rand_indx} measures the consistency between a cluster  partition $\mathcal{C}$ to a reference partition $\mathcal{R}$ counting how many times a pair of points:

$a$ are placed in the same group in $\mathcal{C}$ and $\mathcal{R}$;

$b$ are placed in different groups in $\mathcal{C}$ and $\mathcal{R}$;

$c$ are placed in different groups in $\mathcal{C}$ but in the same group in $\mathcal{R}$;

$d$ are placed in the same group in $\mathcal{C}$ but in different groups in $\mathcal{R}$;\\
and measures the consistency with $RI = (a+b)/(a+b+c+d)$. 
The Rand Index is not corrected for chance meaning that it does not give a constant value (e.g. zero) when the two assignments are random. 
\citet{ari_hubert} proposed to adjust $RI$, taking into account the expected value of the Rand Index, $n_c$, under a suitable null model for chance:
\begin{equation}
    ARI = \dfrac{a + b-n_c}{a+b+c+d-n_c}
\end{equation}
$ARI$ is equal to 1 when the two partitions are consistent is 0 when the assignments are random, and can take negative values. 
A large value of $ARI$ not only implies that instances of the same class are put in the same cluster (homogeneity) but also that the data points of a class are assigned to a single cluster (completeness). 

\paragraph{Reproducibility}  The source code of our experiments with the instructions required to run it on a selection of layers is available at \href{https://github.com/diegodoimo/hierarchical_nucleation.git}{https://github.com/diegodoimo/hie-rarchical\_nucleation}.

\section{Results}
\label{sec:hier_nucl_results}
It is well known that neural networks modify the representations of the data from an initial state where all the data are randomly mixed, to a final state where they are orderly clustered according to their ground truth labels \citet{frosst2019analyzing}. 
But where in the network does this order arise? In the output layer, the nearest neighbors of, say, the image of a cat are very likely other images of cats. But in which layer, and in which manner do cat-like images come together?
We describe the ordering process by analyzing the change in the probability distribution across the layers. 

\subsection{The evolution of the neighbor composition in a CNN}\label{sec:nucl}
\begin{figure}[!tp]
\centering
\includegraphics[width=1\columnwidth]{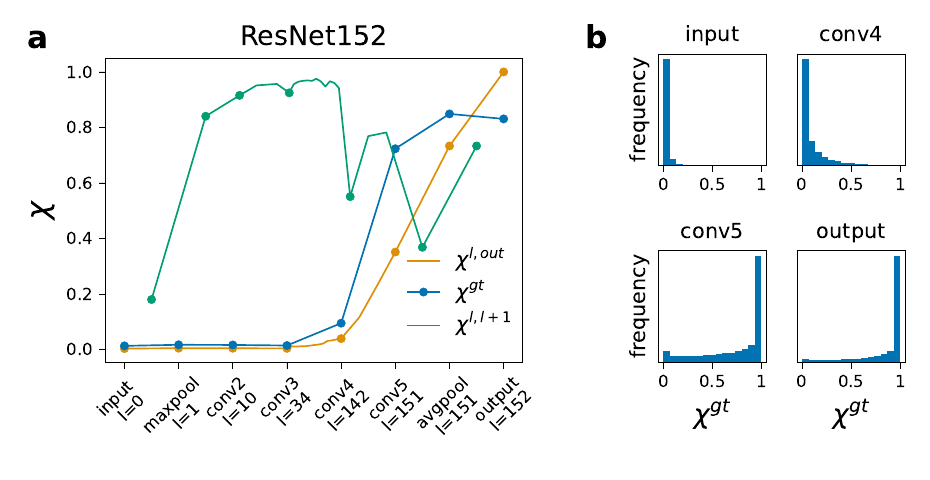} 
\caption{ {\bf Overlap profiles in ResNet152 and in different architectures}. ({\bf a}:) Overlap with the ground truth $\chi^{l,gt}$ (blue), with the output $\chi^{l,out}$ (orange) and between nearby layers $\chi^{l,l+1}$ (green). Checkpoint layers are evenly spaced on the $x$-axis and marked with dots.
({\bf b}:) Probability distribution of $\chi^{l,gt}$ approximated by its frequency histogram. The histograms are built collecting $\chi^{l,gt}(i) = \sum_j {A^l}_{ij} A^m_{ij}/k$ (see Eq. \ref{eq:overlap}) for each data point $i$ in four layers.
}
\label{fig:ov_gt}
\end{figure}

We first characterize the probability flux  by computing the neighborhood overlap $\chi^{l,out}_{30}$: the fraction of $30$-neighbors of a data point which are the same in layer $l$ and in the output layer (Eq. \ref{eq:overlap}). For the sake of brevity from now on, we will omit the subscript $k=30$.
Figure \ref{fig:ov_gt}-a shows the  behavior of  $\chi^{l,out}$ as a function of $l$ for the checkpoint layers of the ResNet152 described in Sec. \ref{sec:hier_nucl_data} (orange line).
The neighborhood overlap remains close to zero up to $l$=142.
In the next nine layers, it starts growing significantly, reaching a value of 0.35 in layer 151 and 0.73 in layer 152, the last before the output.
In the same figure, we also plot the neighborhood overlap of each layer with the ground truth classification $\chi^{l,gt}$ (blue line).
After layer 142, $\chi^{l,gt}$ changes even more abruptly than $\chi^{l,out}$, increasing from 0.10 to 0.72. 

We can obtain more insights into this transition process by looking at the probability distribution of $\chi^{l,gt}$ across the data set in four different layers (see Fig. \ref{fig:ov_gt}-b).
In the input and output layers, the probability distribution is unimodal.
In layer 142 (conv4), before the onset of the transition, the distribution is still strongly dominated by disordered neighbors (i.e. $\chi^{l,gt} \approx 0$ 
for most of the data points), but an ordered tail starts to emerge. 
In layer 151 (conv5), immediately after the transition, the distribution indicates the coexistence of some data points in which the neighborhood is still disordered or only partially ordered ($\chi^{l, gt} \approx 0$) and some data in which it is already ordered ($\chi^{l, gt} \approx 1$).
These results show that ordering, when measured by the consistency of the neighborhood of data points with respect to their class labels, changes abruptly, in a manner that qualitatively resembles the phase transition of a \q{nucleation} process. 

\begin{figure}
\centering
\vspace{0.cm}
\includegraphics[width=0.5\columnwidth]{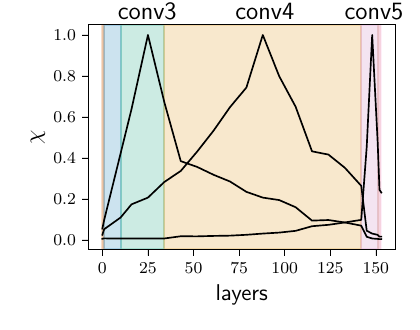}
\caption{Overlap with layers 25, 88, 148 in ResNet152. Different background colors indicate different ResNet blocks.
On average, the number of layers required to change half of the neighbors is $\sim 20$ in conv3 and $\sim 30$ in conv4, while in conv5, where the nucleation takes place, the same change occurs in just one layer.
The neighborhood composition changes significantly also between two blocks when the channels are downsampled. 
}
\label{fig:app3_memory}
\vspace{0cm}
\end{figure}

The green profile of \ref{fig:ov_gt}-a reinforces this evidence showing the overlap between two nearby layers $\chi^{l,l+1}$. This quantity is a measure of the probability flux between any two consecutive layers. 
In the first layer, the neighborhoods are almost completely reshuffled, as indicated by $ \chi^{0, 1} \sim 0.2$. 
Afterward, up to layer 142, $\chi^{l, l+1} \sim 0.95$ indicating that the neighborhoods change their compositions smoothly like in a slow diffusion process. 
In the two central blocks, from layer 10 to layer 142, it takes 20-30 layers to change half of the neighbors of each data point, i.e. to decrease
to 0.5 (see Fig. \ref{fig:app3_memory}). 
At layer 142, the first ordered nuclei appear and $\chi^{l,l+1}$ drops to 0.55 in just one layer.
A significant reshuffling of the neighborhoods takes place again at the last average pooling layer where $\chi^{l,l+1}$ drops again to 0.31.
We will see in Sec. \ref{sec:hier_nucl_evolution_cluster_ari} that  in  this layer the structure of the probability density changes significantly, and the probability peaks corresponding to the \q{correct} categories appear. 
The effective \q{attractive force} acting between data with the same ground truth label overcomes the entropic-like component coming from the intrinsic complexity of the images, and clusters of akin images emerge almost all at the same time (i.e., at the same layer), giving rise to a sharp transition.
\begin{figure}
\centering
{\includegraphics[width=1.0\columnwidth]{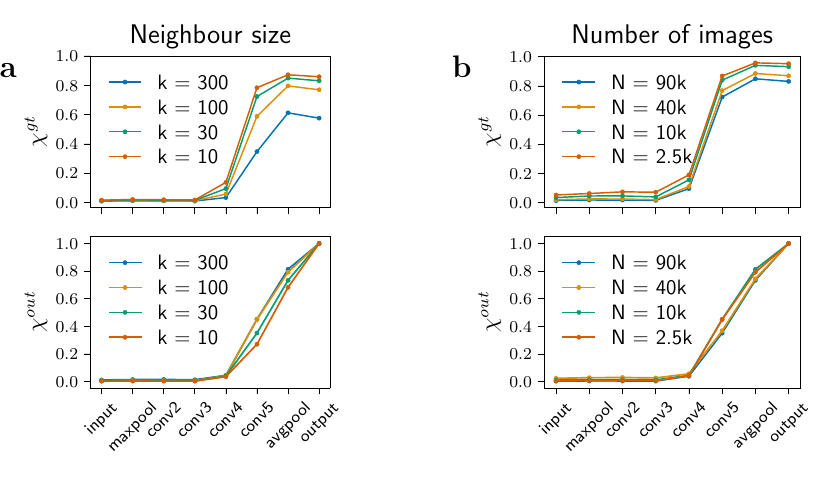}
	\caption{({\bf a}:) Profiles of the overlap with the ground truth labels (top) and with the output layer (bottom) as a function of the neighbor size $k$ for $N = 90000$. ({\bf b}:) Profiles of the overlap with the ground truth labels (top) and with the output layer (bottom) as a function of the total number $N$ of images. As the number of examples $N$ increases we keep the ratio between the number of classes and images per class constant and set $k$ to one tenth the number of images per class}
\label{fig:app1}
}
\end{figure}
Despite this analysis is done with a $k=30$ and a data set size $N = 90000$, as we show in Fig. \ref{fig:app1} the trends of $\chi^{l, gt}$ and $\chi^{l_out}$ are fairly robust over a wide range of $k$ and $N$.

\begin{figure}
\centering
{\includegraphics[width=0.9\columnwidth]{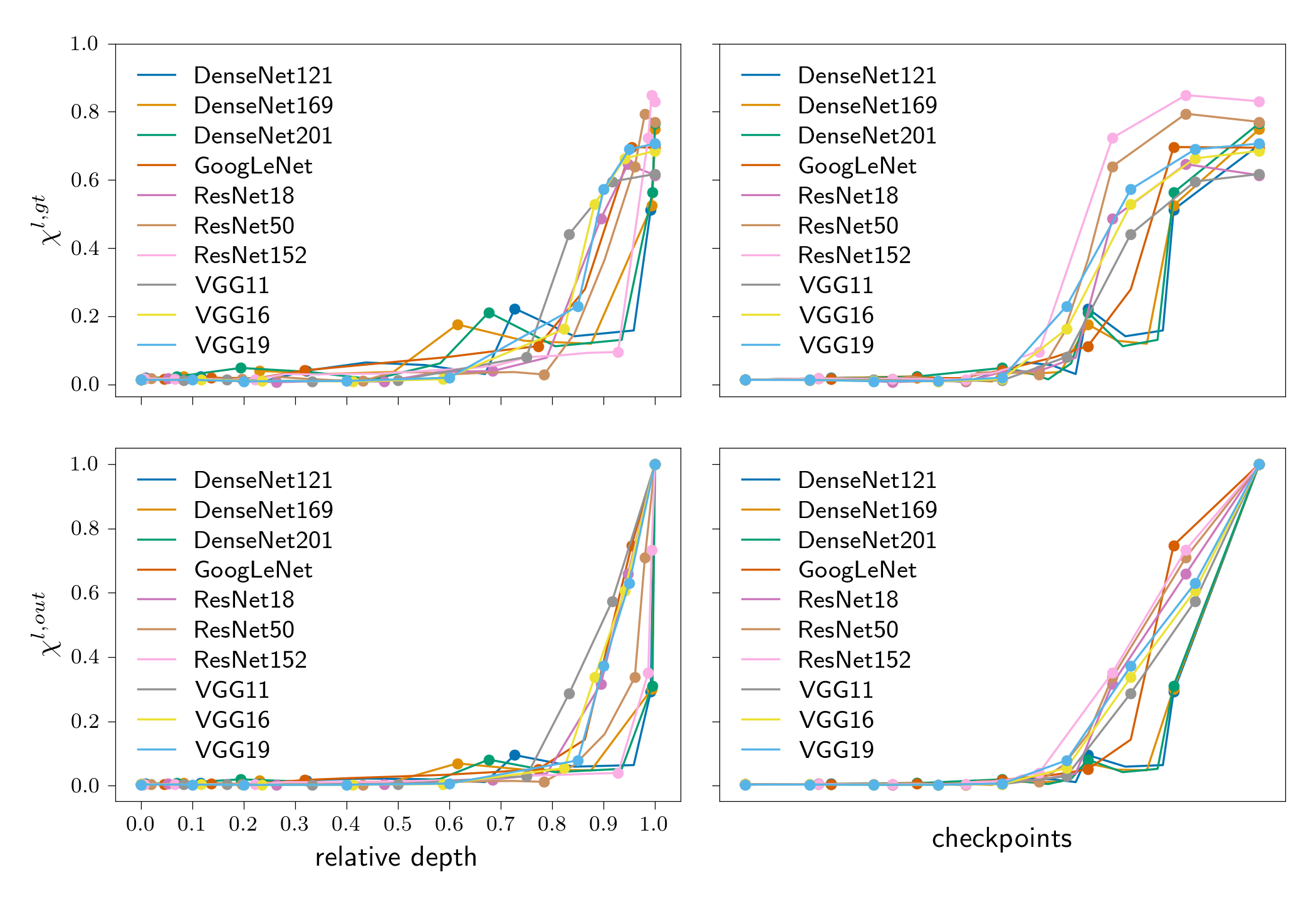}
	\caption{Profiles of $\chi^{l, gt}$ (top panels) and $\chi^{l, out}$ (bottom panels) for a subset of layers of ten deep neural networks. The layers that downsample the channels (checkpoints) are represented with dots. On the $x$-axis of the left panels, we rescale the layer depths by the network size; on the $x$-axis of the right panels, we instead display the checkpoints uniformly spaced. }
\label{fig:app1rev}}
\end{figure}

Is the sudden change we observed specific to this architecture or is it a common feature of deep networks trained for similar tasks? 
To answer this question, we repeated the same experiment on 3 architectures of different depths belonging to the ten networks ResNets, DenseNet, VGG family, plus GoogleNet architecture (see Fig. \ref{fig:app1rev}).
All the profiles increase significantly only close to the output.
The left panels, where the layer depth has been normalized with respect to the network depth, show that both $\chi^{l,gt}$ and $\chi^{l, out}$ tend to increase more abruptly and closer to the output layer when the network is deeper. 
In the DenseNets, the overlaps do not increase monotonically for the layers \emph{inside} a dense block. Within a dense block, the sequence of mapping is not "Markovian" \cite{2017densenet}, meaning that there are a lot of skip connections embracing the representations on which we compute $\chi^{l, out}, \chi^{l, gt}$.
The monotonic trend can be restored by plotting the overlaps only in correspondence of the transition layers, the representations \emph{between} the dense blocks.

\subsection{The data landscape before the onset of ordering}\label{sec:entropy}

\begin{figure}
\includegraphics[width=0.9\columnwidth]{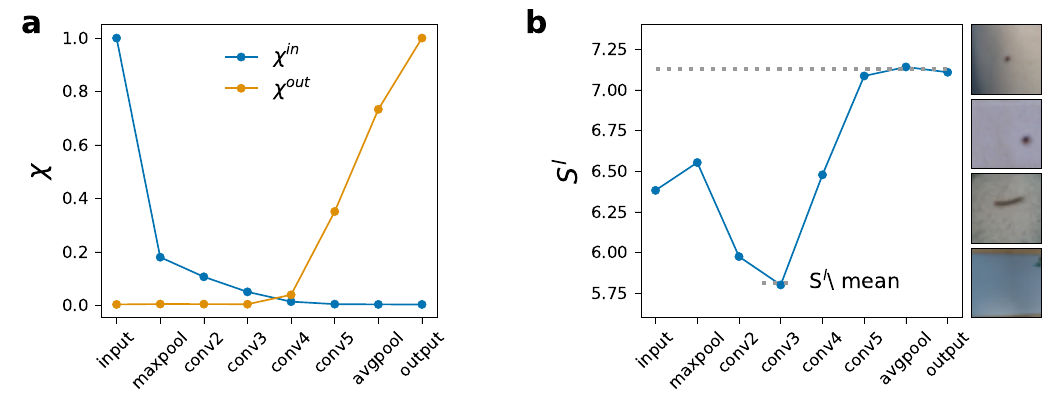}
\caption{{\bf  Image entropy in ResNet152.} ({\bf a}:) Overlap with the input $\chi^{in}$ (blue), and output $\chi^{l,out}$ (orange) layers.
({\bf b}:) Average image image entropy $S^l$ within the first 30 neighbors; the errorbars are shorter than the marker size; the most frequent images found in conv3 are shown on the right.}
\label{fig:2_ov_ll_entr}
\end{figure}

It has been argued that the first layers of deep networks serve the important task of getting rid of unimportant structures present in the data set \cite{shwartz2017opening, achille2018emergence, id, lecun2015deep}.
This phenomenon is illustrated in the upper panel of Fig.~\ref {fig:2_ov_ll_entr}, which shows that any overlap with the input layer is lost roughly after the conv3 landmark (layer 34). 

In the intermediate layers, the CNNs analyzed in this work construct high-dimensional hyperspherical arrangements of points with very few images at the center.
This is related to the high intrinsic dimension (ID) of these layers \cite{id}.
When the ID is very high, few data points act as \q{hubs} \cite{hubs}, namely they fall in a large fraction of the other point's neighborhoods while the others fall in just a few.
%
%
%
%
%
Moving from the input to conv3, the same images appear in a growing number of neighborhoods.
In conv3, the top ten most frequent images are found in almost half of the 90,000 neighborhoods, with a high of 75,663 for the most frequent of all.

Interestingly, we found that hub images are particularly \q{simple}, looking in most cases like elementary patterns (dots, blobs, etc.) lying on almost uniform backgrounds (see Fig.~\ref{fig:2_ov_ll_entr}, bottom right).
To quantify this perceptual judgment, we computed the Shannon entropy of an image $S = -\sum_{n_c}\sum_{v} p_v  \log_{2}(p_v)/n_c$ where $p_v$ is the normalized frequency of pixels of value $v$ and $n_c = 3$ is the number of channels of RGB images. The average entropy of the neighbors of an image $i$ in a layer $l$ is given by $S_i^l = \sum_{j \in \mathcal{N}_k(i)} S_j^l /k$, and averaging across all images, we obtain a measure of the neighborhood entropy of a layer $S^{l}$. A low value of $S^{l}$  means that, in that layer, the neighborhoods contain many low-entropy images. 
In the bottom panel of Fig. ~\ref{fig:2_ov_ll_entr} we show how  $S^{l}$ changes: in intermediate layers, and most prominently in conv3 (layer 34) -- where the intrinsic dimension is at its peak (see Fig. \ref{fig:id_resnet152}) -- the representation is organized around low-entropy hubs whose centers are low-$S$ images (blue line, and left a stack of hub images).
As a reference, we also report the entropy computed by shuffling the neighbors' assignments (grey dashed line).
\subsection{The evolution of the probability density across the hidden layers}\label{sec:hier_nucl_evolution_cluster_ari}
We have seen that at layer34 (conv3)  all images are arranged in a high-dimensional hypersphere and that at layer 151 (conv5) the neighborhoods are already organized consistently with the classification.    
Clearly, the most important transformations of the representation happen in between these layers. 
To shed some light on the evolution of the representations in the intermediate layers, we use the approach described in Sec. \ref{sec:density_peak} that allows characterizing a multidimensional probability distribution, finding its probability peaks, and localizing all the saddle points separating the peaks. 
Unlike other methods for the analysis of hidden representations based on dimensional reduction  \cite{2017tsne_repr, 2015pca_repr}, here we do not perform any low-dimensional embedding of the data. 

We will see that the nucleation-like transition described in Sec. \ref{sec:nucl} is a complex process, in which the network separates the data in a gradually increasing number of density peaks laid out in a hierarchical fashion that closely mirrors the hyperonymous-hyponymous relations of the ImageNet data set.

\begin{figure}[t]
    \centering
\includegraphics[width=0.95\columnwidth]
{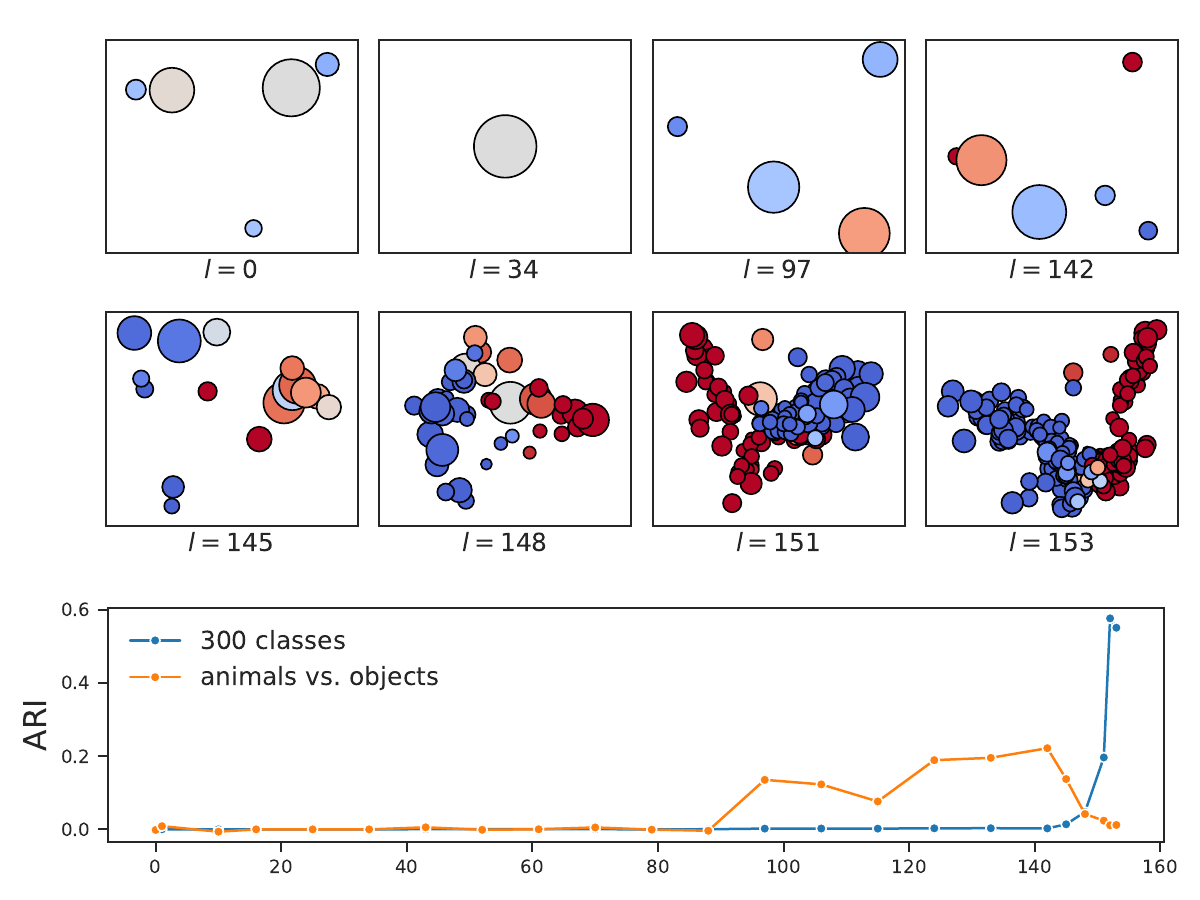}
    \caption{{
    \bf Structure and composition of density peaks of representations}. 
    ({\bf a}:) A schematic view of the peaks in 8 layers. Color tones refer to the relative presence of animals and artifacts in each peak: dark red = $100\%$ of animals, dark blue = $100\%$ of artifacts; the area of the circles is proportional to the number of data points in a cluster.
    ({\bf b}:) $ARI$ profiles for animal/artifact partition and the 300 low-level classes (blue and orange). 
    \label{fig:dendogram}
    }
\end{figure}

\paragraph{Evolution of the probability peaks across the hidden layers.}
Figure~\ref{fig:dendogram}-a shows a two-dimensional visualization of the number and organization of the probability peaks of the representations in some of the layers.
More precisely, we plot the cluster coordinates using the two principal axes of the matrix $K_{\alpha, \beta} = e^{\nicefrac{-S_{\alpha, \beta}}{\sigma}}$ where $S$ is given in  Eq. \eqref{eq:similarity} and $\sigma$ is the average of the entries of $S$. 

In the input layer ($l$=0), the data are split into two major peaks, which roughly divide the training set into light and dark images. 
This structure is not useful for classification and is wiped out within the first 34 layers of the network. In conv3, the probability density becomes unimodal, consistent with the analysis of the previous section.
In the subsequent layers, the network creates a structure that is useful for classification, and in layer 97, a bimodal distribution appears. The other peaks shown in the figure are very small and retain only a few hundred data points each.
The same density peaks persist until layer 142, where $97\%$ of the images still reside in the two biggest ones.
Finally, after layer 142 the two large peaks break down into smaller ones representing individual classes.

In order to assess the population of the density peaks in terms of ground-truth categories, we use the Adjusted Rand Index ($ARI$) \cite{ari_hubert,ref-rand_indx, rand_indx} (see Sec. \ref{sec:hier_nucl_methods_ari}). 
Roughly speaking, $ARI$ is zero if the density peaks do not correspond to the reference partitions of the data set, and is one if they match it. 
In Fig. \ref{fig:dendogram}-b we plot the $ARI$ with respect to the high-level animal/artifact categories ($ARI^{macro}$, orange line), and with respect to the $300$ low-level classes we sampled ($ARI^{cl}$, blue line). 
From layer 97 to layer 142, artifacts and animals predominantly populate one of the two major peaks, increasing the $ARI^{macro}$ value to 0.22 while the correlation with low-level classes remains absent.
The following breakup of the peaks leads to a drop of $ARI^{macro}$ to 0 and to a concomitant sharp rise of $ARI^{cl}$ from 0 to 0.55, consistent with the nucleation mechanism detected by $\chi^{l,gt}$ and described in Sec. \ref{sec:nucl}. 
Moreover, some classes are separated before others (Fig.~\ref{fig:dendogram}-a, layer 145), consistently with the bimodality in the distribution of $\chi^{l,gt}$ observed in the bottom panels of Fig.~\ref{fig:ov_gt}-a.

Interestingly, many of the density peaks in the layers between 142 and 153 (i.e., during the nucleation transition) closely resemble the hierarchical structure of ImageNet.
For instance, in layer 148, one can find peaks corresponding to insects, birds, but also ships and buildings.
\begin{figure}
    \centering
    \includegraphics[width =\columnwidth]{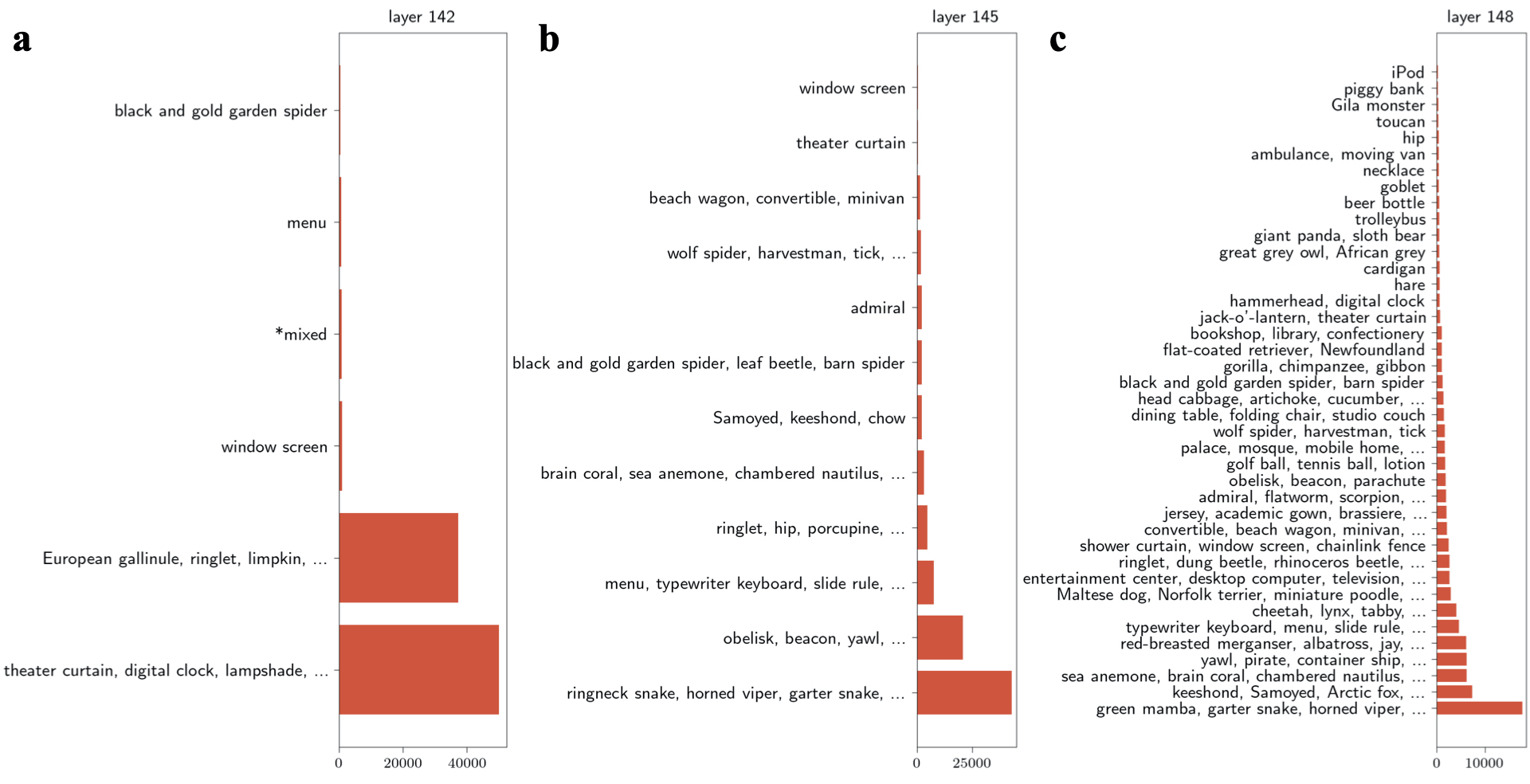}
    \caption{{\bf Composition of density peaks in layers 142 (\textbf{a}), 145 (\textbf{b}) and 148 (\textbf{c})}. The x-axis indicates the size of the peak, the y-axis reports the categories represented with more than 150 points in the peak. Consecutive dots (\q{...}) indicate that more than three categories are well represented in the peak. The peaks are ordered from the smallest to the largest from top to bottom.}
    \label{fig:app_layers_142_148}
\end{figure}
In Fig. \ref{fig:app_layers_142_148} we report a visualization of the density peaks appearing during the \q{nucleation transition} of ResNet 152. 
In particular, the image shows the size and approximate composition of the peaks present in layers 142, 145, and 148.
As discussed in Section 3.3, in layer 142 the data density is dominated by two large peaks composed of images of animals and artifacts, respectively.
This structure is visible in panel (a), in which one can easily identify the two large peaks.
In the subsequent layers, the animal and artifact peaks break down into small peaks containing images of the same class.
The process happens hierarchically: peaks corresponding to multiple classes sharing a lot of semantic similarities appear first, and subsequently break down into smaller peaks corresponding to the single classes.
This phenomenon can be observed in panels (b) and (c).
In layer 145 (panel (b)), one can identify peaks corresponding to certain kinds of arachnids (wolf spider, harvestman, tick, ...), insects (black and gold spider, leaf beetle, barn spiders) 4-wheel means of transportation (beach wagon, convertible, minivan), dogs (Samoyed, keeshond, chow), and so on.
In layer 148 (panel (c)), this process continues, and one finds many more peaks, corresponding either to single classes (e.g., iPod, piggy bank, and beer bottle) or to groups of similar classes.
At the end of the nucleation process described, from layer 152 (not shown here) one finds approximately one peak corresponding to each class label.

\paragraph{Probability landscape at the output of the network}
In the last layer ($l$ = 153), the different peaks correspond to the different classes, but the structure of the probability density is much richer than a simple collection of disjointed peaks. Indeed, the hierarchical process that shaped the density landscape across the layers leaves a footprint on the organization of the peaks.
For instance, the division in macro-classes of animals and artifacts formed in layer 97 is still present in the last layer as indicated by the fact that red and blue clusters are found primarily on the left and on the right of the corresponding low-dimensional embedding (Fig.~\ref{fig:dendogram}-a).
But much more structure is present.
\begin{figure}
\includegraphics[width=\textwidth]{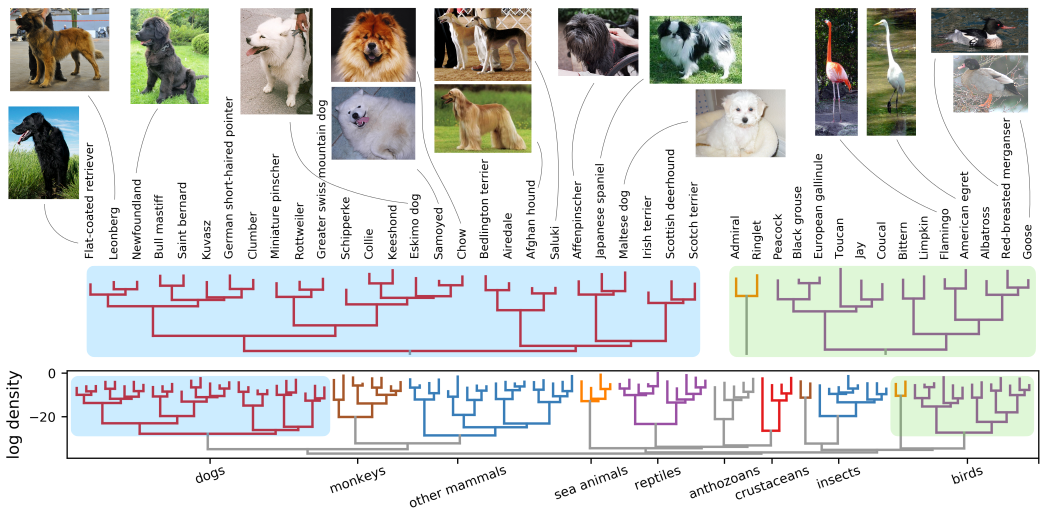}
    \caption{
    {\bf Structure and composition of density peaks of representations}. 
    The dendrogram portrays the hierarchical connections between the density peaks of the animal branch. On the $y$-axis the value of the density peaks is plotted in logarithmic scale. Two insets above show the detailed composition of specific branches (light blue and light green). 
    \label{fig:dendogram_b}
    }
\end{figure}
In Figure \ref{fig:dendogram_b} we visualize the probability landscape of the animal classes as a dendrogram, in which each leaf corresponds to a peak.
Since the leaves do not always contain a single class we consider just those where the most represented category has more than 150 data points. This procedure eliminates 10 small peaks out of the final 229. In the remaining peaks, the average frequency of the most populated class is 89.9\%, so most of the peaks can be fairly assigned to a single class.

The leaves are then merged sequentially, following the WPGMA algorithm \cite{sokal1958statistical}, according to the height of
the saddle point of the probability density between them (see Sec. \ref{sec:hier_nucl_methods_ari}). 
In this manner, the secondary probability peaks belonging to the same large-scale structure form a branch of the dendrogram. The height of a leaf in Fig. \ref{fig:dendogram_b} is proportional to the logarithm of the density of the peaks.
The morphological similarities of animals with similar genetic material make it possible for the dendrogram in Fig.~\ref {fig:dendogram_b} to reproduce the taxonomy of a phylogenetic tree to an astonishing degree.
At the root of the dendrogram, we can notice the first distinction between mammals on the left and other animals on the right.
At the following hierarchical level, we can find a more specific separation of animal types.
Dogs, reptiles, birds and insects, and so on can be easily identified.
Finally, within each species, say dogs (Fig. \ref{fig:dendogram_b}), alike breeds are linked by tighter bounds, that is, saddle points of higher density.

\subsection{Probability landscape of other deep image models}
\label{sec:hier_nucl_clusters_alex_vgg_vit_igpt}
We conclude this section by extending the analysis of the density landscape to other convolutional models (AlexNet and VGG19) and to attention-based models (Vision Transformer and iGPT). We will see that while in AlexNet and VGG19 the development of the ordered nuclei mirrors the trend we described for ResNet152 in the Vision Transformer the transition to an ordered structure is more gradual across the hidden representations. The architectures used the PyTorch pre-trained models available at \href{https://pytorch.org/vision/stable/models.html}{https:// pytorch.org/vision/stable/models.html}.  We now describe the main aspects of all the models. We refer to Sec. \ref{sec:imagenet_igpt} for the discussion of iGPT.

\paragraph{AlexNet}
The first example we consider is AlexNet, the architecture that won the Large Scale Visual Recognition Challenge held in 2012 with a top5 error on the ImageNet test set of 16.4\% and outscoring the runner-up (26.4\% top5 error) by 10 points (\href{https://image-net.org/challenges/LSVRC/2012/results.html}{https://image-net.org/challenges/LSVRC/2012/results.html}). 
This was a breakthrough leading to the widespread adoption of deep learning as the go-to choice in the computer vision community. 
The major advances brought by AlexNet were possible thanks to the use of GPUs for training and the effective implementation of the ReLU activation function, dropout, and data augmentations \cite{lecun2015deep, krizhevsky2012imagenet}.

\begin{figure}
\centering
{\includegraphics[width=1.\columnwidth]{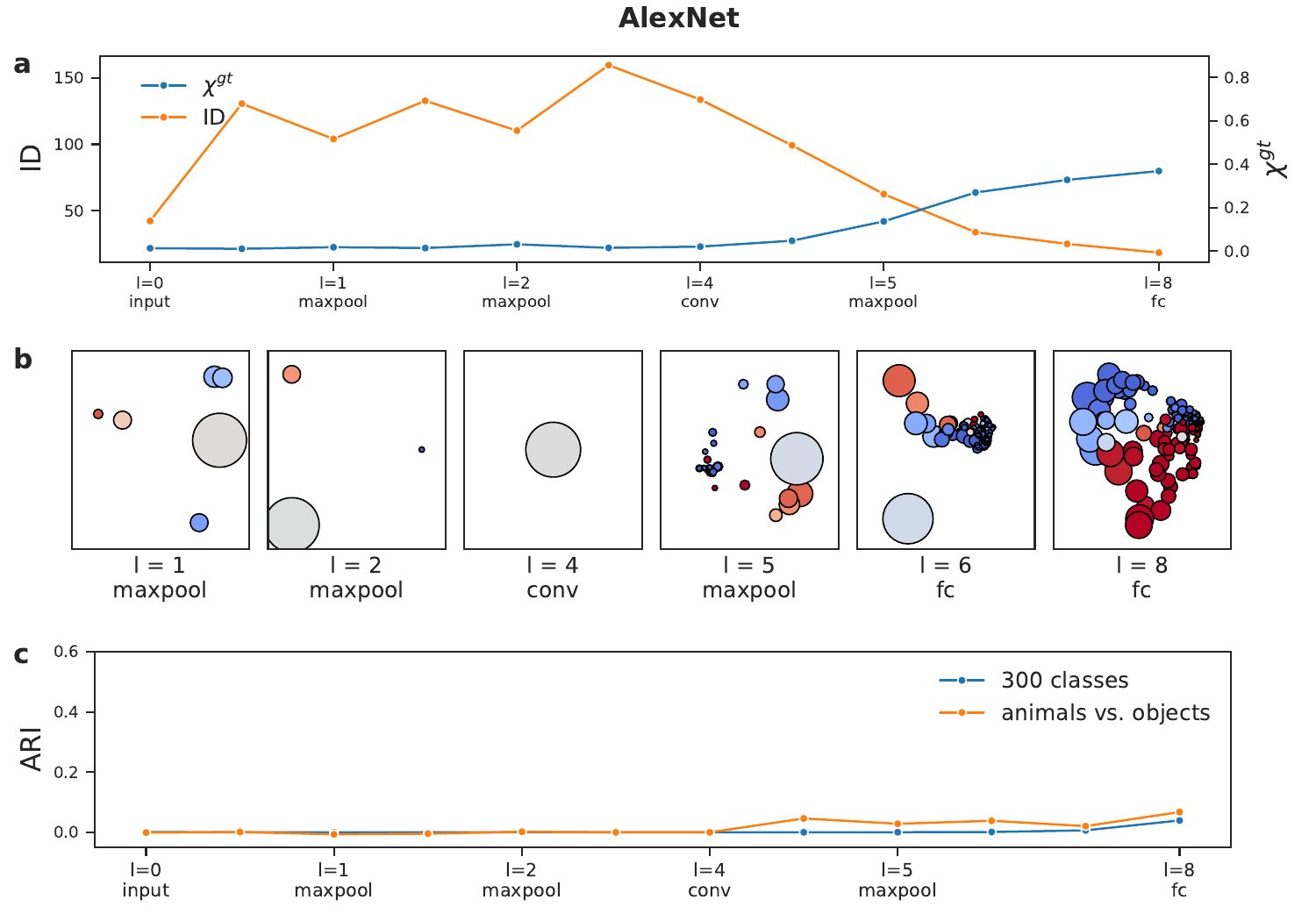}
	\caption{{\bf Characterization of density landscape of the hidden representations of AlexNet}:
	{\bf a}: Intrinsic dimension (orange) and overlap with 300 classes ($\chi^{gt}$, blue) of the hidden representations of AlexNet. The representations are the output of the convolutional, pooling, and final fully connected layers. The ID is measured with Gride ($k_1 = 2, k_2 = 4$).
	{\bf b}: Illustration of the size and composition of the density peaks of AlexNet. The position of the peaks is determined with the kernel PCA method described at the beginning of Sec. \ref{sec:hier_nucl_evolution_cluster_ari}. The area of the circles is proportional to the number of data points in the cluster, and the color tones refer to the relative presence of animals and artifacts in each peak: dark red = 100\% of animals, dark blue = 100\% of
    artifacts.
	{\bf c}: ARI profiles with animals/objects partition (orange) and 300 classes (blue)}
\label{fig:alexnet_clusters}}
\end{figure}

The architecture is composed of eight layers, five convolutional and three fully connected. The convolutional and fully connected layers (except the last one) are followed by a ReLU activation function.
The convolutional part is divided into three blocks separated by pooling layers; the first two blocks contain one convolution and the last block three. In Fig. \ref{fig:alexnet_clusters} we consider the representations at the output of convolutional, pooling, and fully connected layers.
The pre-trained PyTorch network reaches a top5/top1 accuracy of 79.1\%/56.5\% respectively. In the following, we will compare the performance of the network with the top1 accuracy since, as of 2022, this is by far the most used metric to evaluate the classification accuracy on
ImageNet.

\paragraph{VGG19 with batch normalization}
The AlexNet accuracy can be increased by a large margin just by making the architectures deeper \cite{vgg, 2014googlenet, 2015inceptionv3}. 
This is the main contribution of VGG19, made possible by the implementation of a $3 \times 3$ kernel in \emph{all} the layers in place of larger and more computationally expensive kernels.
The convolutional part of VGG19  has five convolutional blocks instead of three, containing $(2, 2, 4, 4, 4)$ convolutions each \cite{vgg}.
VGG19 is one of the best non-residual convolutional networks reaching 72.4\% top1 accuracy. 
When equipped with batch normalization \cite{ioffe2015batch}, the accuracy of VGG19 reaches 74.2\%. We analyze this latter version and extract the representations after the pooling layers, after the last three fully connected layers, and after some intermediate convolutional layers.

\paragraph{Vision Transformer}
With the advent of AlexNet and, importantly, residual networks  \cite{resnet}, convolutional models became the most successful for images and, despite the recent revolution brought by transformers in NLP, until 2020, the attention mechanism was only used together with convolutional layers in computer vision tasks \cite{bello2021lambdanetworks, dosovitskiy2021vision_transformer}.
The Vision Transformer \cite{dosovitskiy2021vision_transformer} (ViT) is one of the first attempts to rely only on attention to classify images. 
We analyze a PyTorch version reaching 85\% top1 accuracy.

\begin{figure}[t]
\centering
{\includegraphics[width=1.\columnwidth]{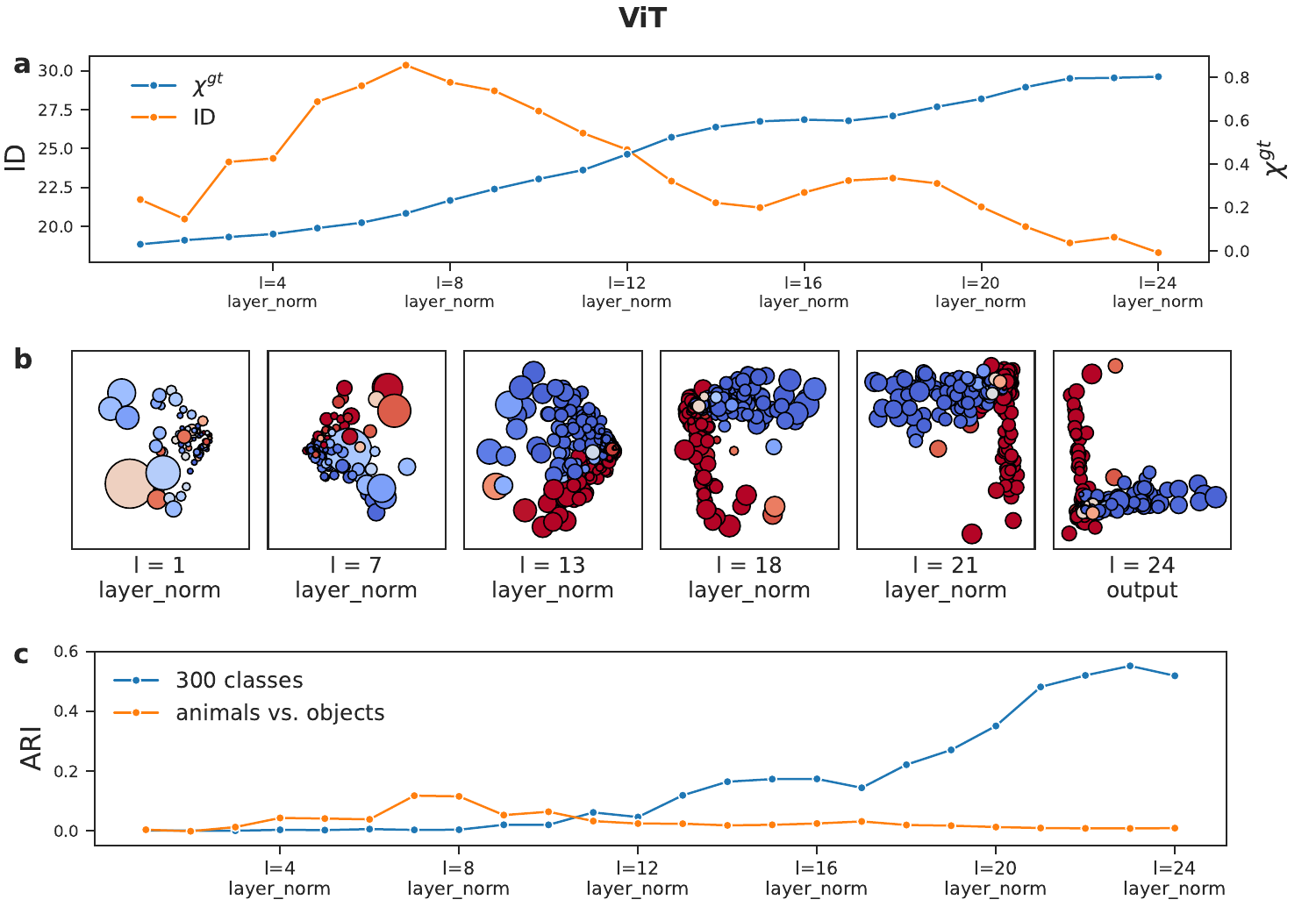}
	\caption{{\bf Characterization of density landscape of the hidden representations of the Vision Transformer}:
	{\bf a}: ID (orange) and $\chi^{gt}$ (blue) of the hidden representations of ViT. 
	The representations are the output of normalization layers before the attention blocks; only the "semantic token" is considered (see main text).
	{\bf b}: Illustration of the size and composition of the density peaks of ViT.
	{\bf c}: ARI profiles with animals/objects partition (orange) and 300 classes (blue)}
\label{fig:vit_clusters}}
\end{figure}

One of the major challenges that prevented the application of the transformers to images is the large memory cost required to store the attention maps for long sequences.
In Sec. \ref{sec:imagenet_igpt}, we saw that the iGPT model (contemporary to the ViT) reduces the sequence length by reducing the image resolution and encoding the color information in the embedding axis. 
ViT instead divides the $H^2$ input pixels into $s$ patches of size $16\times 16$. Each of the  $s = H^2/16^2$ patches is encoded in a 1024-dimensional vector by multiplying the $P = 3\times 16 \times 16$ pixels with a matrix of size $1024 \times P$, shared among the patches. 
A special token, itself of size 1024, is concatenated at the beginning of the sequence to encode the semantics of the representation and is used at the output of the network to predict the image class. Accordingly, the representation of a single image has a shape equal to $(s+1)\times 1024$, where $s = 196$ on the standard ImageNet resolution.
The architecture is composed of 24 attention blocks. 
In Fig. \ref{fig:vit_clusters} we analyze the representation of the first "semantic" token of the sequence, at the output of the normalization layer at the beginning of each block.
Another legitimate possibility would have been averaging over the sequence dimension as we did for the iGPT architectures in Sec. \ref{sec:imagenet_igpt}.
The size of a representation at the output of a block is equal to that at the input; therefore, the embedding dimensions $D$ of the data representations are all equal to 1024. 
This is very different from what happens in convolutional networks, where $D$ changes dramatically across the layers (see Sec. \ref{sec:id_resnet152}).

\paragraph{Intrinsic dimension and overlap profiles}
In this section as well as in Sec. \ref{sec:hier_nucl_dynamics_resnet152}, the ID is computed with Gride $(k_1 = 2, k_2 = 4)$, and the density is evaluated with the "Point Adaptive kNN estimator" (PAk) previously summarized in Sec. \ref{sec:pak}. PA$k$ automatically finds for each data point $i$ the optimal neighborhood size $k_i$ to evaluate the density and therefore reduces the number of hyper-parameters of our approach to the sole $Z$ (see  Eq. \eqref{eq:z_clusters}). We recall that $Z$ fixes the level of statistical confidence to consider two density peaks meaningful. We will set $Z = 1.6$ as we did in the previous section.

\begin{figure}
\centering
{\includegraphics[width=1.\columnwidth]{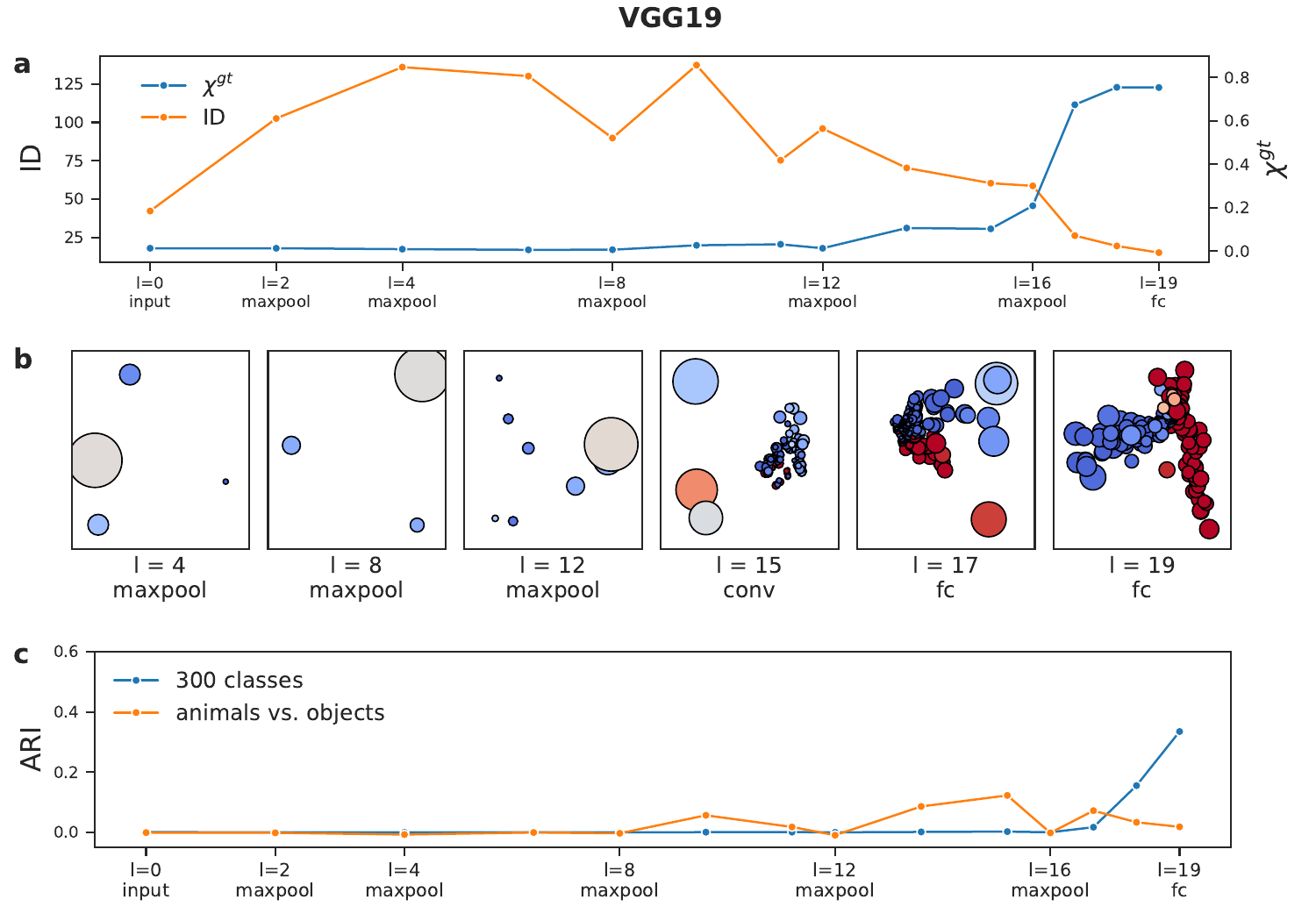}
	\caption{{\bf Characterization of density landscape of the hidden representations of VGG19}:
	{\bf a}: ID (orange) and $\chi^{gt}$ (blue) of the hidden representations of VGG19. The representations are the output of some convolutional layers, pooling layers, and final fully connected layers.
	{\bf b}: Illustration of the size and composition of the density peaks of VGG19.
	{\bf c}: ARI profiles with animals/objects partition (orange) and 300 classes (blue)}
\label{fig:vgg19_clusters}}
\end{figure}

\begin{figure}[t]
\centering
{\includegraphics[width=1.\columnwidth]{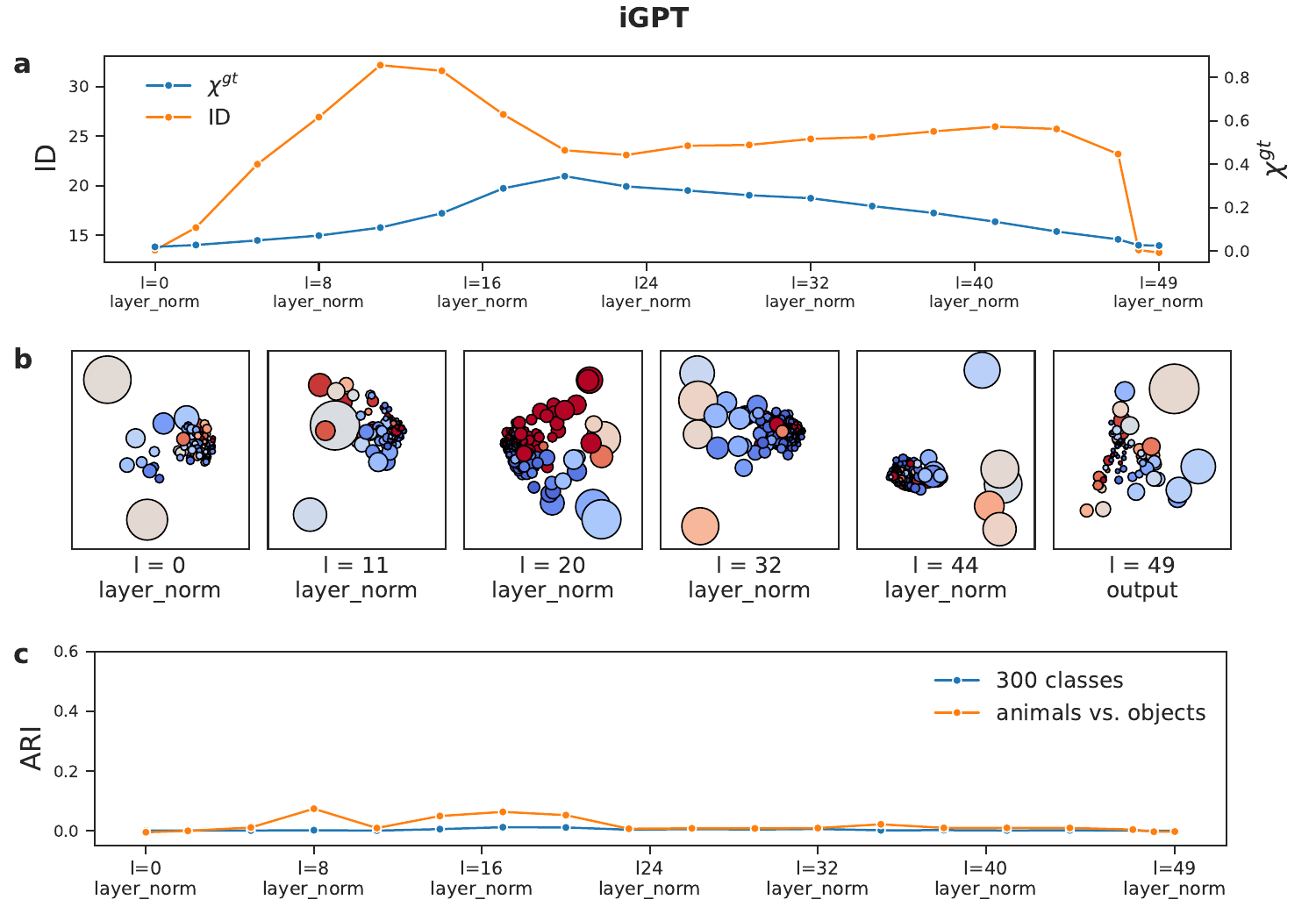}
	\caption{{\bf Characterization of density landscape of the hidden representations of iGPT}:
	{\bf a}: ID (orange) and $\chi^{gt}$ (blue) of the hidden representations of iGPT. The representations are the output of some normalization layers before the attention blocks.
	{\bf b}: Illustration of the size and composition of the density peaks of iGPT.
	{\bf c}: ARI profiles with animals/objects partition (orange) and 300 classes (blue)}
\label{fig:igpt_clusters}}
\end{figure}

In Figs. \ref{fig:alexnet_clusters}, \ref{fig:vgg19_clusters}, \ref{fig:vit_clusters}, and \ref{fig:igpt_clusters}, we show the ID and the overlap with the labels in the top panels ({\bf a}), a schematic view of the density peaks colored by the relative presence of animals and objects in the middle panels ({\bf b}), and the $ARI$ profiles for the animals/artifact and the 300 classes in the bottom panels ({\bf c}).

In all the models, the ID increases in the first layers to reach a peak in the first third of the architecture. For AlexNet, VGG19, and ViT, it decreases afterward, while for iGPT, after layer 22 increases again (see also Sec. \ref{sec:imagenet_igpt}).
Quite generally, when the ID decreases (after layer 4 in AlexNet, after layer 12 in VGG19, between blocks 8 and 14, and from block 19 to the end for ViT, between block 10 and 20 for iGPT) the overlap $\chi^{gt}$ grows and the neighborhoods become populated by data points of the same class. 
The growth of the overlap occurs rather abruptly towards the end of the network in AlexNet and VGG19, similar to what we saw for ResNet152, while in ViT, the increase is more gradual and seems to flatten around 0.58 between blocks 15 and 19, where the ID goes from 21 to 23.
In iGPT $\chi^{gt}$ grows in the first layers, more markedly after layer 10, to reach 0.35 at layer 20 (see Fig. \ref{fig:igpt_clusters}-a) despite the model being trained without labels.
We note that this value is comparable with the overlap observed at the output of AlexNet, which is equal to 0.37. After layer 20, $\chi^{gt}$ decreases and drops to 0 at the end of the network.

At the scale set by the first 30 nearest neighbors, the consistency with the labels at the end of the network in  VGG19, ResNet152, and ViT is quite similar. Indeed, the value of $\chi^{gt}$ is 0.80 and 0.75 for VGG19 and ViT (see Fig. \ref{fig:vgg19_clusters}-a and \ref{fig:vit_clusters}-a) consistent with the 0.83 for ResNet152 (see Fig. \ref{fig:ov_gt}).
A better assessment of the differences between these architectures, which perform quite differently from the ImageNet validation set, 
can be given by the analysis of the probability landscape of the hidden representations reported in panels {\bf b} and {\bf c}.


\paragraph{Density peaks analysis}
The evolution of the peaks across the layers in AlexNet and VGG19 is similar to that of ResNet152: until the end of the convolutional block, the density distribution is unimodal and the development of the peaks occurs between the last convolutional layers and the fully connected ones (from layer 5 onwards for AlexNet, from layer 13/14 onwards for VGG19). 
At the same time, we see that the $ARI$ with the labels at the output of the network are quite different: 0.05 and 0.33 for AlexNet and VGG19, which are smaller than the value observed in ResNet152 (0.44 with PA$k$ in our trained model, 0.55 with $k$NN in the pre-trained Pytorch model). 
In the hidden representations of AlexNet, both the $ARI$ profiles are very small because the clusters contain a mixture of ground truth classes.

In VGG19 instead, the values of the $ARI$ are larger and mirror the trend observed in ResNet152: after layer 12, the consistency with the macro classes grows to reach its maximum value at layer 15, where there are two clusters of size 29480 and 42225 containing 70\% of animals, the first, and 57\% of objects, the second. 
The action of the maxpool operation seems to "destroy" the consistency with the macro classes but the decrease of the $ARI$ is just due to the fact the two peaks are brought closer to each other by the maxpool operation (without mixing the populations of the clusters) and the density peak algorithm merges them in a single one (at $Z=1.6$).
In the last three fully connected layers, these macro clusters are progressively split in smaller ones, and the $ARI$ with the 300 classes grows to 0.33. 
At the output, we found 280 clusters, 114 clusters have a population between 200 and 400 data points, and with the most represented class covering, on average, 87\% of the points. 
These medium-sized clusters contain around 33 thousand data points. The remaining 57 thousand are mostly part of larger peaks that have not been split and include multiple classes; only 2 thousand data points are contained in clusters of smaller size.

The picture is quite different in the case of the Vit transformer, where immediately after the peak of the ID, the density distribution becomes multi-modal with some peaks already populated by images belonging to the same class.
This stark difference with the convolutional models seen so far is partly explained by the fact that, already after a few blocks, the number of non-linear layers is large, as each block contains the attention map plus two fully connected layers. 
Another reason is the low ID. When the ID is low, the difference between the density maxima and the density of the saddle points is more often significant, and more density modes are found.

After block 7, the $ARI$ with the animal/object reaches the highest value of around 0.11. At this stage, two clusters of size 7000 and 11000 containing 80\% and 90\% of animals split from the rest, but 40000 data points lie in a large peak containing a mixture of objects and animals.
The relatively high degree of mixture explains why, until layer 12 the $ARI$ remains low despite locally 
$\chi_{gt}\sim 0.4$ indicating that 40\% the first $30$ neighbors belong to the same class. 
This ordering mechanism is also present in ResNet152, where at layers $(145, 148, 151)$ $\chi^{gt}$ =  (0.2, 0.5, 0.7) to settle down at 0.83 at layer 153, while $ARI$ grows slightly later (0.02, 0.12, 0.4) to finally reach 0.6. 
Between blocks 13 and 18,  16\% of the data points lie in more than 150 clusters where the average consistency with the animals/objects partition is above 98\%, and many of them are also consistent with respect to the partition in 300 classes.

At the final layer, there are around 605 clusters, more than twice those found in VGG19 and ResNet152.
There are 172 clusters with a population between 200 and 400, where the most common class is represented by 88\% of the data points.
These clusters cover around 49 thousand data points, but differently from what happened for VGG19, more than $\sim$ 12 thousand other data points lie in smaller clusters with sizes between 50 and 200, where the class consistency is still around 86\%.
This high fragmentation of the output representation also explains why the $ARI$ coefficient with the classes ($\sim 0.55$) is comparable to that of the Pytorch ResNet152 itself, around 0.55. 
In ResNet152, the number of peaks is 354, closer to the actual number of classes, but on average, the classes are more mixed inside each peak.

In iGPT, the $ARI$ profiles are essentially zero in all the layers, meaning that, according to the class partition, the cluster assignments are random. 
Around layer 20, however, the overlap is 0.35 and the local composition of the neighborhoods is "ordered".
This local ordering is, to some extent reflected by a homogeneous composition at the animals/objects level inside the middle-sized clusters: the large majority of clusters with sizes between 100 and 3000 have a homogeneity higher than 95\% with the animals/objects partition.

\section{Comparison between linear probes and density-based clustering}
\label{sec:linear_probe}
The cross-entropy loss imposes a pressure \cite{2017linear_probes} to the last layer to make the representation linearly separable with respect to the class partition.
To quantify the amount of linear separability "back-propagated" to intermediate representations, \citet{2017linear_probes} propose to probe the performance of a linear classifier on the hidden features of a network. 
Linearly probing a representation allows evaluating how informative a hidden representation is with respect to some external objective.
From a geometrical perspective, the pressure of the cross-entropy loss forces the data points of a class to cluster as close as possible to the vertex of a high-dimensional simplex. The coordinates of the vertices are the one-hot encoding of the classes.
Our approach is therefore complementary to that of \citet{2017linear_probes} as it measures until which stage of the network the density modes of the last representation "back-propagate" before becoming mixed one with another. 

In this paragraph, we directly compare the linear probe accuracy with our approach for the case of ResNet152 and the ViT.
\begin{figure}[t]
\centering
{\includegraphics[width=0.9\columnwidth]{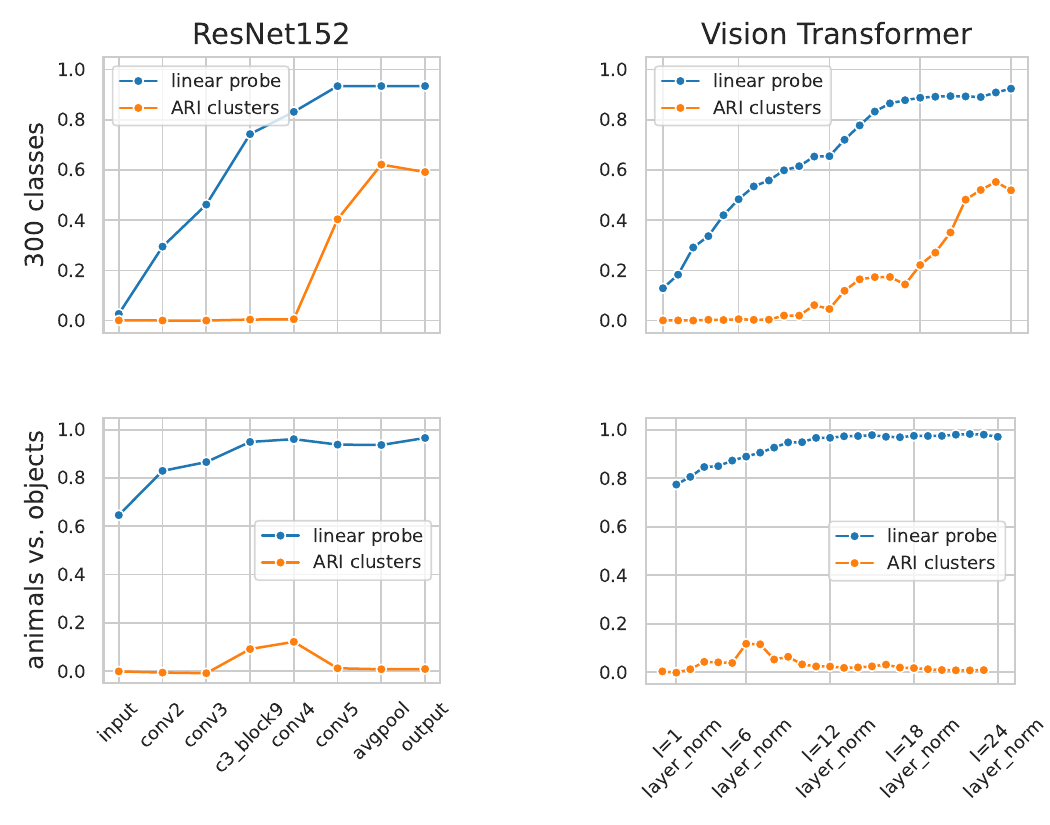}
	\caption{ {\bf Comparison between linear probe accuracy and ARI} Figure shows the accuracy of a linear classifier (blue) trained with the features of some hidden representation on the 300 classes (top) and animals/objects (bottom). In the same panels, we report the ARI for the same partitions (orange). On the left panels, we show the results of the  ResNet152 representation on the right of the ViT representations.}
\label{fig:linear_probes}}
\end{figure}
As in \cite{2017linear_probes}, we attach to the hidden representation of ResNet152 (of size $C \times H^2$) a linear fully connected layer, after averaging the $H^2$ activations of each channel.
In ViT, we take the 1024 activations of the first token of the sequence. 
We evaluate the linear separability of the same data set where we performed the cluster analysis (see Sec. \ref{sec:hier_nucl_data}), splitting the 90000 images in a training and validation set keeping 5/6 of the images for training and 1/6 for validation.
We train the linear classifier for 30 epochs with weight decay equal to 0.0001, an initial learning rate equal to 0.1 in ResNet152 and 0.01 in ViT, and annealing the learning rate with a cosine schedule.
In Fig. \ref{fig:linear_probes} we compare the linear probe validation accuracy (blue) and the $ARI$ of the hidden representations (orange)
for the two ground truth partitions analyzed in this chapter (300 classes, top, and animals vs. objects bottom).

Consistently with  \citet{2017linear_probes}, we find that the accuracy increases almost linearly across the layers of both partitions.
In ResNet152, the linear probe accuracy on the 300 classes reaches its maximum only at the end of the network, after layer 151 (indeed, the output of conv5, its average pool, and the output logits provide equivalent sets of features and must produce the same classification accuracy). 
The accuracy of the animals/objects partition instead saturates around layer 124 ("c3\_block9").
In both cases, the maximum accuracy is reached when the $ARI$ profile becomes for the first time different from 0.
In ViT, the asymptotic accuracy is reached around block 18 for the 300 classes and much earlier, around block 11, for the animals/ objects partition, and the stage at which $ARI$ starts to grow occurs some blocks before (at blocks 7 and 13, respectively).

The two methods assess the information content of a hidden representation in different ways.
Thank to the use of labels, the linear probe approach can divide the data even when the density is uniform.
On the other hand, density peak clustering can detect clusters only if the density maxima are separated by regions of sufficiently low density, and it finds some structure in the later layers of the network.
A piece of additional information given by our approach consists of the identification of the layer where the "information" about the macro classes is maximal and the representation is not yet split in fine-grained structures (e.g. "c3\_block9").
The same information is hard to retrieve from the linear probe analysis, as when the accuracy reaches its asymptotic value, all the layers towards the output are identical in a deep classifier.

More importantly, another crucial difference is that the linear probe approach is an instance of supervised training, and without labels, the method is blind.
In our approach, labels are used \emph{post hoc} to determine the agreement of the structures we find with some external partition of the data. 
If the labels were not available, the information given by the topography of the peaks, their number, sizes, and hierarchical organization (see Sec. \ref{sec:hier_nucl_methods_ari}) could be used to infer how explicit a given representation is according to some unknown distribution of the data. 
For instance, we saw that in the last layer of VGG19 two two-thirds of the probability mass was contained in very large clusters, while in the Vision Transformer, the same probability was distributed in
moderately large peaks with a number of data points between 50 and 400.

\section{Dynamics of probability landscape in ResNet152}
\label{sec:hier_nucl_dynamics_resnet152}
\paragraph{Density landscape of an untrained network.} We now look at how the density peaks arise during training, starting from the study of the density landscape of a randomly initialized ResNet152. 
This analysis serves to set a baseline for comparison with the case of a trained model, but it is by itself a valuable inspection, as randomly initialized CNNs are known to provide 
good classification priors (see \citet{Saxe2011_random_weights}).

\begin{figure}
\centering
{\includegraphics[width=0.75\columnwidth]{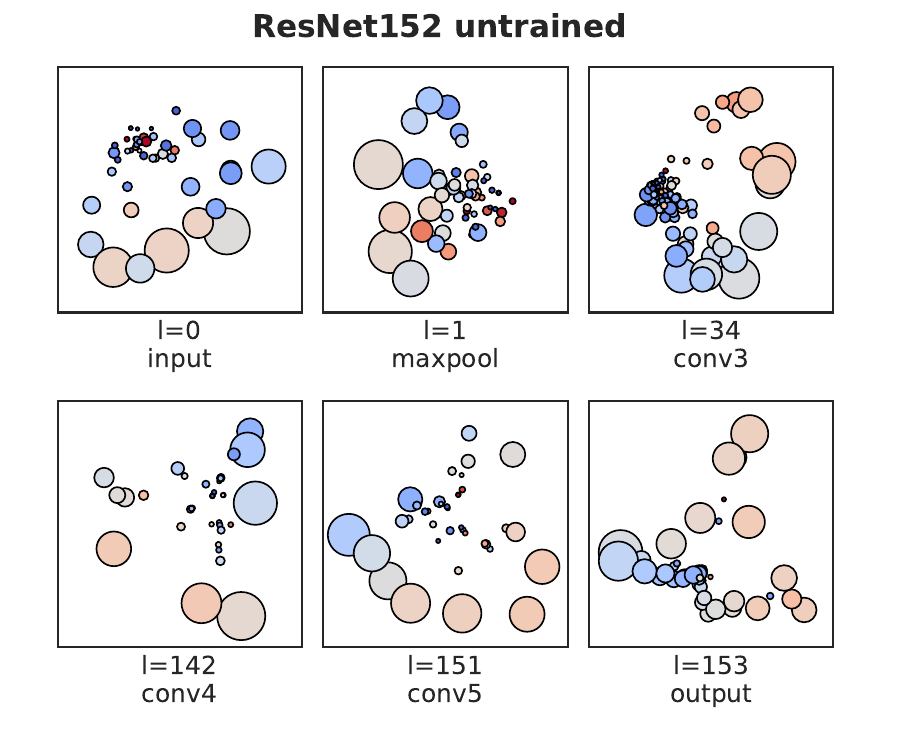}}
	\caption{{\bf visualization of the density peaks in an untrained ResNet152}: 
	The position of the peaks is determined with the kernel PCA method described at the beginning of Sec. \ref{sec:hier_nucl_evolution_cluster_ari}. The area of the circles is proportional to the number of data points in a cluster, and the color tones refer to the relative presence of animals and artifacts in each peak: dark red = 100\% of animals, dark blue = 100\% of artifacts. We report the visualization of the peaks at the input layer and at the output of some of the ResNet152 blocks.
	}
\label{fig:resnet152_untrained}
\end{figure}

When the network is not trained, most of the probability mass (> 85\%) is contained in large clusters with more than 1000 data points  (see Fig. \ref{fig:resnet152_untrained}).
The number of peaks is of the order of 50/100 for all the layers analyzed, and for the first two blocks is larger than what is found when the network is trained.
This is a consequence of the relatively low ID (<40, see Sec. \ref{sec:id_resnet152}) of all the hidden representations.
At the same time, in the last two blocks after layer 142,  the number of clusters is much smaller than what is observed when the network is trained.
We also note that when the density is estimated with PA$k$, the number of peaks is higher than what is found setting $k = 30$. In Sec. \ref{sec:hier_nucl_evolution_cluster_ari} we found four peaks at the input layer, two of which contained most of the images (see Fig. \ref{fig:dendogram}). 
With PA$k$ we find instead 49 peaks, four of which contain more than 10 thousand images while most of the others contain just a few images (see top left panel of Fig. \ref{fig:resnet152_untrained}).
The degree of separation in PA$k$ is higher because the optimal neighborhood size $k^*$ is smaller, with an average ranging from 3 to 20.
With a larger $k$, some relevant density fluctuations are smoothed out, leading to a less structured probability landscape. 

The number of peaks is also affected peculiarly at some specific stages of the network. Where the pooling operations severely shrink the size of channels (between layer 0 and 1 where $H\times H$ goes from $224 \times 224$ to $56 \times 56$, and between layer 151 and layer 152 where the last average pool reduces the $7 \times 7$ channels to a set real numbers) a large portion of the probability mass ($\sim 20\%$) flows from very large clusters containing more than 10 thousand images to smaller ones and the number of clusters increases. 
Between layer 34 and layer 142, there is a long sequence of convolutions where the size of the channels remains constant and the number of density peaks decreases from 150 to 50.

\begin{figure}
\centering
{\includegraphics[width=0.45\columnwidth]{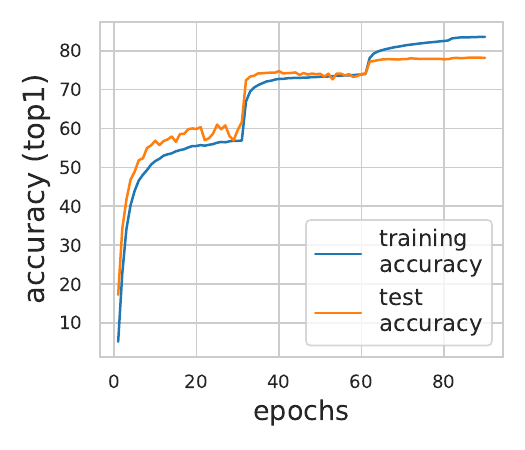}}
	\caption{{\bf Training and test error dynamics in ResNet152}: Profiles of the training accuracy (blue) and test accuracy (orange) over the 90 training epochs. The final test accuracy is 78.8\%. The training protocol is described in Sec. \ref{fig:id_dynamics_resnet152}. The jumps in accuracy correspond to the epochs (30, 60, and 80) where the learning rate is decreased by a factor of 10.
	}
\label{fig:training_curves_resnet152}
\end{figure}

\paragraph{Evolution of the density landscape during training.} 
The sharpest changes during training occur in the first epochs. We train the ResNet152 for 90 epochs following the procedure described in Sec. \ref{sec:id_resnet152}.
At epoch 10 the training accuracy is already 57\%, and the remaining 80 epochs are used to adjust the learned weights to gain an additional 20\% (see Fig. \ref{fig:training_curves_resnet152}). 
In Sec. \ref{fig:id_dynamics_resnet152}, we also showed the largest changes of the ID occur in the first 13 epochs.

\begin{figure}
\centering
{\includegraphics[width=0.98\columnwidth]{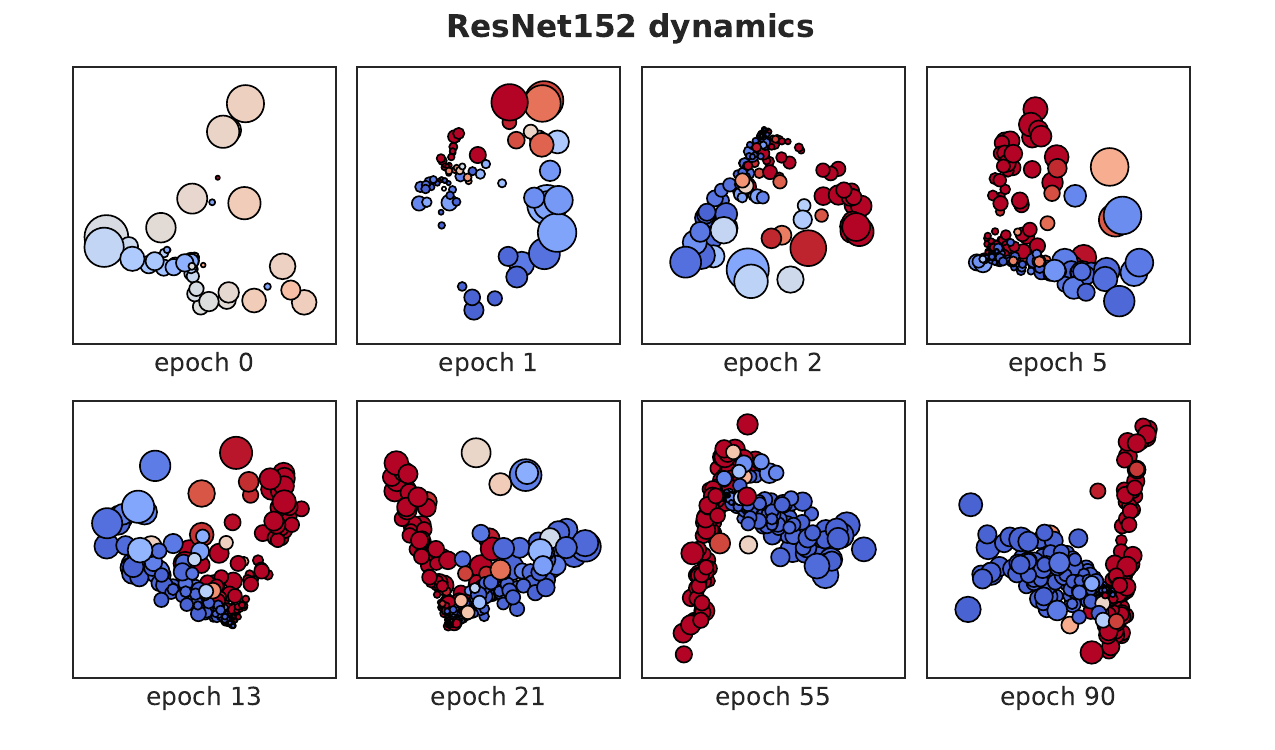}}
\caption{{\bf Dynamics of the density landscape at the output layer of ResNet152}: we display a schematic view of the distribution and size of the density peaks at some training steps, shown below each panel.
	The position of the peaks is determined with the kernel PCA method described at the beginning of Sec. \ref{sec:hier_nucl_evolution_cluster_ari}. The area of the circles is proportional to the number of data points in the cluster, and the color tones refer to the relative presence of animals and artifacts in each peak: dark red = 100\% of animals, dark blue = 100\% of artifacts.
	}
\label{fig:resnet152_dynamics}
\end{figure}

In Fig. \ref{fig:resnet152_dynamics} we show instead the evolution of the density distribution at the output of the network. 
To ease the analysis, we divide the peaks into four groups $(G1, G2, G3, G4)$ based on the number of data points they contain using the following thresholds: $(10000, \allowbreak  500, \,20)$.
The first group, $G1$, contains peaks with more than 10 thousand points, and the second, $G2$, contains peaks with more than 500 points and less than 10 thousand, and so on.
As the training proceeds, the total number of peaks increases from 57 to 394, the probability mass flows from the larger clusters to the smaller ones and the average difference between the density of the peaks and that of the saddle points $\overline{\delta}$ becomes larger.

Before the training begins, the number of data points contained in the four groups is 
$(67\%, \; 21\%,  \;4\%, \; 8\%)$.
At epoch 2, the total number of peaks increases to 137, and around 40\% of the probability mass of $G1$ flows to $G2$.
At epoch 13, $G1$ does not contain any more peaks, and the populations of four groups are $(0\%, \; 54\%, \; 33\%, \;10\%)$ of the total number of points. 
At this stage $\overline{\delta}$ stops to grow and more and more peaks separate, without changing substantially the difference between the density peaks and that of the saddle points. 
After epoch 13, $G3$ grows, draining points both from $G2$ and $G4$ with a distribution of the probability mass after 55 epochs of $(0\%, \; 21\%,  \;75\%, \; 4\%)$, with 16  and 235 clusters in $G2$ and $G3$.

\begin{figure}
\centering
{\includegraphics[width=0.45\columnwidth]{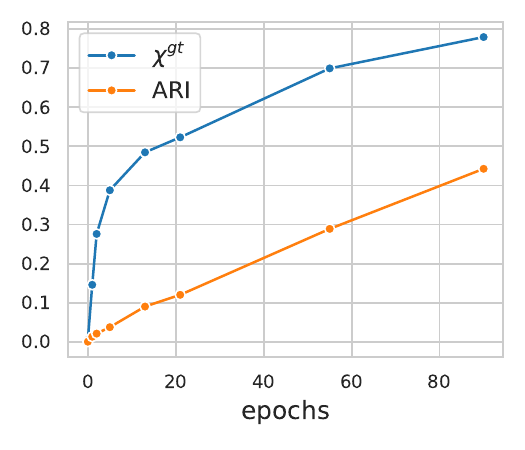}}
	\caption{ {\bf Dynamics of the density peaks at the output of ResNet152 measured by ARI and $\chi^{gt}$}:
	The blue profile shows the growth of $\chi^{gt}$ at the output of ResNet152 after some training epochs. Accordingly, the orange profile shows the ARI with the 300-class partition. $\chi^{gt}$ grows rapidly in the first 13 epochs while ARI increases linearly with the training epochs.}
\label{fig:dynamics_ari}
\end{figure}

The $ARI$ with the animals/objects partition remains low throughout the training as a consequence of the high fragmentation of the probability landscape, but already after the first epoch, the overlap ($k=30$) with the animals/objects partition reaches 90\%, as can also be intuitively seen from the color tones of Fig. \ref{fig:resnet152_dynamics}.
The $ARI$ with the 300 categories instead grows linearly as a function of the training epochs (see Fig. \ref{fig:dynamics_ari}) to reach 0.45 at the end of the training, while the overlap mostly grows in the first 13 epochs. 
The process that leads to the final collections of ordered peaks therefore first populates the neighborhoods with alike data points and then gradually splits the peaks, once at a local scale, the consistency with the ground truth labeling is already high.

\section{Discussion}\label{sec:hier_nucl_discussion}
In this chapter, we provided an explicit characterization of the evolution of the probability density of the data landscape across the hidden representations of various deep learning models trained on ImageNet. 
We showed that this probability density undergoes a sequence of transformations that bring to the emergence of a rugged and complex probability landscape. 
In trained convolutional models, rather surprisingly, we found that the development of these structures is not gradual, as one would expect in a deep network with more than one hundred layers.
Instead, the greatest changes to the neighborhood composition and the emergence of the probability peaks happen in a few layers close to the output (Figs. \ref{fig:dendogram} and \ref{fig:vgg19_clusters}-c).
The analysis of Vision Transformer reveals instead a more regular development of the final probability modes, which appear already  (Fig. \ref{fig:vit_clusters}-c) in the middle of the architecture.

Importantly, however, the external validation analysis of the clusters using the ARI (Sec. \ref{sec:hier_nucl_results}) and the linear probe analysis (Sec. \ref{sec:linear_probe}) show that the hidden representations of both CNNs and the ViT first capture general concepts, for instance, the distinction between animals and objects, and then learn more detailed semantic features.  
The knowledge of the general concepts is not destroyed when the fine-grained categories are learned, as in the logit space, a hierarchical clustering of the density peaks allows recovering the semantic relationships between the classes (Fig. \ref{fig:dendogram_b}).

We finally studied how the density peaks arise during training by looking at the probability density in the logit representation (Sec \ref{sec:hier_nucl_dynamics_resnet152}).
We showed that hierarchical learning occurs also as a function of time, first recognizing animals and objects (Fig. \ref{fig:training_dynamics}) and only gradually developing the final peaks once the neighborhoods of each data are mostly populated by alike data points (Fig. \ref{fig:dynamics_ari}).

In \citet{bilal2017convolutional}, confusion matrices were used to visually analyze the correlations between classes showing results in agreement with our conclusions.
However, the algorithm we use here (Sec. \ref{sec:density_peak}) can reconstruct a probability landscape that faithfully follows the hierarchical structure of categories (Sec. \ref{sec:hier_nucl_evolution_cluster_ari}) in an unsupervised manner, with no need to consider the ground truth labels and estimate the confusion matrix. 
Indeed, our approach works also in the limiting case of 100\% test accuracy.

\paragraph{Difference between overlap and CKA}
Our picture seems qualitatively different from the one emerging from SVCCA \cite{svcca}, projection weighted CCA  \cite{morcos2018insights}, and linear CKA \cite{cka}, which have revealed smoother changes between nearby representations. 

\begin{figure}
\centering
{\includegraphics[width=0.9\columnwidth]{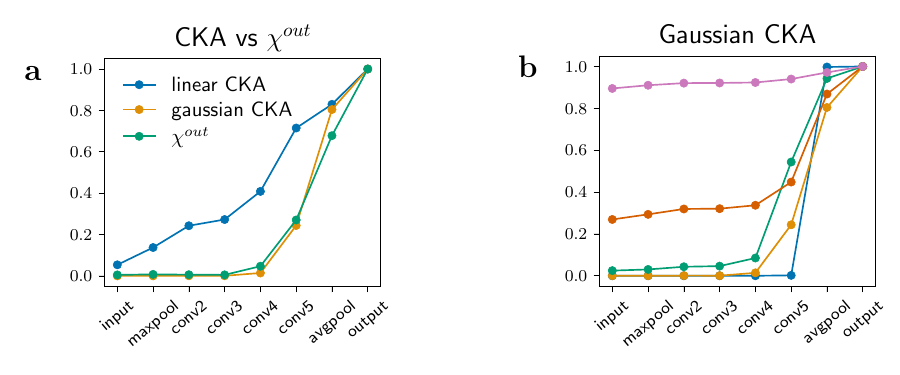}}
	\caption{({\bf a}:) Linear CKA (blu), overlap $\chi^{l, out}$ (green) and Gaussian CKA (orange) with the output layer in ResNet152 for a subset of 5000 ImageNet images. We kept 50 classes and 100 images per class and set the kernel bandwidth $\sigma$ to 0.2 times the average distance with the first nearest neighbor $\overline{d_1}$. ({\bf b}:) Gaussian CKA with the output layer as a function of the kernel bandwidth $\sigma$: $\sigma = 0.1\overline{d_1}$ (blu),  $\sigma = 0.2\overline{d_1}$ (orange),  $\sigma = 0.5\overline{d_1}$ (green),  $\sigma = \overline{d_1}$ (red),  $\sigma = 2\overline{d_1}$ (pink).}
\label{fig:app_cka}
\end{figure}

Centered kernel alignment (CKA) is one of the most used similarity indices between representations and estimates the similarity through a scalar product between two kernel matrices $K_l$ and $K_m$ evaluated on the two representations $l, m$. More precisely, after the data set has been centered, CKA evaluates \cite{cka}:
\begin{equation}
\label{eq:cka}
    CKA(K_l, K_m) = \dfrac{\tr (K_l K_m) }{\sqrt{\tr (K_l K_l) \tr (K_m K_m)}}
\end{equation}
Like the neighborhood overlap, it is invariant under orthogonal transformations and isotropic scaling, but not to an arbitrary invertible linear transformation. 
When the kernel is linear, CKA is equivalent to a CCA between representations in which the canonical variables are weighted by the corresponding eigenvalues \cite{cka} and Eq. \ref{eq:cka} reduces to $CKA(K_l, K_m) = \norm{Y_l^TX_l}^2/(\norm{X_l^TX_l}\norm{Y_l^TY_l})$.
In Gaussian CKA, the kernel $K_l(x_i, x_j) = e^{-\nicefrac{\norm{x_i-x_j}^2}{2\sigma_l^2}}$ probes the local similarity between representations and can be seen as a smoothing of the neighborhood overlap presented in Sec \ref{sec:ov}.
In Fig. \ref{fig:app_cka}-a we compare the similarity measured by linear CKA (blue), Gaussian CKA (orange), and the overlap (green), using as a reference layer the output and setting the kernel bandwidth to 0.2 times the average distance to the first nearest neighbor. 
Linear CKA steadily increases already in the early layers of the network (see Fig. \ref{fig:app_cka}-a blue profile).
Figure \ref{fig:app_cka}-b shows how the Gaussian CKA with the output is affected by different choices of the kernel bandwidth $\sigma$. The smaller $\sigma$, the sharper the transition measured by the index.

A difference between CCA, SVCCA, and linear CKA with the overlap lies in the kind of correlation captured by these similarity indices, the former being more focused on a large-scale correlation, the latter instead on a correlation between the neighborhoods of the data.
In Sec \ref{sec:hier_nucl_results} we saw that the ordering process starts with a "large scale" separation between animals and artifacts, which is functional and correlated to a successful fine-grained classification of the categories. In essence, CCA-based methods capture the correlation between the final categories (the peaks) and the \q{intermediate level} concepts (\q{the mountain chains}) required to construct them, which are recognized already in the middle layers of the network. 
The overlap defined in Eq. \ref{eq:overlap} measures instead a correlation growing only when, locally, the neighbors become consistent with those of the output ($\chi^{l,out}$) or their labels ($\chi^{l,gt}$). 

\begin{figure}
{\includegraphics[width=0.9\columnwidth]{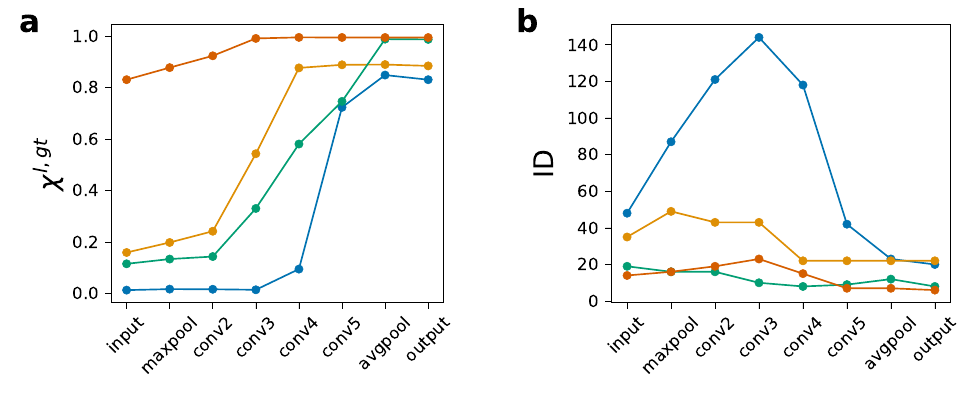}}
\caption{Overlap with the ground truth labels (a)  and intrinsic dimension profiles (b) in ResNet152 for different data sets: MNIST (red), CIFAR10 (orange), modMNIST (green), ImageNet (blue).}
\label{fig:app3}
\end{figure}

\paragraph{Interplay between nucleation transition and data set complexity.}
A second possible reason for the discrepancy between the results reported in this chapter and those based on linear correlation analysis can be attributed to the complexity of the data sets analyzed. %
Most previous studies have focused on data sets like MNIST and CIFAR10 which lack the semantic stratification of ImageNet and hence show a much smoother evolution of the probability landscape. 
This is a consequence of the fewer number of steps needed to disentangle the hierarchy of features of the categories. 
Indeed also the overlap shows a more gradual growth on easier data sets. In Fig. \ref{fig:app3} we compare  $\chi^{l, gt}$ and the ID of data sets of increasing complexity trained ResNet152: MNIST \cite{lecun2010mnist} modMNIST, CIFAR10 \cite{cifar10} and ImageNet \cite{imagenet09}. 
Since in MNIST alike digits are close to each other already in the input space \cite{mnist_knn}, we generated a more challenging version, modMNIST, resizing the dimension of the digits by a random factor ranging from 0.2 to 0.4 and moving them in a random location of the image, and finally scale up the size of the images to 224x224.
This way we decrease $\chi^{l,gt}$ from $\sim 0.78$ to $\sim 0.17$.

MNIST can be directly classified with high accuracy with a $k$NN estimator and consistently already in the input layer $\chi^{l,gt} \sim 0.78$ and reaches one from conv3 onwards.
In modMNIST and CIFAR10 $\chi^{l, gt}$, the initial value is larger than ImageNet because the number of categories is 10 instead of 1000, and in both cases, the overlap starts to grow from conv3 (layer 34) while on ImageNet, the local "disorder-to-order" transition occurs at the end of block conv4 (layer 142).
For complex data sets like ImageNet, the ID has the hunchback shape reported in \cite{id}, while for MNIST and modMNIST, it is almost constant, and it takes much smaller values.

\paragraph{Practical applications of the analysis of probability peaks.}
We believe that the detailed picture of the evolution of the probability density can improve the performance of learning protocols, make transfer learning more effective, and enrich the information extracted from a multi-class classifier.
For instance, the knowledge of the probability landscape can provide a rational criterion to improve training schemes based on triplet loss \cite{2015Facenet}. In this setting it is crucial to select challenging triplets where the so-called \q{anchor} image is closer to an example of a different class ({\it negative}) than to that of the same class ({\it positive}). 
Data points lying at the boundary of a probability peak could be used as anchor,s as they are on average closer to negatives than those lying close to cluster centers. 
One can also imagine defining training losses targeting the development of probability peaks according to a pre-defined semantic classification.  
This can be enforced in the intermediate layers of a network, going well beyond a simple disentanglement of the feature space \cite{2016corr_features}, enhancing the separation between macro categories which arise spontaneously. 
Moreover the analysis of the topography of the 
density peaks developed by a deep neural network can be used as a hierarchical classifier, going beyond the sharp classification in mutually exclusive categories \citet{deng2014large}.

An appropriate understanding of the nucleation mechanism could also be beneficial to transfer learning since it gives a simple rational criterion to judge the generality of the features of a representation \citet{yosinski2014transferable}.

\paragraph{Hierarchical learning as a form of "simplicity bias".}
The problem of the generalization of deep models is a very active area of work. 
Some recent lines of research observed that deep models have a tendency to learn simple patterns
~\citep{saad1995a, farnia2018spectral, valle-perez2018deep, kalimeris2019sgd, rahaman2019spectral} in spite of their very large capacity.

For instance, \citet{kalimeris2019sgd} showed that deep networks capture solutions that can be very well explained by linear classifiers in the first epochs and then they refine the fit to more complex functions while retaining the memory of the linear classifier learned in the initial stages. They then provide some experimental evidence on CIFAR-10. 
The hierarchical dynamics discussed in Sec. \ref{sec:hier_nucl_dynamics_resnet152} is connected with the 
he mechanism of "simplicity bias" described by \citet{kalimeris2019sgd}, as we also show that the ResNet152 first learns simpler (more general) concepts and then focuses on the classification of the detailed categories in the second part of training.
We hope that our analysis can serve as a first step to understanding the generalization of deep models through the lenses of the simplicity bias in a complex data set like ImageNet.

\paragraph{Looking ahead.}
In the next chapter, we look at generalization in deep neural networks, but rather than focusing on the simplicity of the solution found by SGD, we will approach the problem through the analysis of the redundancy of a representation. It is well known that ensembles of classifiers can improve the generalization performance and robustness of predictors. 
We will see that, implicitly, wide neural networks at the end of training develop different copies of features useful for the classification task instead of overfitting noisy, non-generalizable signals.

\chapter{Redundant representations help generalization in wide neural networks}
\label{ch:repr-mitosis}
\section{Introduction}

Deep neural networks (DNN) typically have enough parameters to achieve zero
training error, even with random labels~\citep{zhang2016understanding, arpit2017closer}.
In defiance of the classical bias-variance trade-off, the performance of these \emph{interpolating classifiers} continuously improves as the number of parameters increases well beyond the number of training
samples~\citep{geman1992neural, Neyshabur2015, spigler2018jamming,
  nakkiran2020deep}.
Despite recent progress in describing the implicit bias of stochastic gradient descent towards ``good'' minima~\citep{gunasekar2018characterizing, gunasekar2018implicit, Soudry2018, ji2019implicit, arora2019implicit, chizat2020implicit}, and the detailed analysis of solvable models of learning~\citep{
advani2020highdimensional, neal2018modern, mei2019generalization, belkin2019reconciling, hastie2022surprises, dascoli2020double, adlam2020understanding, lin2021causes, geiger2020scaling}, the mechanisms underlying this ``benign overfitting''~\citep{bartlett2020benign} in \emph{deep} neural networks remain unclear, especially since their loss landscape contains ``bad'' local minima and SGD can reach them~\citep{liu2020bad}.

\begin{figure*}[!t]
  \centering
   \includegraphics[width=1.\textwidth]{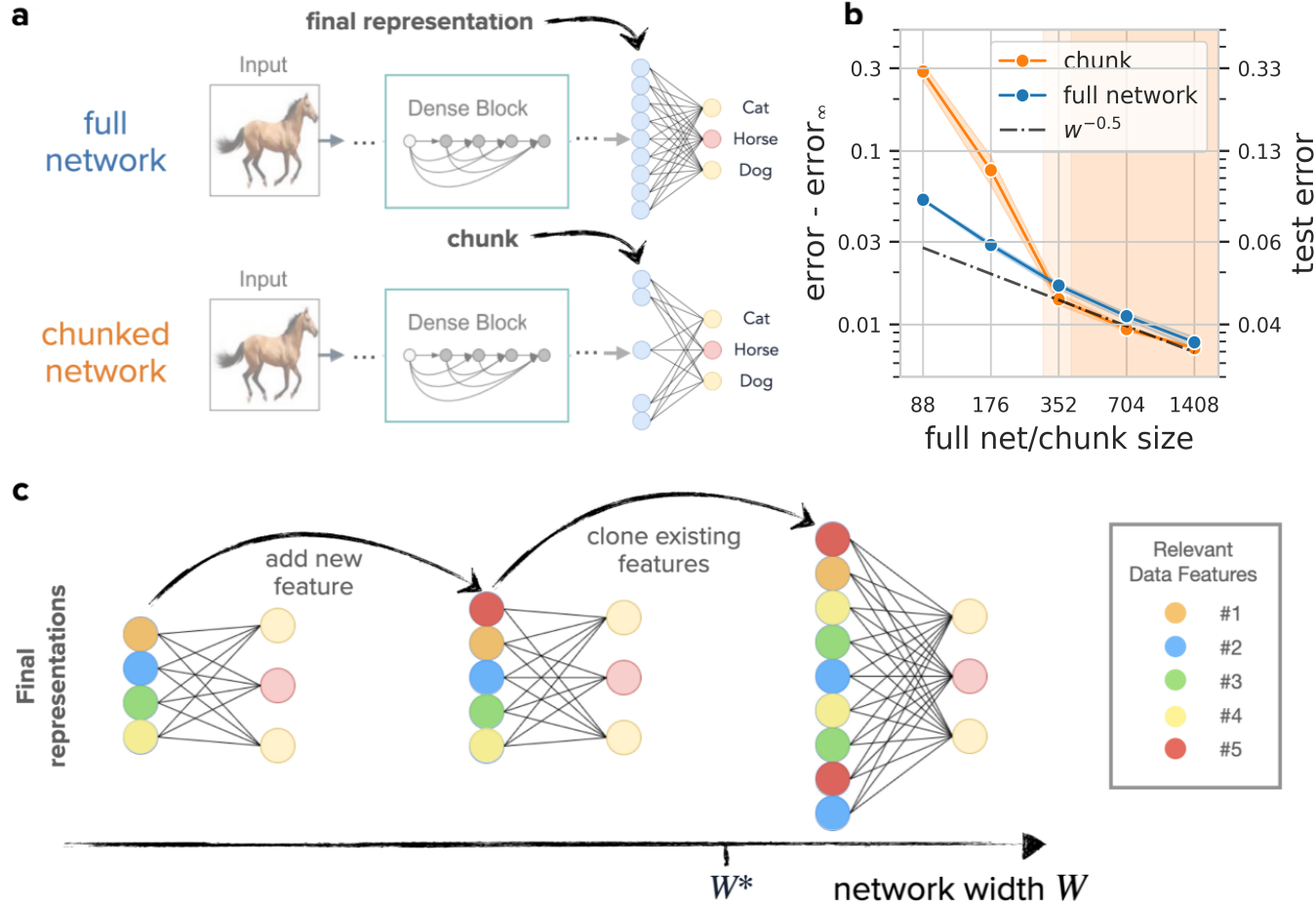}
  \caption{\label{fig:cartoon_} \textbf{The redundancy of representations in wide neural networks.} \textbf{a}: We analyze the final representations of
    deep neural networks (DNN),  namely the activities of the last hidden
    layer of neurons (light blue)
    We focus on
    the performance and the statistical properties of randomly chosen subsets of $w_c$ neurons, which we call ``\emph{chunks}''. In the chunked network shown here, $w_c=5$ out of 9 neurons are kept and used to predict the output.
    \textbf{b}: 
    As we increase the size of the chunk $w_c$ that we keep in a state-of-the-art DNN, here a DenseNet40, the test error of the chunk (orange line) becomes similar to the test error of a full network of width $W=w_c$ (blue line). In this regime, which is reached when $w_c$ is larger than a threshold $w_c^*$ (shaded area) the error approaches its asymptotic value $\mathrm{error}_\infty$ as a power-law $w_c^{-1/2}$ (dashed line). $\mathrm{error}_\infty$ is the error of an ensemble average of 20 networks of the widest size.
    \textbf{c}: Illustration of three final representations for networks of increasing width. In small networks, an additional neuron fits new features of the data
    (red neuron). As the network width goes beyond a critical width $W^*$,
    additional neurons instead copy features already learned from data, instead of over-fitting to features that are not relevant to the task. This mechanism is
    suggested by the $w_c^{-1/2}$ decay of the chunk error, and by the
    statistical analysis, we present in this paper.}
\end{figure*}

In this chapter, we describe a phenomenon observed in wide DNNs that offers a possible mechanism for benign overfitting. 
We illustrate this mechanism in Fig.~\ref{fig:cartoon_} for a family of increasingly wide DenseNet40s~\citep{huang2017densely} on CIFAR10~\citep{krizhevsky2009learning}. 
For simplicity, we refer to the width $W$ of the last hidden representation as the width of the network. 
The blue line in Fig.~\ref{fig:cartoon_}-b shows that the average classification error
($\texttt{error}$) approaches the performance of a large ensemble of  networks ($\texttt{error}_\infty$) \citep{geiger2020scaling} as we increase the network width $W$. In agreement with \citep{zagoruyko2016wide}, we find that the performance of these DenseNets improves continuously with width. 
For widths larger than 350, the networks are wide enough to reach zero training error
(see Fig. \ref{fig:app_densenet}-c) and, interestingly,  their test error decays approximately as $W^{\nicefrac{-1}{2}}$. Our goal is to understand how the error of the network can keep decaying beyond the interpolation threshold, and why it decays as $W^{\nicefrac{-1}{2}}$.

We make our key observation by performing the following experiment: we randomly
select a number $w_c$ of neurons from the last hidden layer of the widest Dense-Net40, and remove all the other neurons from that layer as well as their connections (Fig. \ref{fig:cartoon_}-a). 
We then evaluate the performance of this ``chunk'' of $w_c$
neurons, \emph{without} retraining the network. The orange profile of Fig. \ref{fig:cartoon_}-b shows the test error of chunks of varying
sizes.
There are two regimes: for small chunks, the error decays faster than $w_c^{\nicefrac{-1}{2}}$, while beyond a critical chunk size $w_c^*$ (shaded area), the error of a chunk of $w_c$ neurons is roughly the same as the one of a full network with
$w_c$ neurons. Furthermore, the error of the chunks decays with the same
power-law $w_c^{\nicefrac{-1}{2}}$ beyond this critical chunk size.

The decay rate of $\nicefrac{-1}{2}$ suggests that in this regime chunks of  $w_c$ neurons can be thought of as statistically independent estimators of the same features of the data, differing only by small, uncorrelated noise. 
In other words, beyond the critical width
$w_c^*$, the final hidden representation of an input in a trained, wide DNN, becomes highly redundant.
This motivates a possible mechanism for benign overfitting, schematically portrayed in Fig. \ref{fig:cartoon_}-c: as the network
becomes wider, additional neurons are first used to learn new features of
the data. 
Beyond the critical width~$w_c^*$, additional neurons in the final layer don't fit new features in the data, and hence over-fit; instead, they make a copy, or a \emph{clone}, of a feature that is already part of the final representation. The last layer thus splits into more and more clones as the network grows wider, as we illustrate at the bottom of
Fig.~\ref{fig:cartoon_}. The accuracy of these wide networks then improves with
their width because the network implicitly averages over an increasing number of clones in its representations to make its prediction. 

This paper provides a quantitative analysis of this phenomenon on various data sets and architectures. Our main findings can be summarized as follows:

\begin{enumerate}
\item A chunk of $w_c$ random neurons of the last hidden representation of a wide neural network predicts the output with an error that decays as $w_c^{-1/2}$ if the layer is wide enough and $w_c$ is large enough. In this regime, we call the chunk a ``clone'';
\item  Clones can be linearly mapped one to another, or to the full representation, with an error that can be described as uncorrelated random noise.
\item Clones appear if the training error is zero and the model is trained with weight decay.
If training is stopped too early or if the training is performed without regularization, 1.~and 2.~do not take place, even if the last representation is very wide. 
\end{enumerate}

\section{Methods}
\label{sec:methods}

\subsection{Neural network architectures}
\label{sec:methods-nn}

We report experimental results obtained with several architectures (fully
connected networks, Wide-ResNet-28, DenseNet40, ResNet50) and data sets
(CIFAR10/100~\cite{krizhevsky2009learning},
ImageNet~\cite{imagenet_cvpr09}). 
%
%
All our experiments are run on Volta V100 GPUs. In the following paragraphs, we describe our training protocols and hyperparameters, as well as how we vary the width $W$ of the architectures.
We report the validation accuracy of each architecture in table \ref{table:accuracy}.

\paragraph{Fully-connected networks on MNIST.}
We train a fully-connected network to classify the parity of the MNIST
digits~\citep{lecun1998} (pMNIST) following ~\citet{geiger2020scaling}. The MNIST digits are projected on the first ten principal
components, which are then used as inputs of a five-layer fully-connected
network (FC5).
The four hidden representations have the same width $W$ and the output is a real number whose sign is the predictor of the parity of the input digit.

We train the fully-connected networks for 5000 epochs with stochastic gradient descent using the following hyperparameters: batch size = 256, momentum = 0.9, learning rate = $10^{-3}$, weight decay = $10^{-2}$.
We optimize our networks using Adam.

\paragraph{Wide-ResNet-28 and DenseNet40 on CIFAR10/100.} CIFAR10 and CIFAR100 are instead trained on family of
Wide-ResNet-28~\citep{zagoruyko2016wide} (WR28). 
The number $W$ of the last hidden neurons in a WR28-$n$ is $64 \cdot n$, obtained after average pooling the last $64 \cdot n$ channels of the network. In our experiments, we also analyze two narrow versions of the standard WR28-1, which are not typically used in the literature. We name them  WR28-0.25 and WR28-0.5 since they have 1/4 and 1/2 of the number of channels of WR28-1.
Our implementation of DenseNet40 follows the DenseNet40-BC variant described in \citet{huang2017densely}. The width of the architecture is determined by the number of channels $c$. We consider a set of architectures with $c = \{8, \:16, \:32, \: 64, \:128, \:256\}$. The number $W$ of the last hidden features of this architecture is $5.5\cdot c$. 

All the models are trained for 200 epochs with stochastic gradient descent with a batch size = 128, momentum = 0.9, and a cosine annealing scheduler starting with a learning rate of 0.1. The training set is augmented with horizontal flips with 50\% probability and random cropping the images, padded with four pixels on each side.  
The weight decay $wd$ on CIFAR10 is $5 \cdot 10^{-4}$ for all the WR28 and Densenet40-BC architectures.
On Densenet40-BC we kept the label smoothing $ls$ equal to 0.05 for all the architectures, while for WR28 we vary $ls$ according to the width setting: $ls =0.1$ for WR28-$\{0.25, 0.5, 1, 2\}$ and $ls = 0$ for WR28-$\{4, 8\}$. 
%
%
On CIFAR100, we set $wd =  \{10, \: 7,\: 5, \: 5, \: 5\}\cdot 10^{-4}$ and $ls = \{0.1, \; 0.07, \; 0.05, \: 0, \: 0\}$ for WR28-$\{1, 2, 4, 8, 16\}$ respectively.
All the hyperparameters were selected with a small grid search.

\paragraph{ResNet50 on ImageNet.} We modify the ResNet50 architecture~\citep{he2016deep} by multiplying by a constant factor $c \in \{0.25, 0.5, 1, 2, 4\}$ the number of channels of all the layers after the input stem. 
When $c = 2$, our networks differ from the standard Wide-ResNet50-2 since~\cite{zagoruyko2016wide} only double the number of channels of the bottleneck of each ResNet block. 
As a consequence in our implementation, the number of features after the last pooling layer is $W = 2048 \cdot c$ while in~\cite{zagoruyko2016wide} $W$ is fixed to 2048. 

We train all the ResNet50 models with mixed precision
\citep{micikevicius2018mixed} for 120 epochs with a weight decay of
$4\cdot10^{-5}$ and label smoothing rate of 0.1 \citep{bello2021revisiting}.
The input size is $224\times 224$ and the training set is augmented with random crops and horizontal flips with 50\% probability. The per-GPU batch size is set to 128 and is halved for the widest networks to fit in the GPU memory. 
The networks are trained on 8 or 16 Volta V100 GPUs to keep the batch size $B$ equal to 1024. The learning rate is increased linearly from 0 to 0.1$\cdot B$/256 \citep{goyal2017accurate} for the first five epochs and then annealed to zero with a cosine schedule.

\begin{table}[h!]
  \caption{Validation accuracy for the architectures analyzed in this chapter (average over four runs)}
  \label{table:accuracy}
  \centering
\begin{center}
  \begin{tabular}{llllll}
    \toprule
    \multicolumn{2}{c}{CIFAR10} &     \multicolumn{2}{c}{CIFAR100} &  \multicolumn{2}{c}{ImageNet (top1)}  \\
    \cmidrule(r){1-6} 
    network     & acc. & network  & acc. & network  &  acc. (top1)\\
    \midrule
    Wide-RN28-0.25 & 84.1 &    Wide-RN28-1  & 70.4  &     RN50-0.25 & 67.0   \\
    Wide-RN28-0.5 & 90.3 &     Wide-RN28-2  & 75.7  &     RN50-0.5  & 74.1 \\
    Wide-RN28-1 & 93.4 &       Wide-RN28-4  & 79.6  &     RN50-1  & 77.6 \\    
    Wide-RN28-2 & 95.2 &       Wide-RN28-8  & 80.8  &     RN50-2  & 79.1 \\ 
    Wide-RN28-4  & 95.9  &     Wide-RN28-16 & 81.9 &      RN50-4  & 79.5 \\ 
    Wide-RN28-8  & 96.1  \\
    DenseNet40-BC (k=8)  & 91.6  \\
    DenseNet40-BC (k=16)   & 93.9  \\
    DenseNet40-BC (k=32)   & 95.1 \\
    DenseNet40-BC (k=64)   & 95.7 \\
    DenseNet40-BC (k=128)   & 96.0 \\ 
    
    \bottomrule
  \end{tabular}
  \end{center}
  
\end{table}

\subsection{Analytical methods}
\label{sec:methods-analysis}

\paragraph{Reconstructing the wide representation from a smaller chunk.}
To determine how well a subset of $w$ neurons can reconstruct the full representation of size $W$ we search for the $W \times w$ linear map~$\mathbf{A}$, able to minimize the squared difference 
${({\mathbf{x}}^{(W)}-\hat{\mathbf{x}}^{(W)})}^2$ between the $W$ activations of the full layer representation, $\mathbf{x}^{(W)}$, and the activations predicted from the chunk of size $w$, $\hat{\mathbf{x}}^{(W)}$:
\begin{equation}
    \label{eq:fit}
    \hat{\mathbf{x}}^{(W)} = \mathbf{A} \mathbf{x}^{(w)}.
\end{equation}
This least-squares problem is solved with ridge regression \citep{hastie01statisticallearning} with regularization set to $10^{-8}$, and we use the $R^2$ coefficient of the fit to measure the predictive power of a given chunk size.
The $R^2$ value is computed as an average of the  single-activations $R^2$ values corresponding to the $W$ output coordinates of the fit, weighted by the variance of each coordinate.
We further compute the $W \times W$ covariance matrix $C_{ij}$ of the \emph{residuals} of this fit, and from $C_{ij}$ we obtain the correlation matrix as:
\begin{equation}
    \label{eq:correlation}
    \rho_{ij} = \frac{C_{ij}}{\sqrt{C_{ii}C_{jj}} + 10^{-8}},
\end{equation}
with a small regularization in the denominator to avoid instabilities when the standard deviation of the residuals falls below machine precision.
To quantify how much the errors of the fit are correlated, we average the absolute values of the non-diagonal entries of the correlation matrix $\rho_{ij}$.
For short, we refer to this quantity as a \emph{mean correlation}.

\paragraph{Reproducibility.} The code to reproduce our experiments and our
analysis is available at~\url{https://github.com/diegodoimo/redundant_representation}.

\section{Results}
\label{sec:results}
\subsection{The hallmarks of redundant representations.}\label{sec:hallmarks_redundancies}
\begin{figure}
  \centering
  \includegraphics[width=0.45\textwidth]{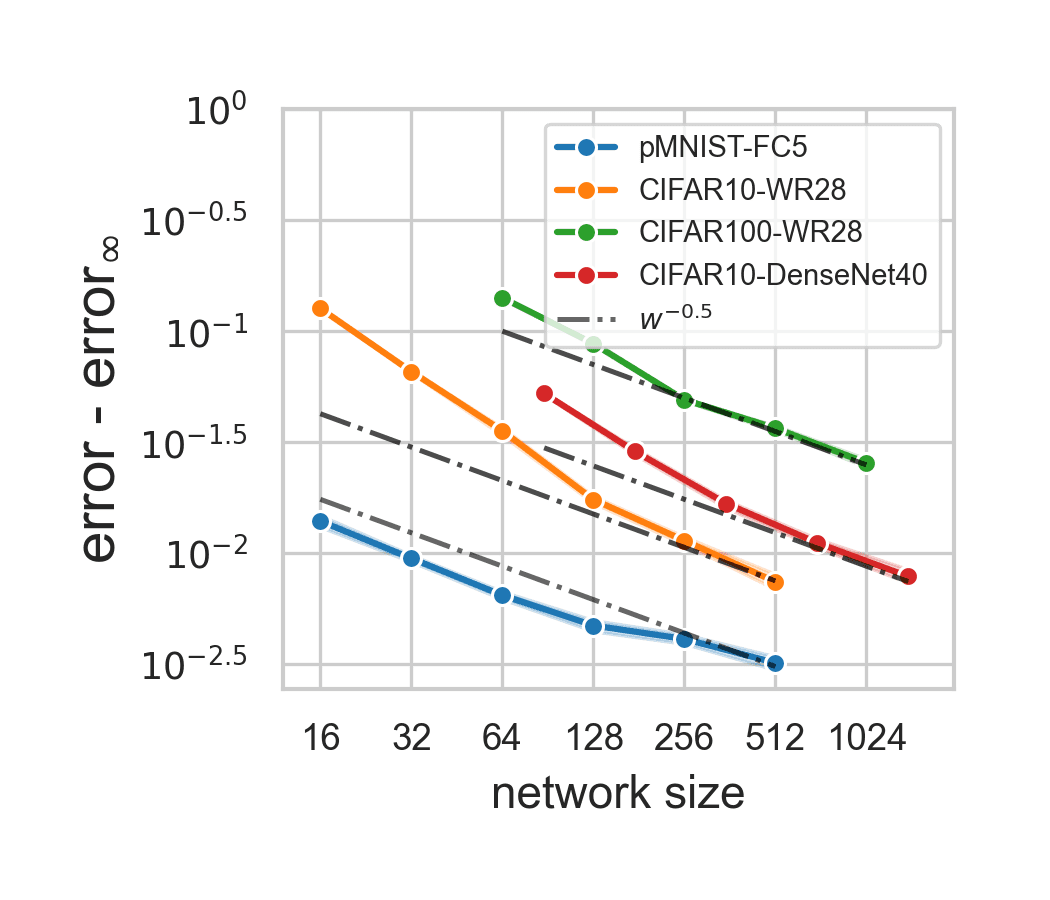}
\caption{\label{fig:error-full-network} 
      \textbf{Scaling of the test error with
      width for various DNNs.} The average test error of neural networks with various architectures approaches the test error of an ensemble of such networks as the network width increases. The network size shown here is the width of the final representation. For large width, we find a power-law behavior $\mathrm{error} - \mathrm{error}_\infty \propto W^{-1/2}$ across data sets and architectures. Full experimental details in Sec.~\ref{sec:methods-nn}}
\end{figure}

\paragraph{The test error of chunks of $w_c$ neurons of the final representation asymptotically scales as~$w_c^{-1/2}$.}
 The mechanism we propose is inspired by the following experiment: we compute the test accuracy of models obtained by selecting a random subset of $w_c$ neurons from the final hidden representation of a wide neural network. 
 We consider three different data sets: pMNIST, trained on a fully connected network, CIFAR10, and CIFAR100, trained on a convolutional network. The width $W$ of the architecture is 512 for pMNIST and CIFAR10,  and $W=1024$ for CIFAR100 (see Sec. \ref{sec:methods-nn}).
 In all these cases, $W$ is large enough to be firmly in the regime where the accuracy of the networks scales (approximately) as $W^{-1/2}$ (see Fig.~\ref{fig:error-full-network}). 
 We select $w_c$ neurons at random and we compute the test accuracy of a network in which we set to zero the activation of all the other $w-w_c$ neurons. 
Importantly, we do not fine-tune the weights after selecting the $w_c$ neurons: all the parameters are left unchanged, except that the activations of the ``killed'' neurons are not used to compute the logits.
We take 500 random samples of neurons for each chunk width $w_c$. 

\begin{figure*}[!t]
  \centering
  \includegraphics[width=1.\textwidth]{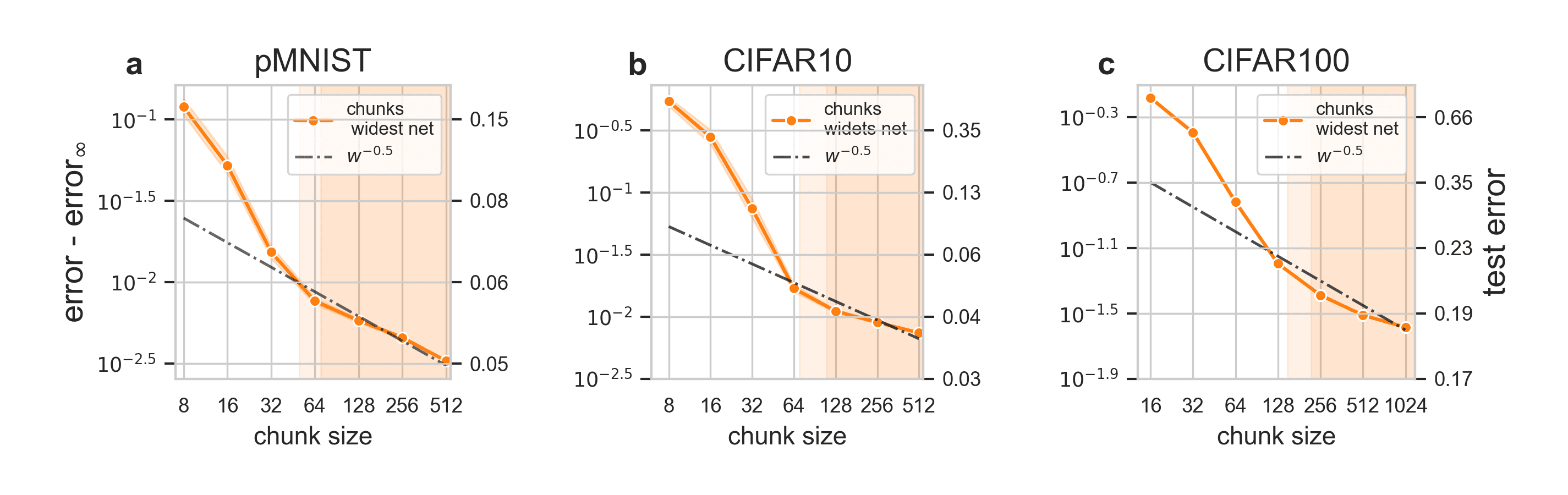}
  \caption{\label{fig:chunk-error-scaling} \textbf{Scaling of the test error of chunks of neurons extracted from the final representation of wide NNs.} We plot how the test error of chunked networks approaches $\mathrm{error}_\infty$, the error of an ensemble of 20 networks of the widest size (e.g. $W = 1024$ for CIFAR100), as the chunk size $w_c$ increases. Chunks are formed by selecting $w_c$ neurons at random from the final representation of the widest networks: an FC5 on pMNIST (width $W=512$), and Wide-ResNet-28 for CIFAR10 ($W=512$) and CIFAR100 ($W=1024$). The shaded regions indicate regions where the error of the chunks with $w_c$ neurons decays as $w_c^{-1/2}$.}
\end{figure*}

In  Fig.~\ref{fig:chunk-error-scaling}, we plot the test error of the chunked models as a function of $w_c$ (orange lines). The behavior is similar in all three networks: the test error of the chunks decays as  $w_c^{\nicefrac{-1}{2}}$ for chunks that are larger than a critical value $w_c^*$, which depends on the data set and architecture used.  
This decay follows the same law observed for full networks of the same width  (Fig.~\ref{fig:error-full-network}). This implies that a model obtained by selecting a random chunk of $w_c > w_c^*$ neurons from a wide final representation behaves similarly to a full network of width $W=w_c$. 
Furthermore, a decay with rate~$\nicefrac{-1}{2}$ suggests that the final representation of the wide networks can be thought of as a collection of statistically independent estimates of a finite set of data features relevant for classification. Adding neurons to the chunk hence reduces their prediction error in the same way an additional measurement reduces the measurement uncertainty, leading to the~$\nicefrac{-1}{2}$ decay.

\vspace{-0.0cm}

At  $w_c$ smaller than $w_c^*$ instead, the test error of the chunked models decays faster than $w_c^{\nicefrac{-1}{2}}$ in all the cases we considered, including the DenseNet architecture trained on CIFAR10 shown in Fig.~\ref{fig:cartoon_}-b.
In this regime, adding neurons to the final representation improves the quality of the model significantly quicker than it would in independently trained models of the same width (see Fig. \ref{fig:cartoon_}-c for a pictorial representation of this process).
We call chunks of neurons of size  $w_c \geq w_c^*$ \emph{clones}. 
In the following, we characterize more precisely the properties of the clones. 

\begin{figure*}[t]
    \centering
    \includegraphics[width=1.\textwidth]{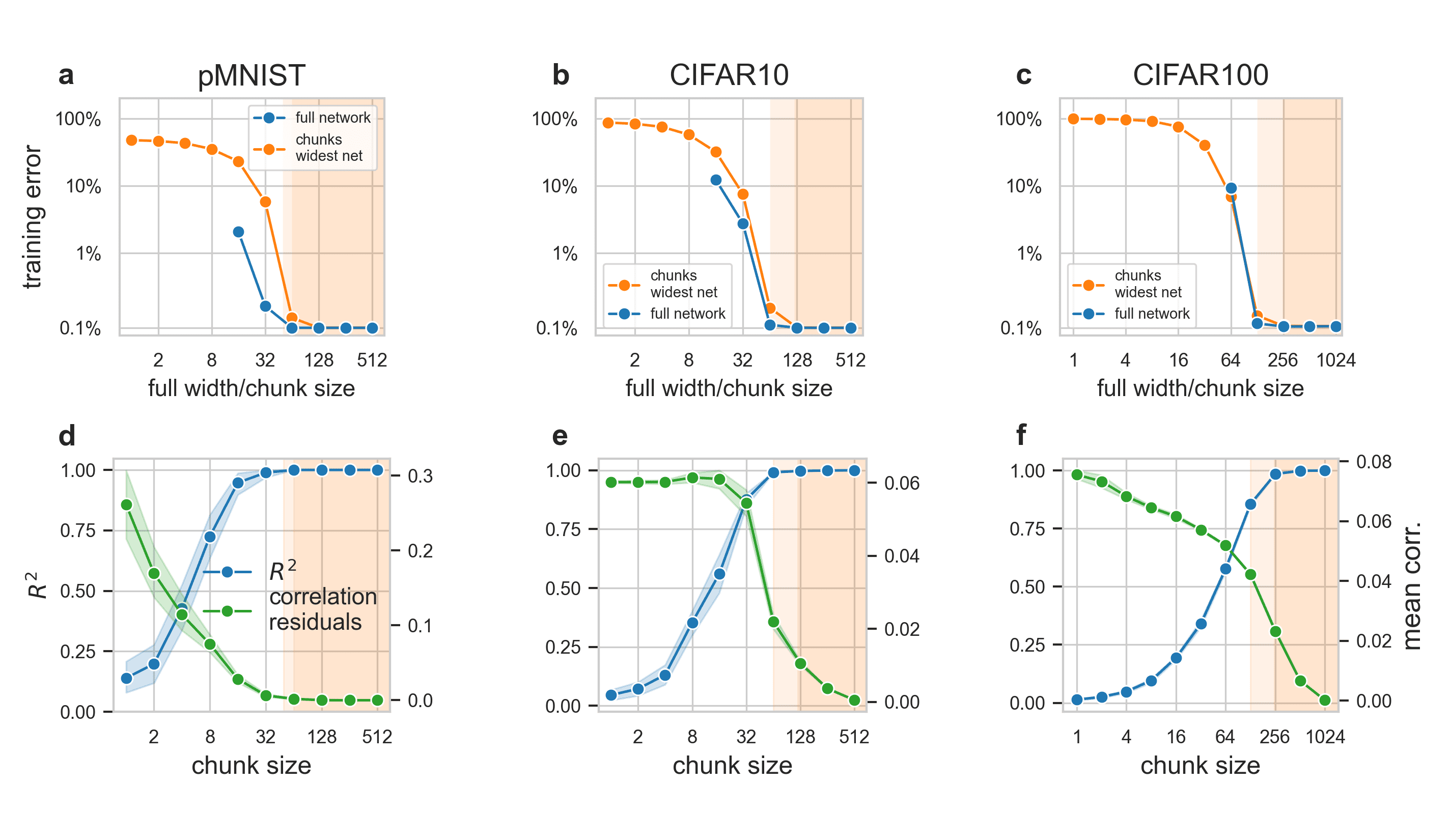}
    \caption{\label{fig:hallmarks-redundancy}\textbf{The three signatures of representation redundancy.} \textbf{(i)} The training errors of the full networks (blue) and of the chunks taken from the widest network (orange) approach zero beyond a critical width/chunk size, resp. (panels {\bf a}-{\bf c}). \textbf{(ii)} The final representation of the widest network can be reconstructed from a chunk using linear regression~\eqref{eq:fit} with an explained variance $R^2$ close to 1 (blue lines in panels {\bf d}-{\bf f}). \textbf{(iii)} The residuals of the linear map can be modeled as independent noise: we show this by plotting the mean correlation of these residuals (green line, panels {\bf d}-{\bf f}), averaged over 100 reconstructions starting from different chunks. A low correlation at high $R^2$ indicates that the chunk contains the information of the full representation with some statistically independent noise. \emph{Experimental setup:} FC5 on pMNIST, Wide ResNet-28 on CIFAR10/100. Full details in Methods section~\ref{sec:methods-nn}}
\end{figure*}

\paragraph{Clones interpolate the training data.}
A trained deep network often represents the salient features of the data set well enough to achieve (close to) zero classification error on the training data. 
In the top panels of Fig.~\ref{fig:hallmarks-redundancy}, we show that wide networks can interpolate their training set also using just a subset of $w_c > w_c^*$ random neurons: the dark orange profiles show that when the size of a chunk is greater than $w_c^* \sim$ 50 for pMNIST, 100 for CIFAR10 and 200 for CIFAR100, the predictive accuracy on the training set remains almost 100\%.
The minimal size of a clone $w_c^*$ can be identified with the minimal number of neurons required to interpolate the training set.
Beyond $w_c^*$, the neurons of the final representation become redundant since the training error remains (close to) zero even after removing neurons from it. 
The number of distinct clones in a network of width $W$ is $n = {{W}/{w_c^*}}$. If distinct clones provide independent measures of the same salient features of the data, the \emph{test} error decays approximately as $n^{\nicefrac{-1}{2}}$ or equivalently $W^{\nicefrac{-1}{2}}$. 
In the following, we will indeed see that distinct clones differ from each other by uncorrelated random noise.

\paragraph{Clones reconstruct the full representation almost perfectly.} From a geometrical perspective, the important features of the final representation correspond to directions in which the data landscape shows large variations \citep{bengio2013representation}.
A clone is a chunk that is wide enough to encode almost exactly these directions 
(since its training error is almost zero), 
but using much fewer neurons than the full final representation.
We analyze this aspect by performing a linear reconstruction of the $W$ activations of the last hidden representation of the widest network, starting from a random subset of $w_c$ activations using ridge regression with a small regularization penalty according to Eq.~\eqref{eq:fit}.
The blue profiles in Fig.~\ref{fig:hallmarks-redundancy}-(d,e,f),  show the $R^2$ coefficient of fit as a function of the chunk size $w_c$ for pMNIST (left), CIFAR10 (center), CIFAR100 (right).
When $w_c$ is very small, say below $6$ for pMNIST, $20$ for CIFAR10, and $60$ for CIFAR100, the~$R^2$ coefficient grows almost linearly with $w_c$. In this regime, adding a randomly chosen activation from the full representation to the chunk increases substantially  $R^2$.
When $w_c$ becomes larger  $R^2$ reaches almost one.%
This transition happens when $w_c$ is still much smaller than $W$  and corresponds approximately to the regime in which the test error starts scaling with the inverse square root of $w_c$ (see Fig.~\ref{fig:chunk-error-scaling}).
The almost perfect reconstruction of the original data landscape with few
neurons can be seen as a consequence of the low \emph{intrinsic dimension} (ID) of
the representation~\citep{ansuini2019intrinsic}.
The ID of the widest representations gives a lower bound on the number of coordinates required to describe the data manifold, and hence on the neurons that a chunk needs to have the same classification accuracy as the whole representation. The ID of the last hidden representation is 2 in pMNIST, 12 in CIFAR10, and 14 in CIFAR100, numbers which are much lower than $w_c^*$, the width at which a chunk can be considered a clone.

\paragraph{Clones differ from each other by uncorrelated random noise.} 
When~$w_c > w_c^*$, the small residual difference between the chunked representation and the full representation can be approximately described as statistically independent random noise.
The green profiles of the bottom panels of Fig. \ref{fig:hallmarks-redundancy} show
the \emph{mean correlation} of the residuals of the linear fit (see Sec. \ref{sec:methods-analysis}), a measure that indicates the level of correlation of the chunk representations and the full representation.
Below $w_c^*$, the residuals are not only large but also significantly correlated, since they are related to relevant features of the data that are not covered by the neurons of the chunk. As the chunk width increases above  $w_c^*$, the correlation between residuals drops basically to zero.
Therefore, in networks wider than $w_c^*$, any two chunks of equal size $w_c > w_c^*$ can be effectively considered as equivalent copies, or clones, of the same representation (that of the full layer), differing only by a small and non-correlated noise, consistently with the scaling law of the error shown in Fig.~\ref{fig:chunk-error-scaling}.
\vspace{-0cm}

\paragraph{The dynamics of training.}
In the previous paragraphs, we set forth evidence in support of the hypothesis
that large chunks of the final representation of wide DNNs behave approximately
like an ensemble of independent measures of the full feature space. This allowed
us to interpret the decay of the test error of the full networks with the 
network width observed empirically in Fig.~\ref{fig:error-full-network}. The
three conditions that a chunked model satisfies in the regime in which its test
error decays as $w_c^{\nicefrac{-1}{2}}$ are represented in
Fig.~\ref{fig:hallmarks-redundancy}: (i) the training error of the chunked model is close to zero; (ii) the chunked model can be used to reconstruct the full final representation with an
$R^2\sim 1$ and (iii) the residuals of this reconstruction can be modeled as
independent random noise. These three conditions are all observed at the end of
the training.
We now analyze the training \emph{dynamics}. We will see that for the clones to arise, 
models not only need to be wide enough but also, crucially, they need to be trained to maximize their performance.

\begin{figure*}[t]
\centering
\includegraphics[width=1.\textwidth]{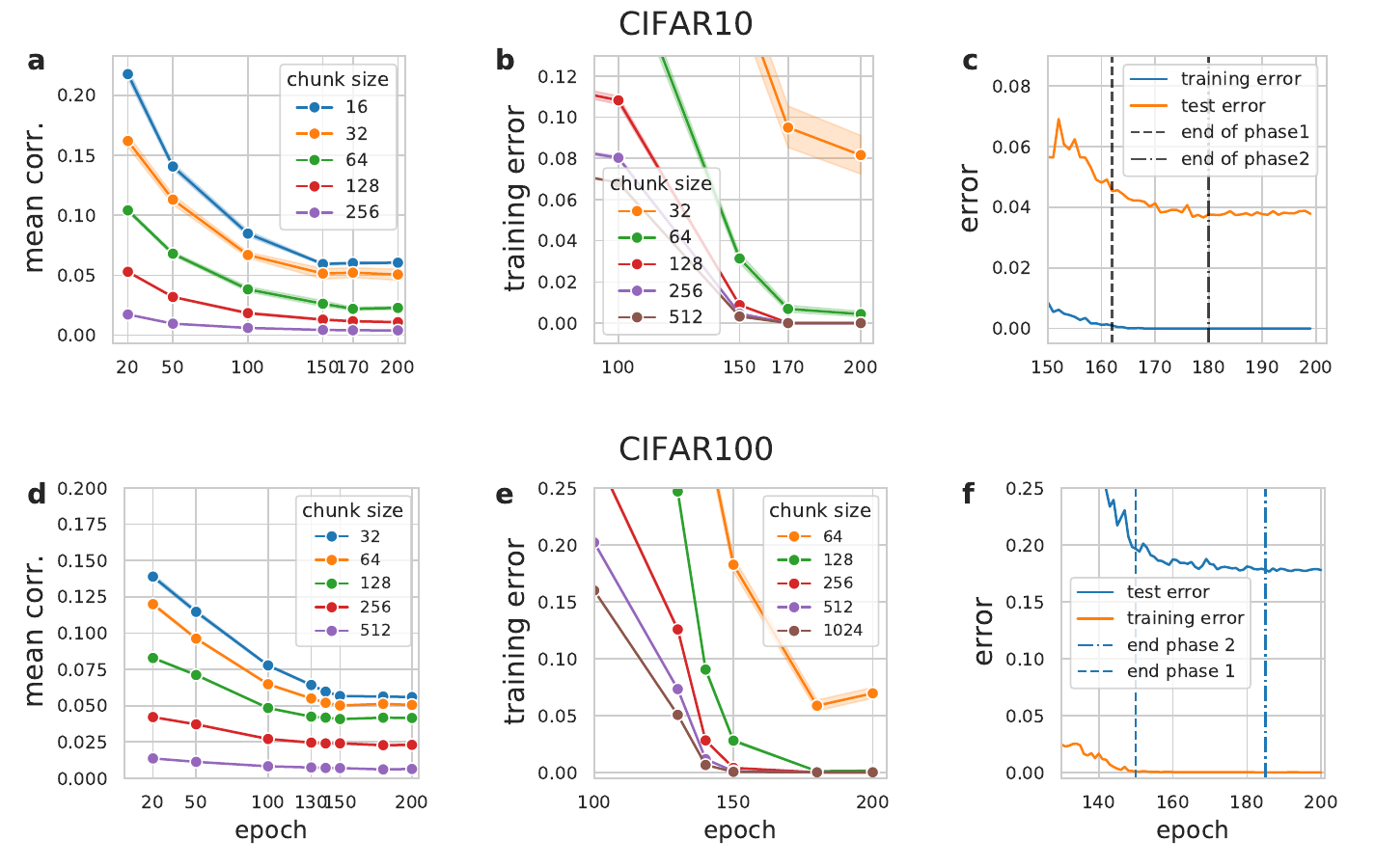}
\caption{
\textbf{The onset of clones during training.}
\textbf{a:} As in Fig.~\ref{fig:hallmarks-redundancy}, we show the mean correlation of the residuals of the linear reconstruction of the final representation from chunks, but this time as a function of training epochs. A small correlation indicates that the reconstruction error in going from chunks to the final representation can be modeled as independent noise. 
\textbf{b:} Training error during training for chunks of different sizes.
After the network has reached zero training error at $\sim 160$ epochs, continuing to train improves the training accuracy of the chunks.
\textbf{c:} Test and training error during training for the full network. Between epochs 160 and 180, the clones of the full network progressively achieve zero training error. In the same epochs, one observes a small improvement in the test error.
}
\label{fig:training_dynamics}
\end{figure*}

Clones are formed in two stages, which occur at different times during training.
The first phase begins as soon as training starts: the network gradually adjusts the chunk representations to produce independent copies of the data manifold.
This can be observed in Fig.~\ref{fig:training_dynamics}-a for CIFAR10 and Fig.~\ref{fig:training_dynamics}-b for CIFAR100, which depict the mean correlation between the residuals of the linear fit from the chunked to the full final representations of the Wide-ResNet28 architectures.
The mean correlation is the same quantity that we analyze in Fig.~\ref{fig:hallmarks-redundancy}-(e-f, green profiles), but now as a function of the training epoch. 
%
%
%
As training proceeds, the correlations between residuals diminish gradually until epoch 160 for CIFAR10 and 150 for CIFAR100. After that, further training does not bring any sizeable reduction in their correlation.
For CIFAR10 (Fig. \ref{fig:training_dynamics}-a) the mean correlation becomes particularly small for chunks larger than 64, which is the size of a clone in CIFAR10 (see for instance shaded area of Fig. \ref{fig:chunk-error-scaling}-b.
For CIFAR100 instead, a clear gap in the mean correlation at the end of training is less evident (see also Fig. \ref{fig:hallmarks-redundancy}-d), but the residuals of the linear fit done with chunks larger than 256 and 512 (red and purple profiles of \ref{fig:training_dynamics}-d) are clearly smaller than those of smaller chunks.

At epoch 160 for CIFAR10, and 150 for CIFAR100, the full network also achieves zero error on the training set, as shown in  Fig.~\ref{fig:training_dynamics}-(b-e) (brown) and Fig.~\ref{fig:training_dynamics}-(c-f) (blue).
This event marks the end of the first phase and the beginning of the second phase, where the training error of the clones keeps decreasing while the full representation has already reached zero training error.
For instance, CIFAR10 chunks of size 64 at epoch 150 have training errors comparable to the test error, around 0.04 (see  Fig.~\ref{fig:training_dynamics}-b).
In the subsequent $\sim 20$  epochs, the training error of clones of size 128 and 256 reaches exactly zero, and the training error of chunks of size 64  plateaus around 0.5\%. 
In CIFAR100 (Fig.~\ref{fig:training_dynamics}-d), from epoch 150 to epoch 185, the training error of the chunks with size 128/256 decreases below 0.5\%, while for smaller chunk sizes it remains above 5\%. 
Random chunks with sizes larger than $W = 64/128$ in CIFAR10 and $W = 128/256$ in CIFAR100 can fit the training set, thus having the same representational power as the whole network on the training data. 
In these regimes, the test accuracies decay approximately with the same law as that of independent networks with the same width (see Fig. \ref{fig:chunk-error-scaling}-(b, c)). This picture suggests that for CIFAR100, the size of a clone is 128/256, slightly larger than the size of the clones in CIFAR10.

Importantly, both phases improve the generalization properties of the network.
This can be seen in Fig.~\ref{fig:training_dynamics}-(c-f), which reports the training and test error of the network, with the two phases highlighted.
After the end of the first phase, the training error of the full networks remains consistently smaller than 0.1\% (blue profiles) while the test error continues to decrease from 4.8\% to 3.9\%  for CIFAR10, from 19,4\% to 17.6\% for CIFAR100.  
Both phases lead to a reduction in the test error, although the first phase leads by far to the greatest reduction, consistent with the fact that the greatest improvements in accuracy typically arise during the first epochs of training.
The formation of clones can be considered finished around epochs 180 and 185 for CIFAR10 and CIFAR100, respectively, when all the clones have reached almost zero error on the training set. Then, we observe that the test error stops improving.
%

\begin{figure}
\centering
\includegraphics[width=1.\columnwidth]{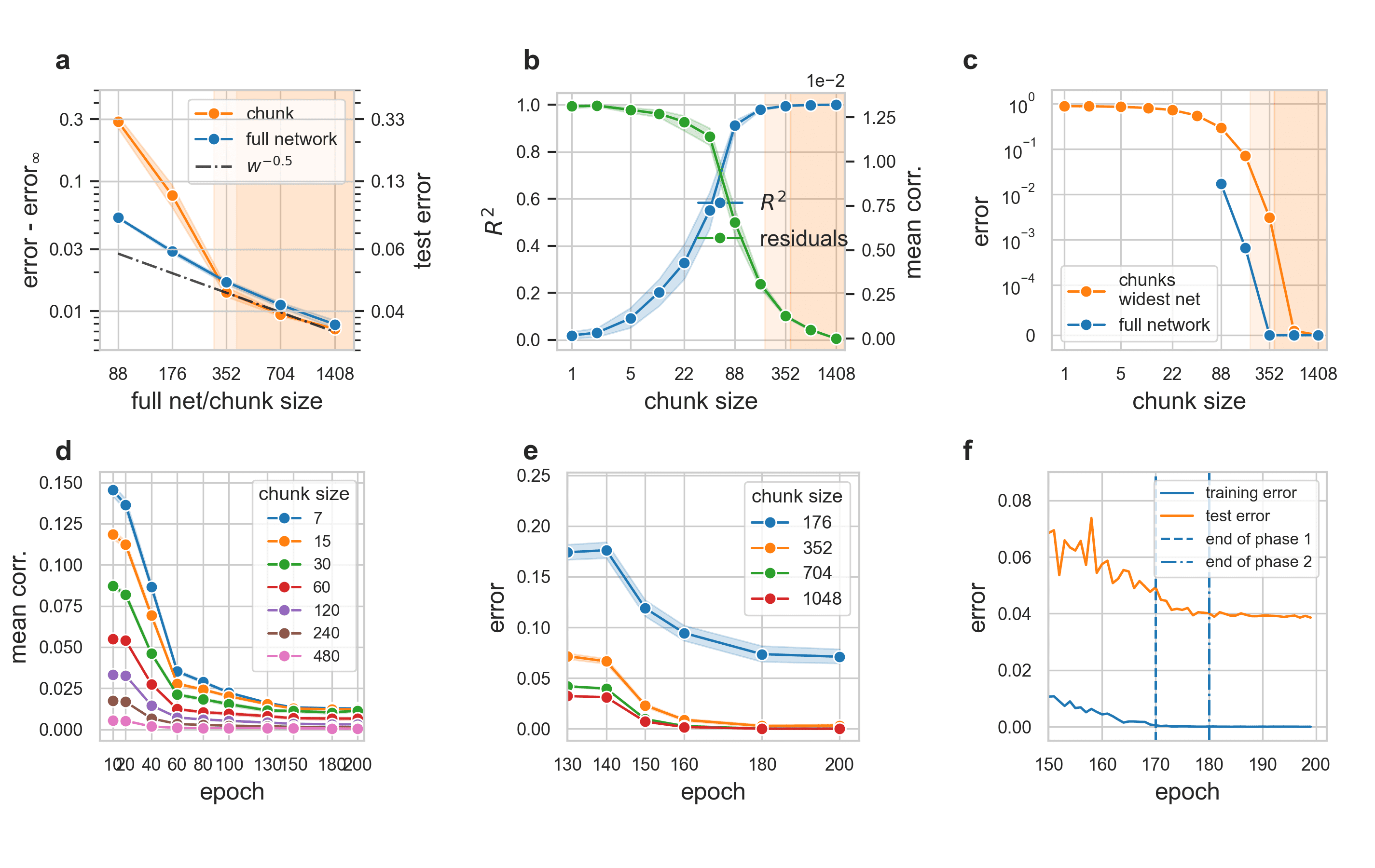}
\caption{
\label{fig:app_densenet} 
\textbf{A DenseNet40 architecture.}
\textbf{a:} Decay of the test error of independent networks (blue) and chunks of the widest network (orange) to the error of an ensemble average of ten of the widest networks (DenseNet40-BC, k=128)
\textbf{b:} 
Blue profile: $R^2$ coefficient of the ridge regression of a chunk of $w_c$ neurons ($x$-axis) to the full layer representation.
Green profile: mean correlation of the residuals of the mapping as described in Sec. \ref{sec:methods-analysis}.
\textbf{c:} 
Training error of various DenseNet40 of increasing width (blue) and of chunks of the widest architecture (orange).
\textbf{d:} 
The mean correlation of the residuals from the linear reconstruction of the final representation from chunks of a given size for a DenseNet40-BC (k=128) during training. 
\textbf{e:} 
Training error dynamics of chunks of a DenseNet40-BC (k=128).
\textbf{f:} 
Training and test error dynamics for a DenseNet40-BC (k=128).}
\end{figure}

\vspace{10pt}
Figure \ref{fig:app_densenet} summarizes the results discussed so far for the case of 
a Densenet40 trained on CIFAR10.

\newpage

\subsection{Two necessary conditions to observe redundant representations.}
\begin{figure}
\centering
\includegraphics[width=0.75\columnwidth]
{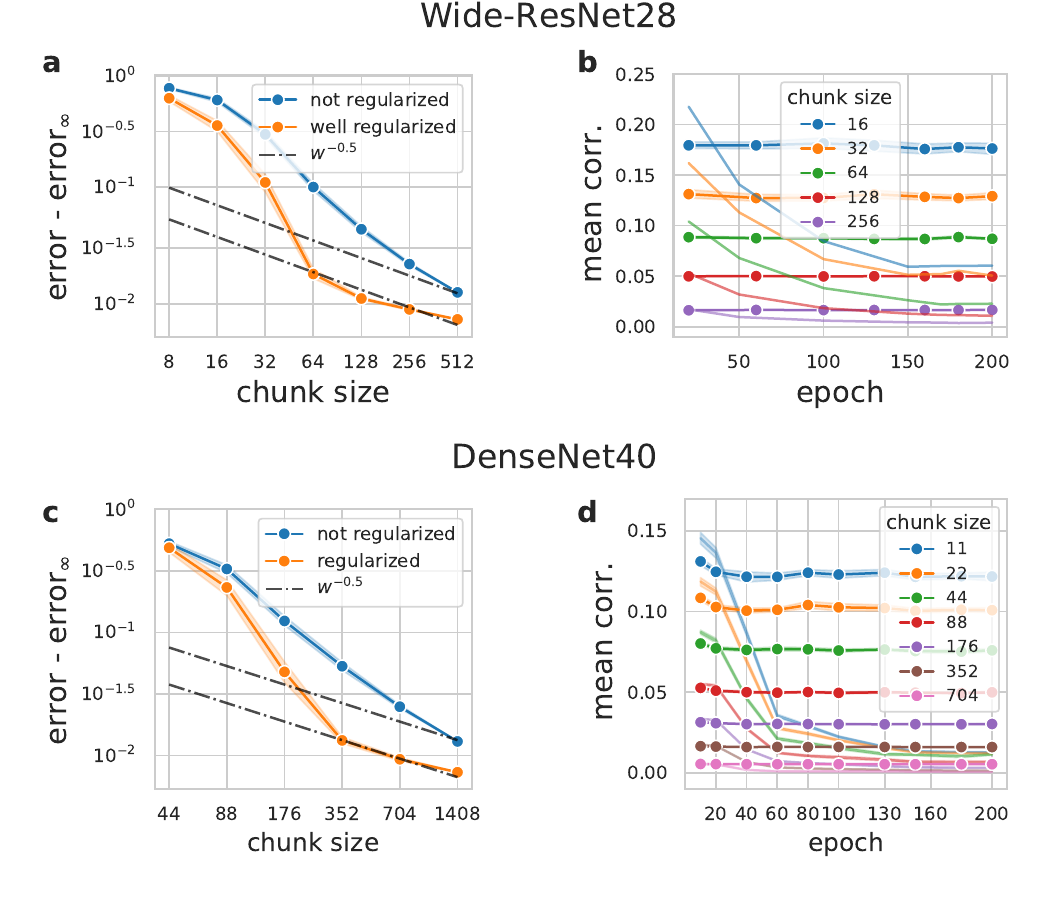}
\caption{\label{fig:narrow_bad-trained} \textbf{Networks trained without weight decay on CIFAR10.} \textbf{a:}  $\mathrm{error}$ -$\mathrm{error}_\infty$ for chunks of a Wide-ResNet28-8 trained without weight decay (blue) and with weight decay (orange, taken from Figure \ref{fig:chunk-error-scaling}-b). \textbf{b:}  Mean correlation between residuals of the linear reconstruction of the full representation from chunks of different sizes for two networks: one trained without weight decay (thick lines), and one using weight decay (thin lines, same data as in Fig.~\ref{fig:training_dynamics}-a).
\textbf{c:} DenseNet40 (k = 128) trained without weight decay (blue) and with weight decay (orange, taken from Figure \ref{fig:app_densenet}-a)
\textbf{c:}  Mean correlation of the residuals for the DenseNet40 trained without weight decay (thick lines), and one using weight decay (thin lines, same data as in Fig.~\ref{fig:app_densenet}-d).
}
\end{figure}

\paragraph{Clones appear only in regularized networks.}
So far in this work, we have shown only examples of networks and data sets in which representations are redundant. 
If the network is not regularized, some of the signatures described above don't appear even if the width of the final representation is much larger than $W^*$ (the minimum interpolating width). 
Figure \ref{fig:narrow_bad-trained} shows the case of the Wide-ResNet28-8 and DenseNet40 analyzed in Figs. \ref{fig:training_dynamics} and \ref{fig:app_densenet}, both trained on exactly the same data set (CIFAR10) but
without weight decay.
Panels a and c of Fig. \ref{fig:narrow_bad-trained} show that when the networks are trained without regularization (blue line), the error does not scale as $w_c^{-1/2}$. 
This, as we have seen, indicates that the last hidden representation cannot be split into clones equivalent to the full layer.
Indeed, the mean correlation of the residuals of the linear map of the chunks to the full representation remains approximately constant during training (see panels b-d), and is always much higher than what we observed for the same architecture and data set when training is performed with weight decay.

\begin{figure}
  \centering
  \includegraphics[width=1.\textwidth]{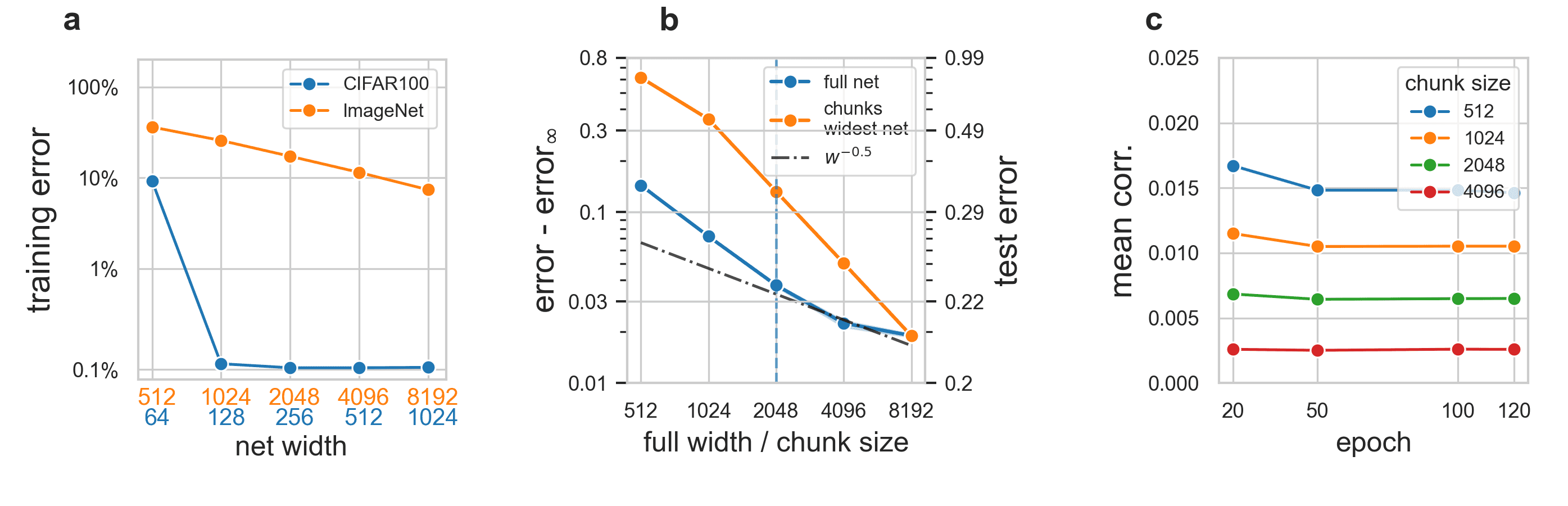}
  \caption{\label{fig:imagenet}\textbf{In ResNet50 trained on ImageNet clones do not appear.}
\textbf{a:} Training error as a function of the network width for CIRA100 (blue) and ImageNet (orange).
  \textbf{b:} Decay of the test error as a function of the network width (blue) and for chunks of the widest ResNet50 (orange) to the error of an ensemble of ResNet50-4.  The ensemble consists of four networks.
  \textbf{c:}
  Mean correlation (see Sec. \ref{sec:methods-analysis}) of the residuals of the linear map of a chunk of the last hidden representation to the full representation. The network examined is ResNet50-4.
 }
\end{figure}

\paragraph{Clones appear only if a network interpolates the training set: the case of ImageNet.} 
We saw that a chunk of neurons can be considered a clone if it fully captures the relevant features of the data, achieving almost zero training error (see Fig. \ref{fig:hallmarks-redundancy}). 
This condition is not satisfied for most of the networks trained on ImageNet \cite{belkin2019reconciling}, therefore we do not expect to see redundant representations in this important case.
We verified this hypothesis by training a family of ResNet50s where we multiply all the channels of the layers after the input stem by a constant factor $c \in \{0.25, 0.5, 1, 2, 4\}$. In this manner, the widest final representation we consider consists of $8192$ neurons, which is four times wider than both the standard ResNet50 \citep{he2016deep} and its wider version \citep{zagoruyko2016wide} (see Sec. \ref{sec:methods-nn}). 
We trained all the networks following the standard protocols and achieved test errors comparable to or slightly lower than those reported in the literature (see table \ref{table:accuracy}).
We find that even in the case of the largest ResNet50, the top-1 error on the training set is $\sim$ 8\% (see Fig. \ref{fig:imagenet}-a) and the network does not achieve interpolation, as discussed also in \cite{belkin2019reconciling}.

In this setting, none of the elements associated with the development of independent clones can be observed. The scaling of the test error of the chunks is steeper than $w_c^{-1/2}$ (see Fig. \ref{fig:imagenet}-b), suggesting that chunks remain significantly correlated to each other. Figure \ref{fig:imagenet}-c shows that the mean correlation of the residuals does not decrease during training, as it happens for the networks we trained on CIFAR10 and CIFAR100 (see Fig. \ref{fig:training_dynamics}-(a-d)).
We conclude that in a ResNet50, a representation with 8192 neurons is too narrow to encode all the relevant features redundantly on ImageNet, and a chunk as large as 4096 activations is not able to reconstruct all the relevant variations of the data as it does in the cases analyzed in Sec. \ref{sec:results}.

\section{Discussion}
\label{sec:discussion}
This work is an attempt to explain the paradoxical observation that over-parame-terization boosts the performance of DNNs.
This ``paradox'' is  not a peculiarity of DNNs: if one trains a prediction model with $n$ parameters using the same training set, but starting from independent initial weights and independently receiving samples, one can obtain, say, $m$ models which, in suitable conditions, provide predictions of the same quantity with independent noise due to initialization, SGD schedule, etc.
If one estimates the target quantity by an ensemble average, the statistical error will (ideally) scale with $m^{\nicefrac{-1}{2}}$, and therefore with $N^{\nicefrac{-1}{2}}$, where $N=n\,m$ is the total number of parameters of the combined model.
This will happen even if $N$ is much larger than the number of data.

What is less trivial is that a DNN can accomplish this scaling within a single model, in which all the parameters are optimized collectively via the minimization of a single loss function.
Our work describes a possible mechanism at the basis of this phenomenon in the special case of neural networks in which the last layer is very wide.
We observe that if the layer is wide enough, random subsets of its neurons can be viewed as approximately independent representations of the same data manifold (or clones).
This implies a scaling of the error with the width of the layer as~$W^{\nicefrac{-1}{2}}$, which is qualitatively consistent with our observations. 

\paragraph{The impact of network architecture.}
The capability of a network to produce statistically independent clones is a genuine effect of the over-parametrization of the \emph{whole} network as redundancies appear even if the last layer width is kept constant and the width of all intermediate layers is increased, but if the network is too narrow, increasing the width of \emph{only} the final representation is not sufficient to make the last layer redundant.

\begin{figure}
\centering
\includegraphics[width=0.8\columnwidth]{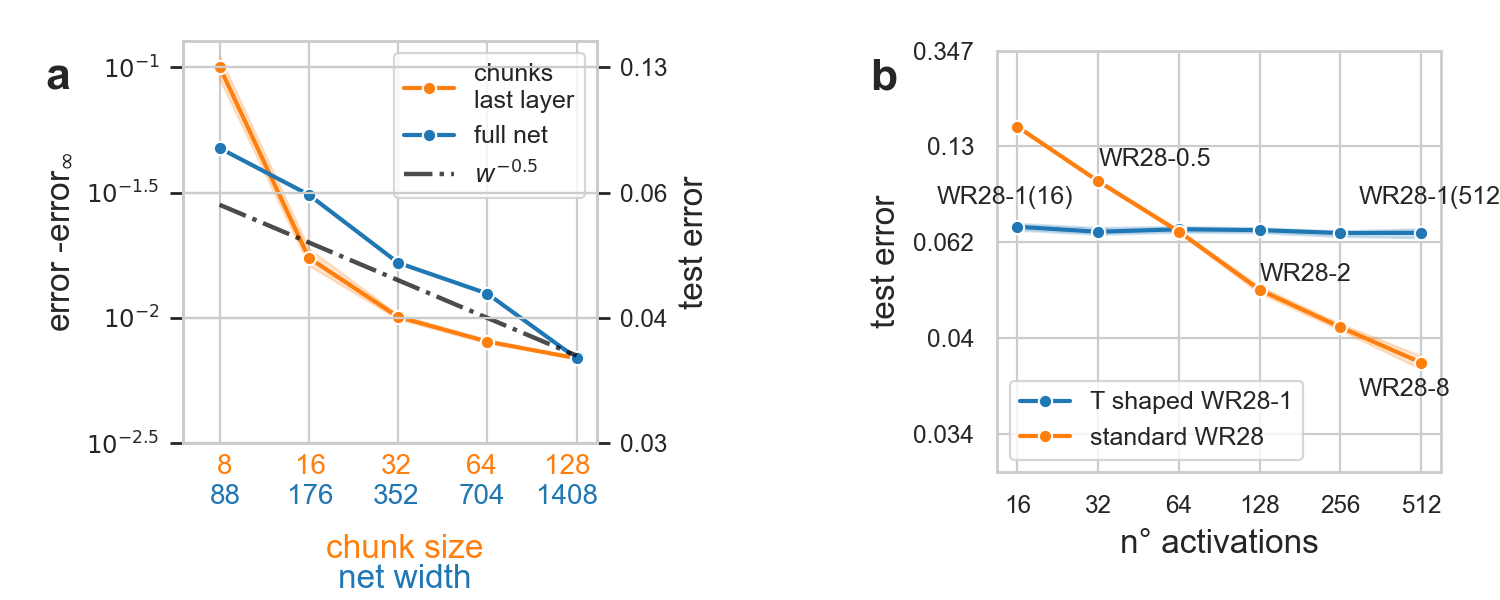}
\caption{\label{fig:app_only_last} 
\textbf{Impact of the width of the intermediate layers.}
}
\end{figure}

To simulate the first scenario we train a DenseNet40 on CIFAR10 with an additional 1 × 1 convolution at the end of the network, before the last fully connected layer, to keep the number of output channels fixed at 128. 
The blue profile of Fig. \ref{fig:app_only_last}-a shows the test accuracy as we increase the width of the architecture (reported on the x-axis in blue). 
The test accuracy decays approximately as $w_c^{-\nicefrac{1}{2}}$, even when the width of the final representation has 128 activations.
The orange profile instead shows the test error of chucks of the widest architecture. This architecture has 1408 activations in the second-to-last convolutional representation and 128 activations in the layer before the output.
The presence of a bottleneck at the end of the architecture makes the clones much smaller: the orange profile shows a strong deviation from the $w^{\nicefrac{-1}{2}}$ only for chunk sizes smaller than 16. 
The clone size of the same network without a bottleneck is around 350 as shown in Fig. \ref{fig:cartoon_}-b.
 We also verified that 16 random activations are sufficient to interpolate the training set with an error smaller than  $5 \cdot 10^{-3}$) and that the $R^2$ coefficient of fit to the full layer is 0.912 (0.98 for chunk sizes $=32$). 
 The phenomenology described in Sec. \ref{sec:results} applies also when a bottleneck of 128 channels is added at the end of the network.
%

In the second scenario, we trained a ResNet28-1, increasing only the number of channels in the last layer with a $1 \times 1$.
The blue profile of Fig. \ref{fig:app_only_last}-{b} shows that the test error of the modified ResNet28-1 is approximately constant. On the contrary, when we increase the width of the whole network, the test error decays to the asymptotic test error with an approximate scaling of  $1/\sqrt{w}$.

\paragraph{The impact of training.}
The mechanism we described is robust to different training objectives 
since we trained the convolutional networks with cross-entropy loss and the fully connected networks with hinge loss.
However, even for wide enough architectures, clones appear only if the training is continued until the training error reaches zero. 
In our examples, by stopping the training too early, for example when the training error is similar to the test error, the chunks of the last representation would not become entirely independent from one another, and therefore they could not be considered clones.

\paragraph{Neural scaling laws.} Capturing the asymptotic performance of neural networks via scaling laws is an active research area. Hestness \emph{et al.}~\citep{hestness2017deep} gave an experimental analysis of scaling laws w.r.t~the training data set size in a variety of domains. Refs.~\citep{rosenfeld2020constructive, kaplan2020scaling} experimentally explored the scaling of the generalization error of deep networks with the number of parameters/data points across architectures and applications
domains for supervised learning, while \citep{henighan2020scaling}
identified empirical scaling laws in generative models.
Geiger \emph{et al.}~\citep{geiger2020scaling} found that the generalization error of networks trained on pMNIST \emph{without} weight decay, momentum, and data augmentation scales as $W^{-1}$. 
Bahri \emph{et al.}~\citep{bahri2021explaining} showed the existence of four scaling regimes and described them theoretically
in the NTK or \emph{lazy regime}~\cite{jacot2018neural,du2018gradient,chizat2019lazy},
where the network weights stay close to their initial values throughout training.
We instead consider the feature learning regime, where weights move considerably as the networks learn task-relevant features. Moreover, we train our networks with weight decay which is unavoidable to obtain models with state-of-the-art performance. 
Previous theoretical work mainly focused on the NTK regime and did not study the impact of weight decay on scaling laws, so we hope that our results can spark further studies on the role of this essential regularizer.

\paragraph{Relation to theoretical results in the mean-field regime.}
Our empirical results also agree with recent theoretical results that were obtained for two-layer neural networks~\citep{Mei2018, rotskoff2018interacting, Chizat2018, Sirignano2018, goldt2019dynamics, refinetti2021classifying}.
These works characterize the optimal solutions of
two-layer networks trained on synthetic data sets with some controlled features.
In the limit of infinite training data, these optimal solutions correspond to
networks where neurons in the hidden layer duplicate the key features of the
data.
These ``denoising
solutions'' or ``distributional fixed points'' were found for networks with wide hidden layers~\citep{Mei2018, rotskoff2018interacting, Chizat2018, Sirignano2018} and wide input
dimension~\citep{goldt2019dynamics, refinetti2021classifying}. Another point of connection
with the theoretical literature is the concept of \emph{dropout stability}. A network is said to be $\epsilon$-dropout stable if its training loss changes by less than $\epsilon$ when 
half the neurons are removed at random from each of its layers~\cite{kuditipudi2019explaining}. Dropout stability has been rigorously linked to several phenomena in neural networks, such as the connectedness of the minima of their training landscape~\cite{shevchenko2020landscape, nguyen2021connectivity}.

\paragraph{Bias-variance trade-off and implicit ensembling} The success of various deep
learning architectures and techniques has been linked to some form of ensembling. 
The successful dropout regularization technique~\citep{hinton2012improving, srivastava2014dropout}
samples from an exponential number of ``thinned'' networks during training to
prevent co-adaptation of hidden units. While this can be seen as a form of
(implicit) ensembling, here we observe that co-adaptation of hidden units in the form of clones occurs \emph{without} dropout, and is crucial for their improving performance with width. Recent theoretical work on random features showed that ensembling and over-parameterization are two sides of the same coin and that both mitigate the increase in the variance of the network that classically leads to \emph{worse} performance with over-parameterization due to the bias-variance trade-off~\citep{dascoli2020double,adlam2020understanding, lin2021causes}.

\begin{figure}
\centering
\includegraphics[width=1.\columnwidth]{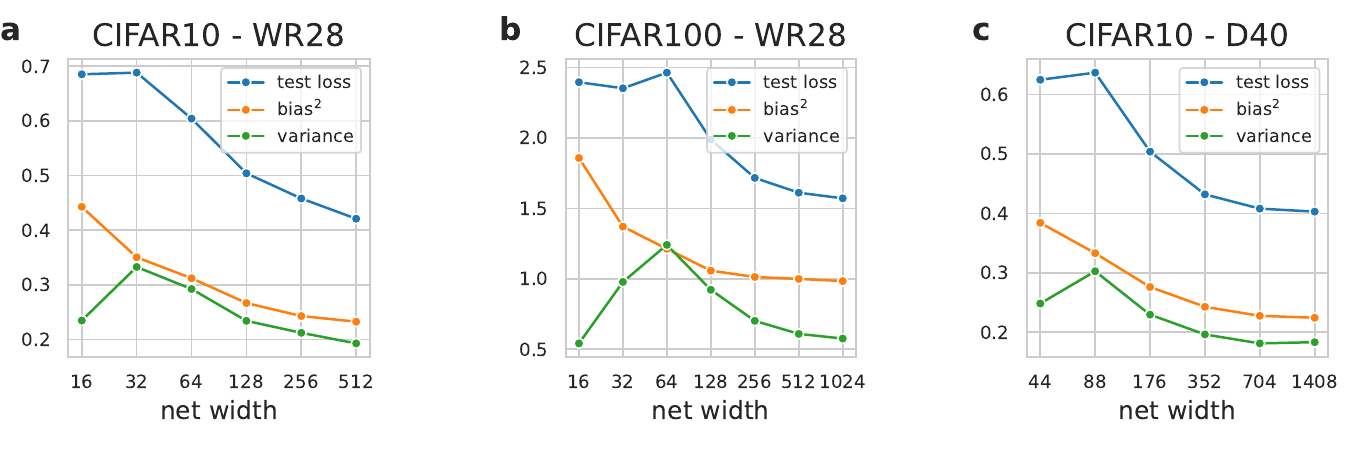}
\caption{\label{fig:app_bias_variance_ls0} \textbf{Bias-variance profiles in CIFRA10 and CIFAR100.} 
Bias and the variance profiles for the convolutional architectures analyzed in the paper: Wide-ResNets and DenseNets trained on CIFAR10, and Wide-ResNets trained on CIFAR100.
The average over the data distribution is approximated by splitting the CIFAR training sets into five disjoint subsets containing 10~000 images each and training the networks from scratch on each of them. We use the same regularization for all the networks, namely that of the largest architectures, with weight decay equal to $5 \cdot 10^{-4}$ and label smoothing equal to 0.
We repeat the procedure 4 times, for a total of 20 training runs for each network width, as described in Ref.~\cite{yang2020rethinking}.
}
\end{figure}

\citet{belkin2019reconciling} showed that the test loss has a double descent profile with a peak in correspondence to the interpolation threshold if deep networks are not regularized, due to a very large model variance.
We have seen that in correspondence to the interpolation width, the last representation starts to develop redundancies (see Sec. \ref{sec:hallmarks_redundancies}).
It is therefore worth analyzing whether the peak of the variance and the development of clones are connected to each other. 
In Fig. \ref{fig:app_bias_variance_ls0} we show the bias and the variance profiles for the convolutional architectures analyzed in the paper: Wide-ResNets and DenseNets trained on CIFAR10, and Wide-ResNets trained on CIFAR100. 
Since we trained the models using the cross-entropy loss, the standard bias-variance decomposition, which assumes the square loss, does not apply. 
Instead, we used the method recently proposed by \citet{yang2020rethinking} to estimate the bias and the variance on networks trained with cross-entropy loss. 
As expected, the bias of the models decreases as we add parameters and make the model more flexible. 
The variance instead initially grows with the network width to reach its peak at $W_\mathrm{peak} = 32$ and $64$ for CIFAR10 and CIFAR100 trained on Wide-ResNet28 (a, b) and  $W_\mathrm{peak} = 88$ on CIFAR100 trained on DenseNet40 (c). 
and then decreases, allowing the model to generalize better and better.
The clone sizes $w_c^*$ for these architectures are slightly above the widths at which the variance peaks and are $w_c^* = 64$, and $128$ for CIFAR10 and CIFAR100 trained on Wide-ResNet28 (compare Figs.~\ref{fig:chunk-error-scaling} and~\ref{fig:hallmarks-redundancy}) and 
$w_c^* = 170/250$ (Fig.~\ref{fig:app_densenet}). Indeed, the peak of the variance occurs at a width slightly smaller than the interpolation width.
In all cases, however, the onset of the clones occurs at a width that is approximately two times larger than $W_\mathrm{peak}$.

\paragraph{Conclusion.} The framework introduced in this work allows verifying if a neural network is sufficiently expressive to encode multiple statistically independent representations of the same ground truth, which, we believe, is a fair proxy of model quality and robustness. 
The capability of deep neural networks to develop clones is a direct consequence of the relatively low intrinsic dimension of the representation, discussed in Chapter \ref{ch:gride}. Indeed, the ID of the final representation is a very robust feature of a data set, ultimately determined only by the classification task\cite{ansuini2019intrinsic}. Making the representation wider by adding activations does not change significantly this ID. This implies that the added neurons must encode the same information.  
In particular, we find that reaching interpolation on the training set is not necessarily detrimental for generalization, and is instead a necessary condition for developing redundancies which, in turn, reduce the test error.

\chapter{Conclusion}
\label{ch:conclusion}
In this Thesis, we developed two parallel and interlaced research topics. The first is the development of unsupervised learning and statistical measures to understand high-dimensional data manifolds. The second is the study of how deep neural networks for real-world tasks represent the data in their inner layers. Since the data representations are very high-dimensional, exploiting advanced unsupervised learning techniques and developing our own tools was essential to address the second task. 
\vspace{7pt}

One of the key ingredients for analyzing the structure of a high-dimensional data manifold is estimating appropriately its dimension, which is typically orders of magnitude smaller than the plain number of variables defining the representation. In Chapter \ref{ch:gride}, we have addressed the problem of the estimate of the intrinsic dimension (ID) of noisy data sets and proposed Gride, an extension of the TwoNN estimator to analyze the scale dependence of the ID without decimating the data points.
Gride uses neighbors of higher order instead of using bootstrap to extract subsets to estimate the ID above the noise scale.
Gride is two times faster than TwoNN and orders of magnitude faster than other neighbor-based estimators on data sets embedded in very high-dimensional spaces, making it particularly well suited, for the analysis of the data representations of neural networks. 
We have applied Gride to analyze the hidden representations of state-of-the-art networks trained to classify ImageNet (ResNet152) or to reconstruct the images from incomplete patches (iGPT). 
These architectures show qualitative differences in the ID profiles across the hidden layers. 
In ResNet152, the ID profile has a single hunchback and reaches its lowest value at the output layer \cite{ansuini2019intrinsic} while in iGPT there are two hunchbacks with a local minimum in the middle of the network where the linear probe accuracy on ImageNet is the best (see \cite{igpt} and Sec. \ref{sec:hier_nucl_evolution_cluster_ari}).

These results point to a possible connection between the low ID and the abstract content of a representation. Besides the example of iGPT,  \citet{ansuini2019intrinsic} showed that ID at the output of convolutional networks is negatively correlated with the accuracy of the models: the lower the ID, the more accurate is the model. 
Consistent with this picture, a recent study of protein Language Models by \citet{protein_language_models},  revealed that the evolutionary and structural information extracted by the network is maximal in a range of layers where the ID plateaus to its lowest value. 
At the end of Chapter \ref{ch:gride}, we also mention a similar trend for the case of the StyleGAN, a network for image generation, studied in \cite{style_gan_id}. 
A direction for future research is to better understand the relation between the ID of hidden representations and their semantic expressiveness, in particular in deep generative models. 
For instance, the recent and influential work by \citet{stable_diffusion} has shown that powerful diffusion models can be trained very efficiently using as input features the latent representation of an autoencoder pre-trained on a large image corpus.
If a connection between (low) ID and semantic expressiveness could be found, the unsupervised knowledge of the ID of a latent space can be used to select, without labels, the best (i.e. most semantically rich) latent space to use for downstream tasks such as image synthesis or text-to-image generation \cite{stable_diffusion}.
\vspace{7pt}

Chapter \ref{ch:hier-nucl} and \ref{ch:repr-mitosis} are fully devoted to the analysis and the characterization of data representations in deep neural networks.  In Chapter \ref{ch:hier-nucl}, we analyze the probability flux and the similarity between representations of neural networks trained on ImageNet, using a similarity measure introduced by us and called neighborhood overlap (Sec. \ref{sec:hier_nucl_methods}). 
Through these analyses, we confirm the striking differences between different architectures highlighted by the intrinsic dimension analysis. In convolutional networks, we find that the neighborhood composition changes very smoothly across most of the layers, and the final ordered structures arise abruptly close to the output, while in the Vision Transformer, the formation of ordered nuclei is more gradual.
We then used the approach introduced in ref. \citet{cluster} to characterize the evolution of the probability peaks across the hidden layers. 
The semantic richness of ImageNet is organized hierarchically, from general concepts like animals and objects to intermediate-level concepts like animal families to specific breeds of animals. 
We find that deep networks trained to classify ImageNet (convolutional and transformers) first recognize the general concepts, splitting the probability landscape into a number of modes equal to the number of \enquote{macro classes} in the data set, and then divide each probability peak according to the fine-grained classes it contains.
The hierarchical splitting of the categories leaves a footprint in the output representations where the probability modes of semantically similar classes are connected by saddle points of high density.
A more thorough analysis of the density topography of the final layer can be used to connect our work with other studies that tried to understand how neural networks extract semantic relationships \cite{saxe_semantic_development, semantic_cognition}.
\citet{saxe_semantic_development} described how semantic knowledge arises in deep linear networks and characterized the final organization of the objects in the hidden representations of the network. 
The semantics of the representation can be characterized by the notions of typical, prototypical, and coherent classes  \citet{saxe_semantic_development}.
These notions are tightly connected to some aspects of the density landscape of the final layer of Resnet152 (see Sec. \ref{sec:hier_nucl_evolution_cluster_ari}).

Coherent classes are characterized by tight clusters highly distinct from the other classes \cite{saxe_semantic_development}. A coherent category could therefore be identified by a cluster divided from the rest by a high-energy barrier i.e., in which the difference between the density of the peak and the density of the saddle points is large for all the neighboring clusters. An example of a coherent class is that of dogs (see Fig. \ref{fig:dendogram_b}, left branch).
A typical category is instead one that expresses at best the characteristics of a group, for instance, the sparrow is a more "typical" bird than a penguin \cite{saxe_semantic_development}. 
In our setting, the cluster of the most typical birds should have the highest saddle point density in the output representation, because this implies that they are separated from the rest of the birds by small "free energy barriers". 
This definition implies that typical categories are those split in the latest stages of the architecture because it is harder for the network to identify those "atypical" features required to divide them from the rest of the family. 
In this sense, the penguins should be divided more easily from the rest of the birds, and possibly some layers before the sparrows.
In our case, the role of the penguins is played by the butterflies (Admiral, Ringlet, represented with yellow leaves in the right-most branch of the dendrogram in Fig. \ref{fig:dendogram_b}). The network puts the butterflies between birds and insects, but closer to the bird classes. 
The butterfly class is also "coherent" and it is learned in early layers (see layer 145 of Fig. \ref{fig:app_layers_142_148} where the cluster of admiral butterflies is already formed).
Finally, a prototypical bird would be, loosely speaking, the one that summarizes at best the properties of all the birds from penguins to sparrows, and can be defined from a weighted average of the features of all the birds \cite{saxe_semantic_development}.
Perhaps, the prototypical bird can be found earlier in the network, in the last layer before the nucleation of the specific breeds of birds, where all the birds belong to the same cluster. 
In this representation, the prototypical bird would be the density peak, or, alternatively, the medoid of the cluster.
These are just tentative hypotheses and further research is needed to establish a sound connection between the geometrical aspects of the representations analyzed in Chapter \ref{ch:hier-nucl} and the theory of semantic cognition \cite{cognition1, cognition2, cognition3}.
The computational cost for the analysis of the probability density described in Chapter \ref{ch:hier-nucl} is dominated by the time of a forward pass to extract the activations plus the time to compute the distance matrix. 
For the convolutional models, the actual bottleneck is the computation of the distance matrices, which can take up to 10 hours on representations with $P = 5 \cdot 10^5$ activations and a data set size $N = 10^5$. 
Increasing further the input resolution can be very expensive (we recall that $P$ scales quadratically with the size of the image (see Sec. \ref{sec:id_resnet152})), limiting the applicability of the method on convolutional networks trained on high-resolution inputs.
A possible solution can be averaging the activations of each channel, reducing the number of features in the hidden representations from $10^5$ to $10^2$/$10^3$ and the overall computational cost factor of 100. 
This approach would be equivalent to that used in the analysis of the transformer architectures, where we averaged and pooled the representations along the sequence dimension (see Sec. \ref{sec:hier_nucl_clusters_alex_vgg_vit_igpt} and Sec. \ref{sec:imagenet_igpt}).

The pipeline for the analysis of the density peaks presented in Chapter \ref{ch:hier-nucl}  is efficient and allows the analysis of models as big as iGPT, which has more than one billion parameters, in a few hours. 
The methods developed and used in this Thesis used are all available at \href{https://github.com/sissa-data-science/DADApy}{https:// github.com/sissa-data-science/DADApy} together with a set of easy-to-run Jupyter notebooks, an online manual, and an extensive code reference. Further additional information about the methods can be found in the companion paper \cite{me_dadapy}.

We are currently using neighborhood overlap and density-based clustering in the context of transfer learning. 
The knowledge of the internal structures and the similarity between representations in networks trained on different tasks can be used to compare the similarity between tasks or data sets and to decide at
which \emph{intermediate} layer the features extracted from the source task can be effectively transferred to a target task. 
%

Chapter \ref{ch:repr-mitosis} is also devoted to the analysis of the representations of deep neural networks. We proposed a mechanism to interpret the scaling law of state-of-the-art deep neural networks trained on image data sets. 
We showed how the low-dimensional structure of the last hidden layer representation of wide neural networks can be thought of as the composition of what we call "clones". 
Clones are groups of activations that achieve perfect classification accuracy on the training set and differ from the full layer representation for an affine transformation up to uncorrelated random noise.
We hypothesize that the scaling law of generalization error with width $W$ as 
$W^{\nicefrac{-1}{2}}$  empirically observed for wide networks in the over-parametrized regime (Fig. \ref{fig:error-full-network}) can be the effect of an implicit combination of almost independent representations of the same manifold. This combination reduces the prediction error in the same way an additional measure reduces the measurement uncertainty, leading to the $\nicefrac{-1}{2}$ decay.
We also show that the scaling law of the generalization error and the dynamics of networks trained with the standard protocols \cite{bello2021revisiting, 2017densenet, cubuk2020randaugment} to achieve the best classification accuracy on CIFAR10 and CIFAR100, in particular, trained with weight decay, is quantitatively different from that of not regularized networks (Fig. \ref{fig:narrow_bad-trained}) which are very often studied in theoretical works \cite{jacot2018neural,du2018gradient,chizat2019lazy, geiger2020scaling}. 

\vspace{7pt}

We end this tour through the hidden pathways of artificial neural networks with some questions that remain unanswered.
In this Thesis, we have mainly analyzed deep networks trained with supervised objectives.
We have seen that the geometry and organization of the hidden representations of deep networks trained with labels are determined both by the low-dimensional structure of the data and by the human priors acting through the supervised cross-entropy loss.
We have also seen that a network trained without labels can develop human-like semantics in some internal representations.
What is the influence of human intelligence in finding architectures that produce human-interpretable representations, and what is the impact of these architectural choices on generalization? 
Is the iGPT example representative of other networks and tasks trained with semi-supervised or unsupervised losses? 
We also saw that the intrinsic dimension of the hidden representation can give a hint about the semantics of a hidden layer. 
In the general case, what are good measures to understand the semantics of hidden representations when labels are not available?






\printbibliography[heading=bibintoc]


\end{document}